\documentclass[letterpaper]{article} % DO NOT CHANGE THIS
\usepackage{aaai25}  % DO NOT CHANGE THIS
\usepackage{times}  % DO NOT CHANGE THIS
\usepackage{helvet}  % DO NOT CHANGE THIS
\usepackage{courier}  % DO NOT CHANGE THIS
\usepackage[hyphens]{url}  % DO NOT CHANGE THIS
\usepackage{graphicx} % DO NOT CHANGE THIS
\usepackage{natbib}  % DO NOT CHANGE THIS AND DO NOT ADD ANY OPTIONS TO IT
\usepackage{caption} % DO NOT CHANGE THIS AND DO NOT ADD ANY OPTIONS TO IT
\usepackage{algorithm}
\usepackage{algorithmic}
\usepackage{xcolor}
\usepackage{amsmath, amsfonts}
\usepackage{multirow}
\usepackage{booktabs}
\usepackage{subcaption}
\usepackage{adjustbox}
\usepackage{arydshln}
\usepackage{pifont}
\usepackage{url}

\usepackage{newfloat}
\usepackage{listings}
\DeclareCaptionStyle{ruled}{labelfont=normalfont,labelsep=colon,strut=off} % DO NOT CHANGE THIS
\floatstyle{ruled}
\newfloat{listing}{tb}{lst}{}
\floatname{listing}{Listing}
\title{DuMo: Dual Encoder Modulation Network for Precise Concept Erasure}
\author{
    Feng Han\textsuperscript{\rm 1,\rm 2}, Kai Chen\textsuperscript{\rm 1,\rm 2}, Chao Gong\textsuperscript{\rm 1,\rm 2}, Zhipeng Wei\textsuperscript{\rm 1,\rm 2}, Jingjing Chen\textsuperscript{\rm 1,\rm 2}\thanks{Corresponding author.},\\ Yu-Gang Jiang\textsuperscript{\rm 1,\rm 2}
}
\affiliations{
    \textsuperscript{\rm 1}Shanghai Key Lab of Intell. Info. Processing, School of Computer Science, Fudan University \\
    \textsuperscript{\rm 2}Shanghai Collaborative Innovation Center on Intelligent Visual Computing \\
    \{cgong20,chenjingjing,ygj\}@fudan.edu.cn, \{fhan23,kchen22,zpwei21\}@m.fudan.edu.cn
}

\usepackage{bibentry}
\begin{document}

\maketitle

\begin{abstract}
The exceptional generative capability of text-to-image models has raised substantial safety concerns regarding the generation of Not-Safe-For-Work (NSFW) content and potential copyright infringement. To address these concerns, previous methods safeguard the models by eliminating inappropriate concepts. Nonetheless, these models alter the parameters of the backbone network and exert considerable influences on the structural (low-frequency) components of the image, which undermines the model's ability to retain non-target concepts. In this work, we propose our \textbf{Du}al encoder \textbf{Mo}dulation network (DuMo), which achieves precise erasure of inappropriate target concepts with minimum impairment to non-target concepts. In contrast to previous methods, DuMo employs the \textbf{E}raser with \textbf{PR}ior Knowledge (EPR) module which modifies the skip connection features of the U-NET and primarily achieves concept erasure on details (high-frequency) components of the image. To minimize the damage to non-target concepts during erasure, the parameters of the backbone U-NET are frozen and the prior knowledge from the original skip connection features is introduced to the erasure process. Meanwhile, the phenomenon is observed that distinct erasing preferences for the image structure and details are demonstrated by the EPR at different timesteps and layers. Therefore, we adopt a novel \textbf{T}ime-\textbf{L}ayer \textbf{MO}dulation process (TLMO) that adjusts the erasure scale of EPR module's outputs across different layers and timesteps, automatically balancing the erasure effects and model's generative ability. Our method achieves state-of-the-art performance on Explicit Content Erasure (detecting only 34 nude parts), Cartoon Concept Removal (with an average $\mathrm{LPIPS}_\mathrm{da}$ of 0.428, 0.113 higher than SOTA at 0.315), and Artistic Style Erasure (with an average $\mathrm{LPIPS}_\mathrm{da}$ of 0.387, 0.088 higher than SOTA at 0.299), clearly outperforming alternative methods. Code is
available at \url{https://github.com/Maplebb/DuMo}
\end{abstract} 


\section{Introduction}
%Exisitng text-to-image models may generate insecure content and reproducing famous artworks. 
   % (a) Previous methods fail to protect non-target characteristics such as the hat on the person. The DuMo framework we propose can conduct precise erasure on target concepts to safeguard the models with minimal devastation on non-target generative ability.
\begin{figure*}[t]
    \centering
    \includegraphics[width=\linewidth]{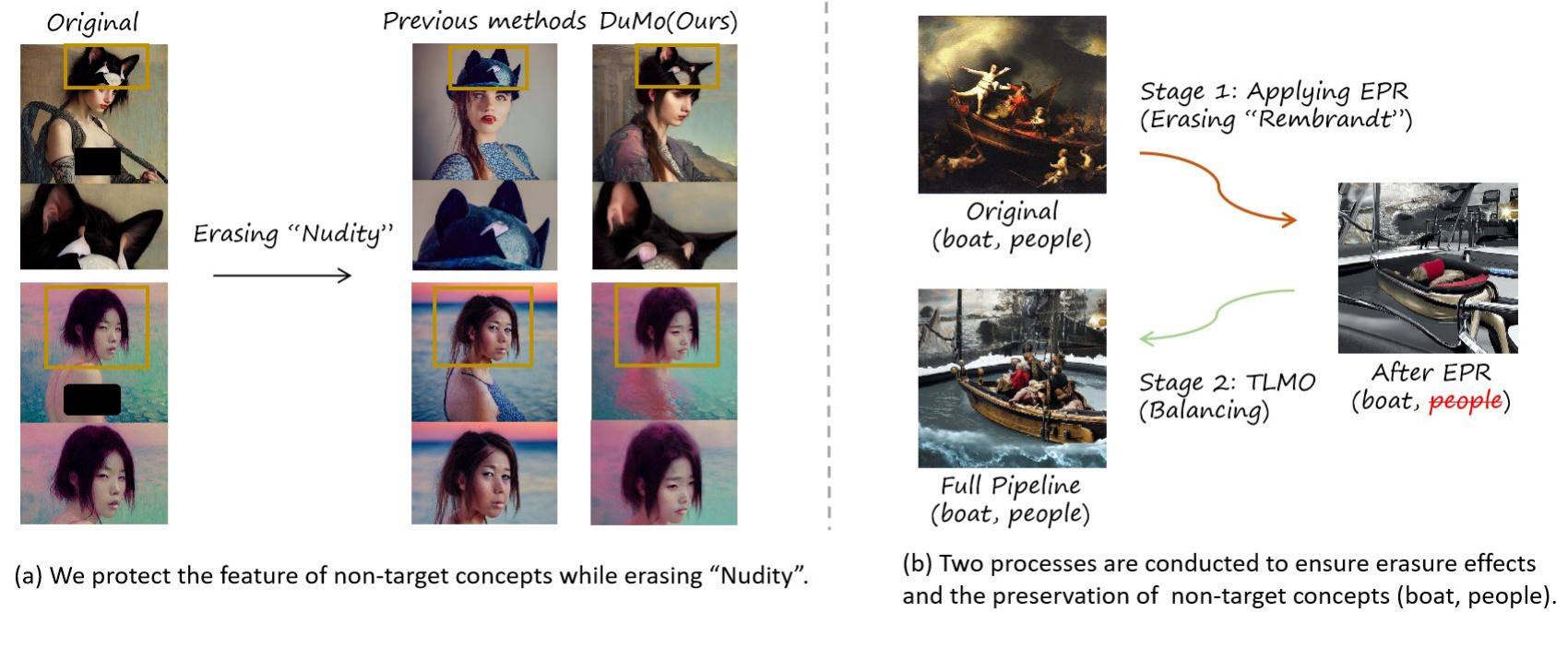}  
    \vspace{-0.8cm}
    \caption{(a) Compared to previous methods, our DuMo framework can precisely erase target concepts to safeguard the models with minimal devastation on non-target concepts' generative ability. (b) While erasing ``Rembrandt", the trade off between erasing and preserving non-target concepts is perfectly accomplished by two steps ``erasing and balancing" procedure. We cover the nude content with \rule{0.8cm }{0.25cm} to prevent negative public influence.
    %The parameters of the original U-Net are frozen to ensure the integrity of the backbone feature and erasure effects are applied to the skip connection feature with the EPR module and Modulation process.(b) The modulation factors operates on the output of the EPR module to band the generative ability of the DM.
    }
    \label{fig:display}
    
\end{figure*}

Recent advancements in text-to-image (T2I) diffusion models~\cite{dhariwal2021diffusion,nichol2021glide,ramesh2022hierarchical,nichol2021improved, saharia2022photorealistic} have made notable progress in creating high-quality images based on textual prompts within seconds. This is primarily attributed to the extensive web-scale datasets for model pre-training. However, the advanced generation capability of the models is accompanied by a number of potential risks including copyright infringement and the propagation of Not-Safe-For-Work (NSFW) content.

Recent studies demonstrate that diffusion models tend to imitate famous artworks and specific painting styles of artist~\cite{jiang2023ai,setty2023ai} or generate sexually explicit content~\cite{hunter2023ai,zhang2023generate}, which violates the societal norms and legal regulations. To tackle this problem, a naive approach is to filter out the inappropriate data of the datasets and retrain the model from scratch. Nevertheless,
this is not only resource-consuming but also unsatisfactory in terms of the results. For instance, Stable Diffusion v2.0 continues to generate explicit content while being trained on a sanitized dataset. Besides, both the interference of classifier-free guidance during generation time~\cite{schramowski2023safe} and the safety checker~\cite{rando2022red} afterwards can be easily circumvented ~\cite{SmithMano2022,fan2023salun,shi2020improving}.

Therefore, recent methods either concentrate on parameter fine-tuning or developing erasure-specific modules to the U-NET~\cite{gandikota2023erasing,fan2023salun,heng2024selective,kumari2023ablating,gong2024reliable,huang2023receler}. Although these methods are effective to the target concept, they alter the backbone features of the U-Net decoder and severely sacrifice the ability of generating non-targeted concepts. In each stage of the U-Net decoder, the skip features from the skip connection and the backbone features are concatenated together. Nevertheless, the skip connection features and backbone features exhibit distinct characteristics during the denoising process. Notably, it is discovered by FreeU~\cite{si2024freeu} that the backbone features have much relevance to the structural (low-frequency) components of the generated image, while the skip connection features are more related to the style and details (high-frequency)~\cite{si2024freeu}. Altering the backbone features of the U-Net decoder engenders harmful effects to the generation of the non-target concepts, compromising their structural integrity (Fig. \textcolor{red}{\ref{fig:display}a}). Besides, the potential of the skip connection features is not fully exploited, which refrains the model from fulfilling excellent erasure ability and superb generative ability~\cite{luo2024diff}.

To cope with the aforementioned challenges, we design a noval framework, dubbed \textbf{Du}al encoder \textbf{Mo}dulation network (DuMo), to conduct effective erasure on multiple concepts with minimal devastation on non-target generative ability. We first propose the \textbf{E}raser with \textbf{PR}ior knowledge (EPR) to perform concept erasure. EPR operates on skip connection features, without affecting the backbone features, thereby preserving the structure of non-target concepts. Meanwhile, to avoid significant degradation of generative ability during the erasure process, the original skip connection features from the original U-NET is maintained, which provides implicit semantic information of the unerased concepts. We refer to the implicit semantic information as the prior knowledge. Furthermore, the result displayed in Fig \textcolor{red}{\ref{fig:layercomparison}}, \textcolor{red}{\ref{fig:timecomparison}}, reveals that, at different skip connection layers and denoising timesteps, the outputs of our EPR module show varying erasure preferences for the structural (low-frequency) components and the detail-specific (high-frequency) components of the images. To further enhance the preservation of non-target concepts while ensuring the erasure ability, we design the Timestep-Layer MOdulation process (TLMO). This process introduces timestep-specific and layer-specific factors to automatically determine the erasure scale for each output of the EPR module. 
%Extensive qualitative and quantitative experiments are conducted on three tasks, explicit content erasure, cartoon concept removal and artistic style erasure, to validate the effectiveness of our model. A higher $\mathrm{LPIPS}\mathrm{da}$ indicates a better ability to preserve non-target concepts and achieve precise erasure. In terms of $\mathrm{LPIPS}\mathrm{da}$, we achieve first place in the Cartoon Concept Removal and Artistic Style Erasure experiments, with average improvements of 0.113 (from 0.315 to 0.428) and 0.088 (from 0.299 to 0.387), respectively. Regarding erasure effects, our method detected the fewest nude parts (34) in the 'Nudity' erasure task. Meanwhile, for the erasure-related metric $\mathrm{LPIPS}\mathrm{e}$, we obtained the best scores of 0.538 and 0.539 for "Snoopy" and "Mickey", improving by 0.035 and 0.011 points, and the second-best score of 0.602 for "Spongebob" while simultaneously erasing "Snoopy", "Mickey", and "SpongeBob". In the task of erasing 'Van Gogh' and 'Rembrandt,' we obtain second-best scores of 0.383 and 0.458, respectively. 
Through extensive experiments, DuMo demonstrates superb erasure performance and fabulous generative ability in comparison to state-of-the-art (SOTA) methods. We briefy summarize our contributions as follows:
\begin{itemize}
    \item We propose a novel concept erasure network DuMo that keeps the original U-NET intact and adopts the EPR module to exploit the protential of skip connection features for effective erasure with protection to the non-target concepts.
    \item We investigate different erasure effects of the EPR module to high-frequency and low-frequency components of the image across different timesteps and layers and adopt modulation factors to balance the erasure effects and model's generative capability.
    \item We conduct extensive experiments on three tasks: explicit content erasure, cartoon concept removal and artistic style erasure to validate the effectiveness of DuMO.
\end{itemize}

\begin{figure*}[t]
    \centering
    % \begin{adjustbox}{max width=0.4\textwidth}
    \includegraphics[width=\linewidth]{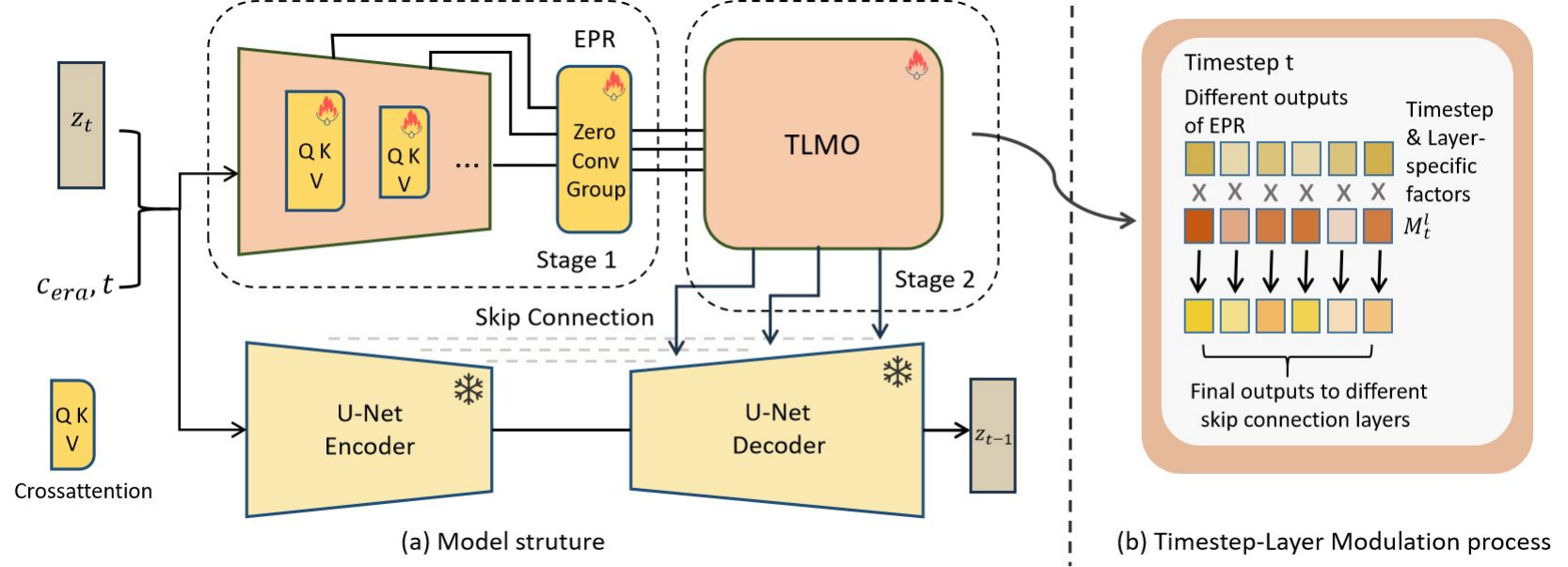}
    \caption{(a) Framework overview. Given a target concept $c_{\text{era}}$, we first fine-tune the EPR module to erase it. In the second stage, we employ TLMO to adjusts erasure effects for each output of the EPR. EPR module and TLMO are applied exclusively to the skip connection features. (b) TLMO applys timestep-layer factors to scale each output of the EPR.
    %The parameters of the original U-Net are frozen to ensure the integrity of the backbone feature and erasure effects are applied to the skip connection feature with the EPR module and Modulation process.(b) The modulation factors operates on the output of the EPR module to band the generative ability of the DM.
    }
    \label{fig:modelstructure}
\end{figure*}

\section{Related Work}
To mitigate copyright infringement and inappropriate content generation, many research efforts have been made recently, including training data filtering for model retraining, 
designing inference time and post-hoc safety mechanisms, as well as parameter fine-tuning and adopting erasure-specific modules for concept erasure. For example, SD 2.0 is retrained on the LAION-5B dataset~\cite{schuhmann2022laion} after filtering out harmful content.
% to enhance the quality of training data, Stable Diffusion 2.0 filter out unsuitable content using an NSFW detector and sanitize the LAION-5B dataset~\cite{schuhmann2022laion} before its training process. 
While this approach aims to reduce harmful outputs,it is not only resource-intensive but also fails to guarantee complete prevention of sexually explicit content generation~\cite{gandikota2023erasing}. Furthermore, such filtering can lead to performance degradation~\cite{schramowski2023safe}.
Unlike training data filtering, SLD~\cite{schramowski2023safe} employs classifier-free guidance 
% and incorporates the pre-existing knowledge to actively direct the model 
to eliminate inappropriate content during inference. However, this approach struggles to balance content quality with effective erasure.
% it's challenging to maintain the quality and fidelity of the generated content while ensuring erasure effects. 
Besides, the safety checker~\cite{rando2022red} for the post-hoc verification is limited in scope and can be easily bypassed~\cite{SmithMano2022}.
% of generated images is limited to a narrow set of categories, and it can be easily circumvented~\cite{SmithMano2022}.

% Recent research has focused on removing specific concepts
Recent research has explored concept removal by fine-tuning model parameters or embedding erasure-specific modules to the U-NET. ESD~\cite{gandikota2023erasing} aligns the probability distributions of the targeted concept with a null string without access to the training data.
% accomplishes concept removal by aligning the probability distributions of the targeted concept with that of a null string without acquiring access to the original training data.
Salun~\cite{fan2023salun} established a hard threshold to create a weight saliency map, identifying critical parameters for erasure. 
% which identify the most critical parameters for effectively erasing a specific concept. 
CA~\cite{kumari2023ablating} aligns the target concept with a broader anchor concept and incorporates a regularization loss term on the surrogate concept. MACE~\cite{lu2024mace} utilizes distinct LoRA modules and refine cross-attention layers using a closed-form solution.
% while preserving the model's original parameters. 
% lightweight
SPM~\cite{lyu2024one}  injects an one-dimensional adapter into the model and adopt the Latent Anchoring strategy and Facilitated Transport mechanism to erase targeted concepts while safeguarding others. However, these methods fail to balance erasure and preservation due to limited information and alterations to backbone features.
% achieve a satisfactory equilibrium between erasure and preservation, due to the reception of limited information and alteration of backbone feature.

\section{Method}

Our gold is to accurately erase specific concepts from a pre-trained DM model, while ensuring that the model retains its generative capability for other non-target concepts. We divide this objective into two components. 
\begin{itemize}
    \item For prompts that explicitly or implicitly include concepts that require removal, the fine-tuned model should maintain the structure of non-target objects while ensuring the erasure effect.
    \item For prompts that do not encompass the concept of erasure, the image generated by the fine-tuned model should be consistent to the pre-trained DM model.
\end{itemize}
To this end, we introduce DuMo, the \textbf{Du}al encoder \textbf{Mo}dulation network. The erasure procedure comprises two stages, the fine-tuning of the EPR module and the \textbf{T}imestep-\textbf{L}ayer \textbf{MO}dulation process (TLMO). Firstly, we set up an additional erasure module, dubbed \textbf{E}raser with \textbf{PR}ior knowledge (Sec. \textcolor{red}{3.1}). EPR is a plug-and-play module that focus full attention on concept erasure. During the erasure process, the prior knowledge of the original skip connection features is utilized to eliminate adverse impairment to non-target concepts. Subsequently, to further improve EPR's ability to retain non-target concepts while guaranteeing the erasure capability, a novel \textbf{T}ime-\textbf{L}ayer \textbf{MO}dulation process (TLMO) is proposed (Sec. \textcolor{red}{3.3}). During the generation of the image, this process leverages timestep-specific and layer-specific factors to adjust the erasure effects of the EPR. Fig. \textcolor{red}{\ref{fig:modelstructure}} overviews our framework.

% At inference time, we utilized the Facilitated Transport mechanism proposed by [] to further leverage the information of user prompt to control the strength of our model.

\subsection{Eraser with Prior Knowledge}
FreeU~\cite{si2024freeu} discovered that for the U-Net of Stable Diffusion, the backbone feature offers a greater abundance of low-frequency semantic information related to the structure of the generated image while denoising. Comparatively, more high-frequency information is provided by the skip connection feature, which is closely associated with the style and details of the generated image. Since our gold is erasing specific concepts while protecting the structure of non-target concepts, We decide to intervene in skip connection without alteration to the backbone features.

Specifically, inspired by~\cite{zhang2023adding}, we freeze the parameters  \(\theta\) of the pre-trained U-Net of DM and make a copy of the encoder block as an external module. EPR (Fig \textcolor{red}{\ref{fig:modelstructure}a}) takes the latent embedding of the image, timestep and text embedding as its inputs and connect its outputs to zero convolution layers, denoted by \(ZeroConv(·; ·)\). \(ZeroConv(·; ·)\) is a 1 × 1 convolution layer with both weight and bias initialized to zeros. The original skip connection features of the original U-Net remains unchanged in the fine-tuning process which provides prior knowledge of the non-target concepts, ensuring generative ability of non-target concepts. The final output of the EPR is computed as 
\begin{equation}
    \mathcal{S}_{c_{\text{era}}}^{t,l}(Z_t, t, \tau_{\theta}(y)) = ZeroConv(\mathcal{E}_{c_{\text{era}}}^{t,l}(z_t, t, \tau_{\theta}(y))|\theta_{zl})
\end{equation}
The final skip connection feature is computed as:
\begin{equation}
    Skip_t^l = x_t^l + \mathcal{S}_{c_{\text{era}}}^{t,l}(z_t, t, \tau_{\theta}(y))
\end{equation}

Here $\theta_{zl}$ denotes parameters of the $l$th zero convolution, $x_t^l$ denotes the $l$th original skip connection feature of the U-Net for a specific denoising timestep $t$, $\mathcal{S}_{c_{\text{era}}}^{t,l}$ denotes the corresponding output of the fine-tuned EPR for concept $c_{\text{era}}$, $z_t$ is latent embedding of the image at timestep t, and $\tau_{\theta}(y)$ is the text embedding of the prompt.

We adopt the erasing loss (\ref{eq:eraloss})~\cite{gandikota2023erasing}, and optimize the parameters $\mathcal{S}_{c_{\text{era}}}$ of EPR module. 
\vspace{-0.2cm}
\begin{multline}
    \mathcal{L}_{\text{era}} = \mathbb{E}_{z_t, t} [ \| \epsilon(z_t, c_{\text{era}}, t \mid \theta, \mathcal{S}_{c_{\text{era}}}) - \epsilon(z_t, c_{0}, t \mid \theta) \\
    + \eta \ast (\epsilon(z_t, c_{\text{era}}, t \mid \theta) - \epsilon(z_t, c_{0}, t \mid \theta))\|_2^2 ].
    \label{eq:eraloss}
    \vspace{-0.7cm}
\end{multline}
Where $c_{\text{era}}$ denotes the concept needs to erase, $\mathcal{S}_{c_{\text{era}}}$ is the parameters of the EPR model for $c_{\text{era}}$, $\theta$ denotes the parameters of pre-trained U-Net, and $c_{0}$ denotes the empty concept `` " that the erasing concept is mapped to. Besides, $\eta$ represents the erasure strength. % and a larger value of $\eta$ signifies a more thorough erasure.

\begin{figure}[t]
    \centering
    \begin{adjustbox}{max width=0.47\textwidth}
    % \begin{tabular}{c@{\hskip 0.03in} c@{\hskip 0.03in} c@{\hskip 0.03in} c@{\hskip 0.03in} c@{\hskip 0.03in} c}
    \begin{tabular}{c: c@{\hskip 0.06in} c@{\hskip 0.03in} c@{\hskip 0.03in} c@{\hskip 0.03in} c@{\hskip 0.03in} c@{\hskip 0.03in}}

        \begin{subfigure}[b]{.11\textwidth}
            \includegraphics[width=\linewidth]{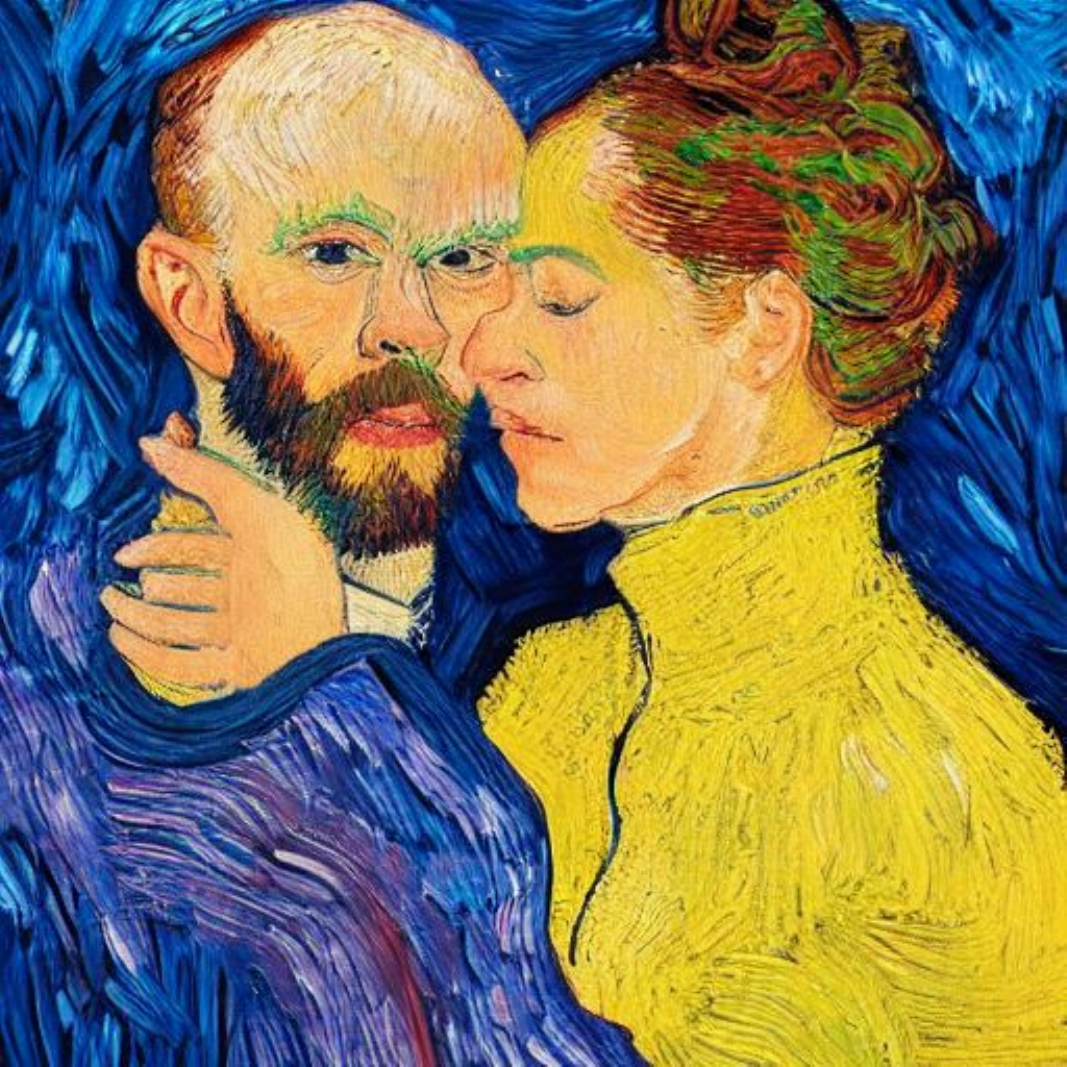}
            \caption{Original}
        \end{subfigure} &
        \begin{subfigure}[b]{.11\textwidth}
            \includegraphics[width=\linewidth]{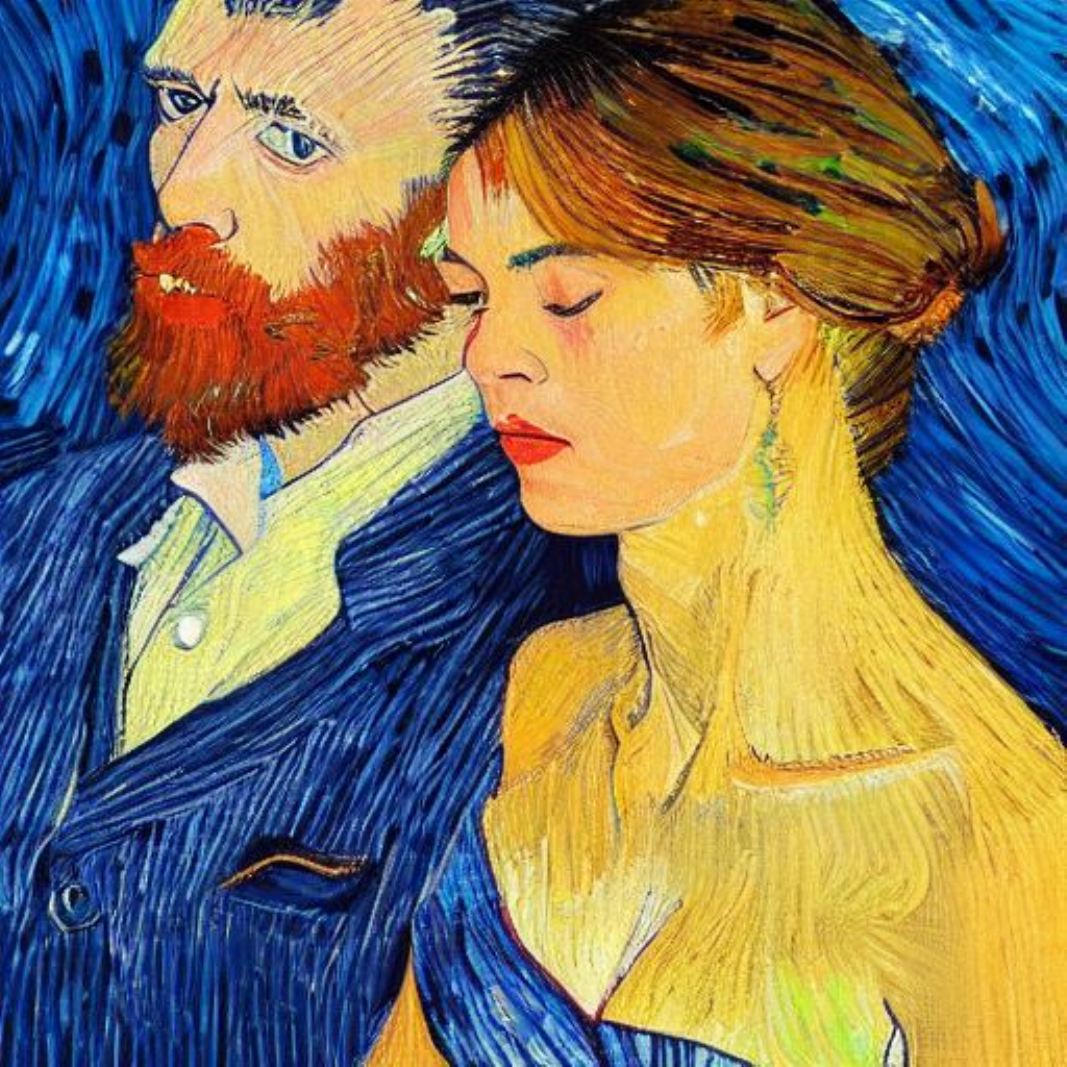}
            \caption{Group 1}
        \end{subfigure} &
        \begin{subfigure}[b]{.11\textwidth}
            \includegraphics[width=\linewidth]{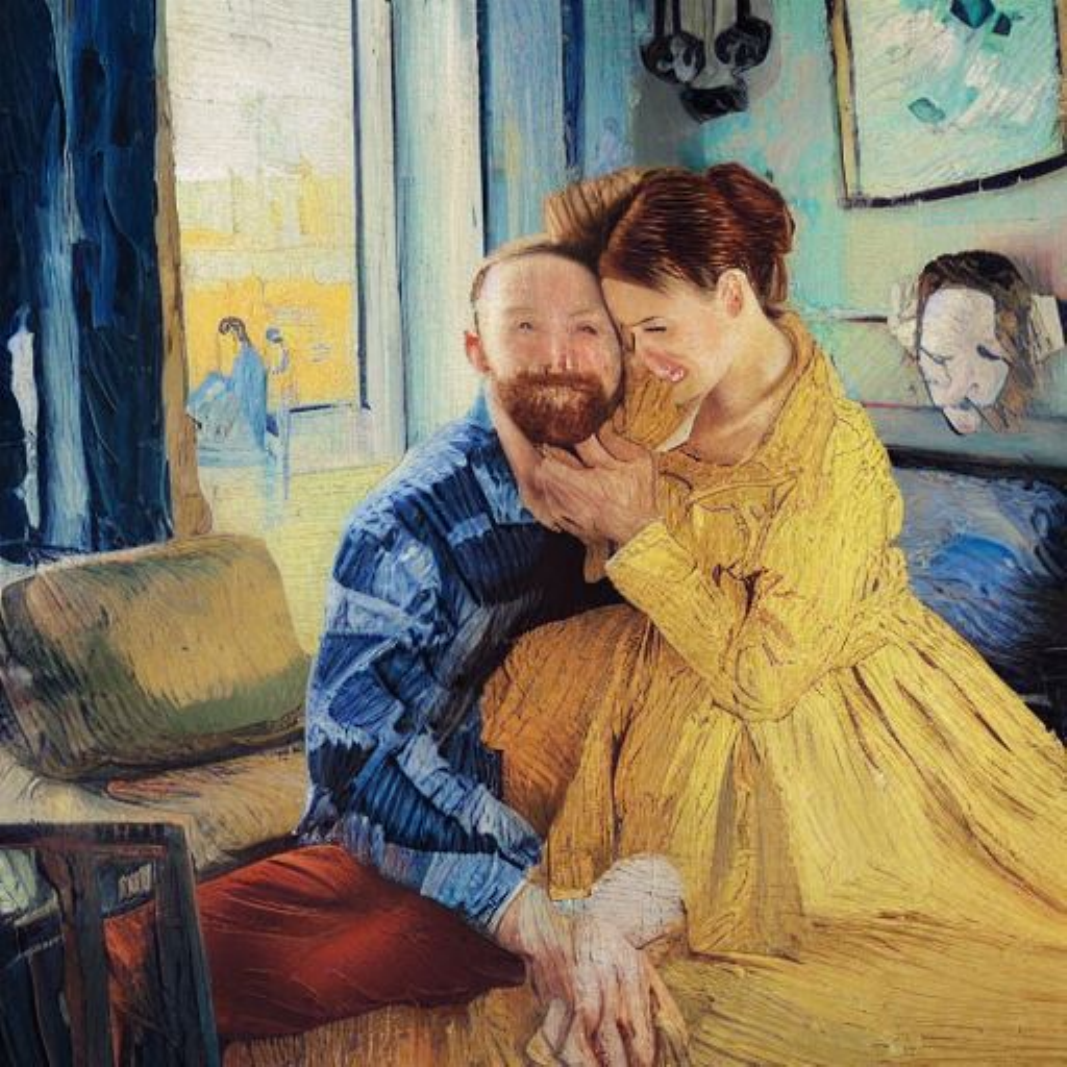}
            \caption{Group 1-2}
        \end{subfigure} &
        \begin{subfigure}[b]{.11\textwidth}
            \includegraphics[width=\linewidth]{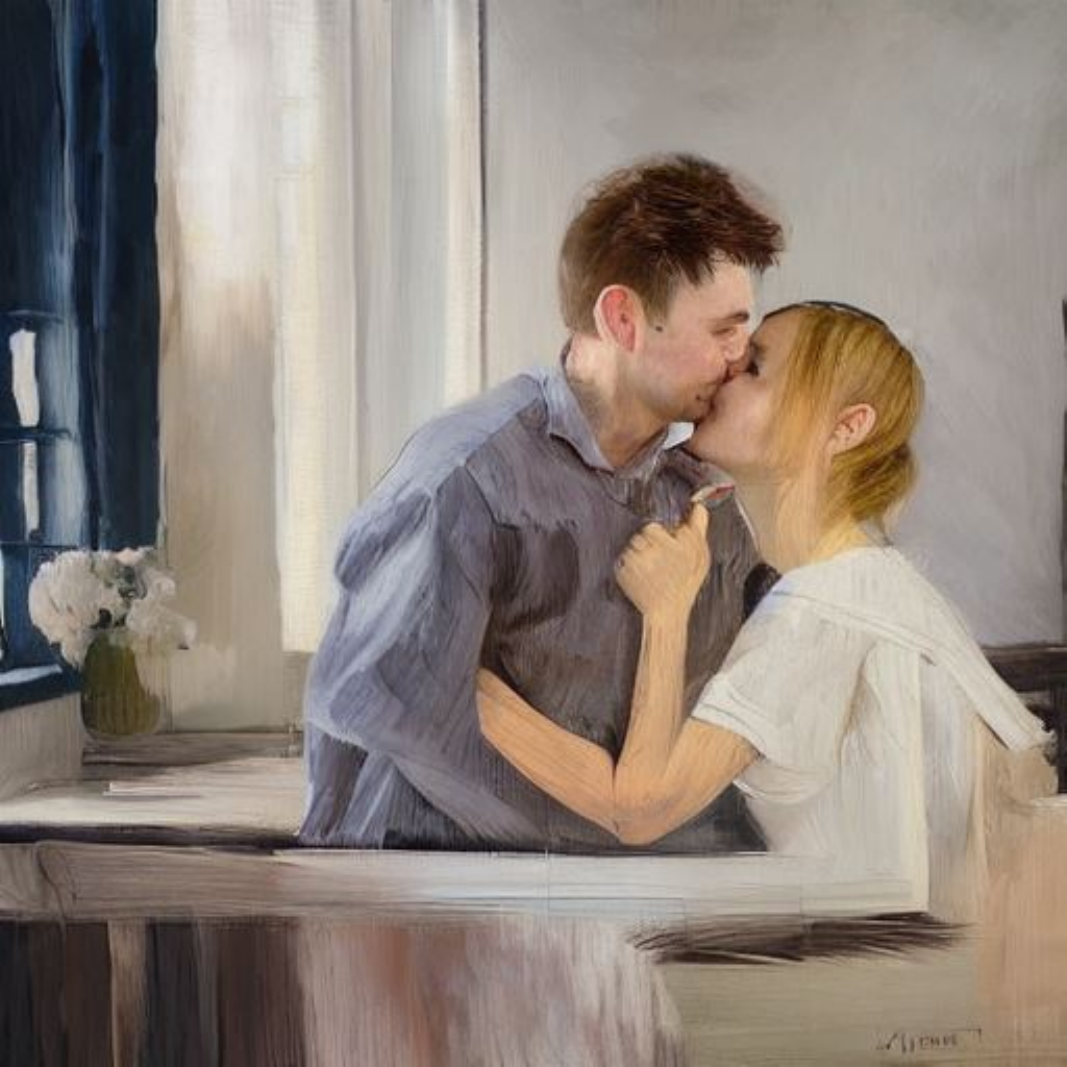}
            \caption{Group 1-3}
        \end{subfigure} &
        \begin{subfigure}[b]{.11\textwidth}
            \includegraphics[width=\linewidth]{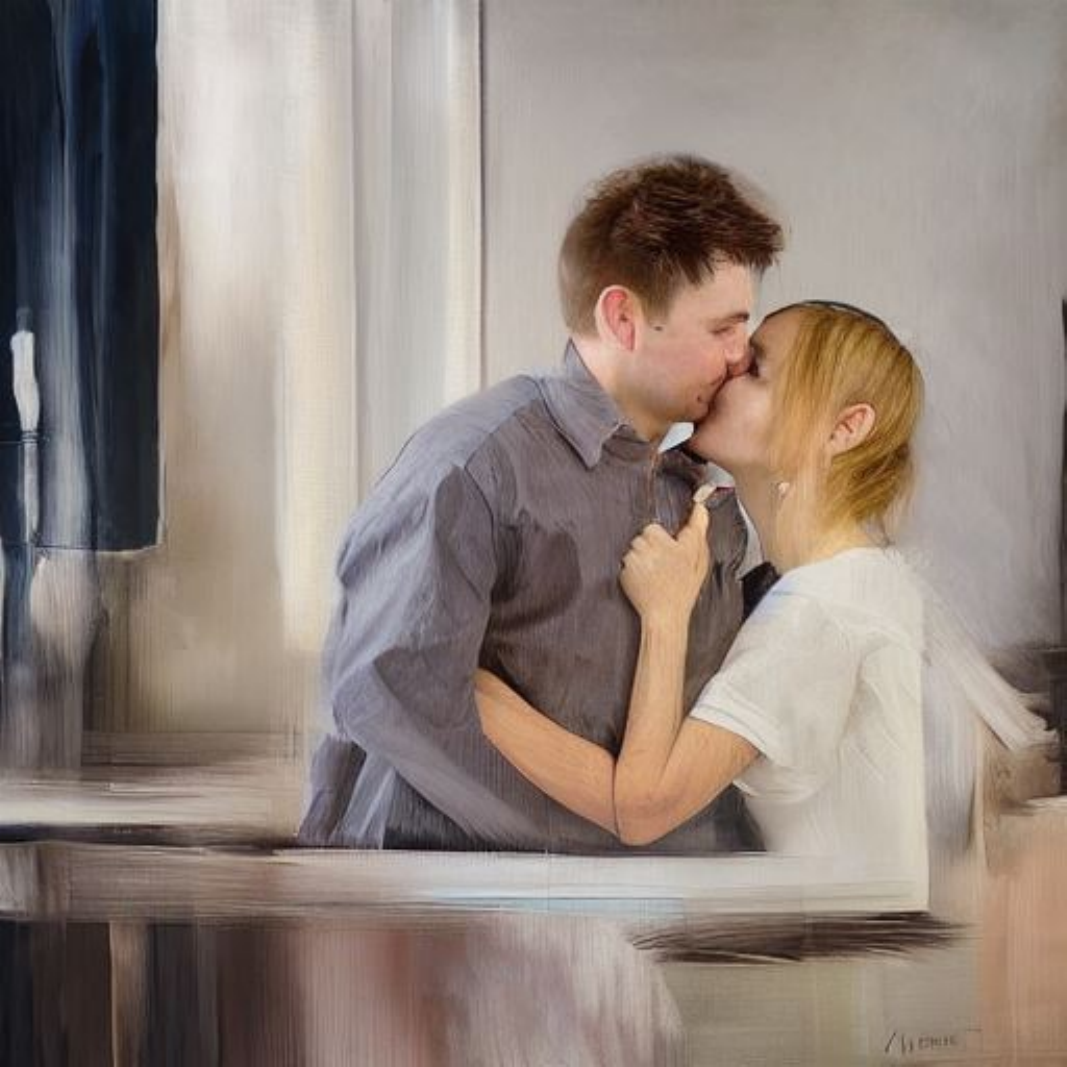}
            \caption{Group 1-4}
        \end{subfigure} &
        \\
        
        &
        \begin{subfigure}[b]{.11\textwidth}
            \includegraphics[width=\linewidth]{images/time-layer-pdf/layer1000.pdf}
            \caption{Group 1}
        \end{subfigure} &
        \begin{subfigure}[b]{.11\textwidth}
            \includegraphics[width=\linewidth]{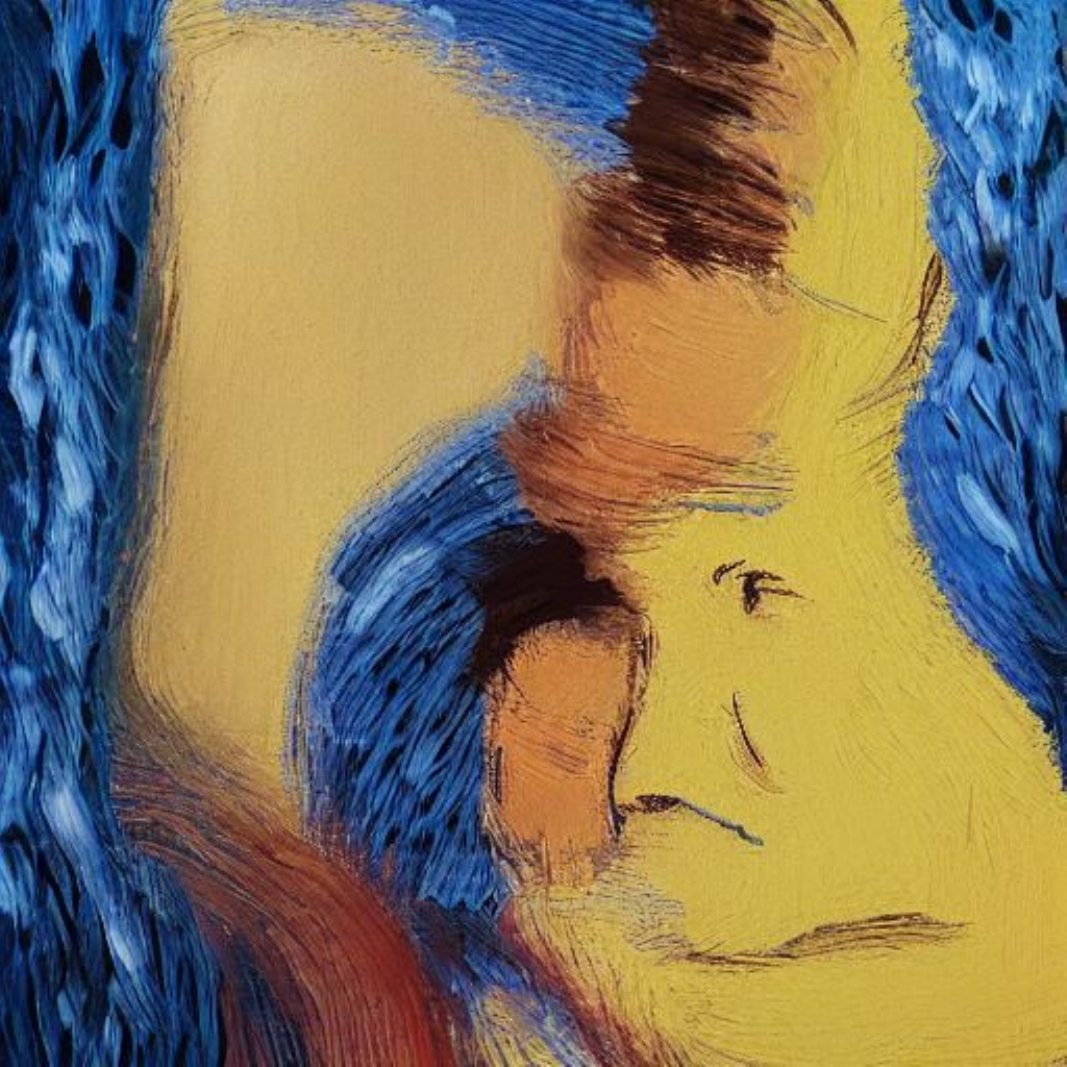}
            \caption{Group 2}
        \end{subfigure} &
        \begin{subfigure}[b]{.11\textwidth}
            \includegraphics[width=\linewidth]{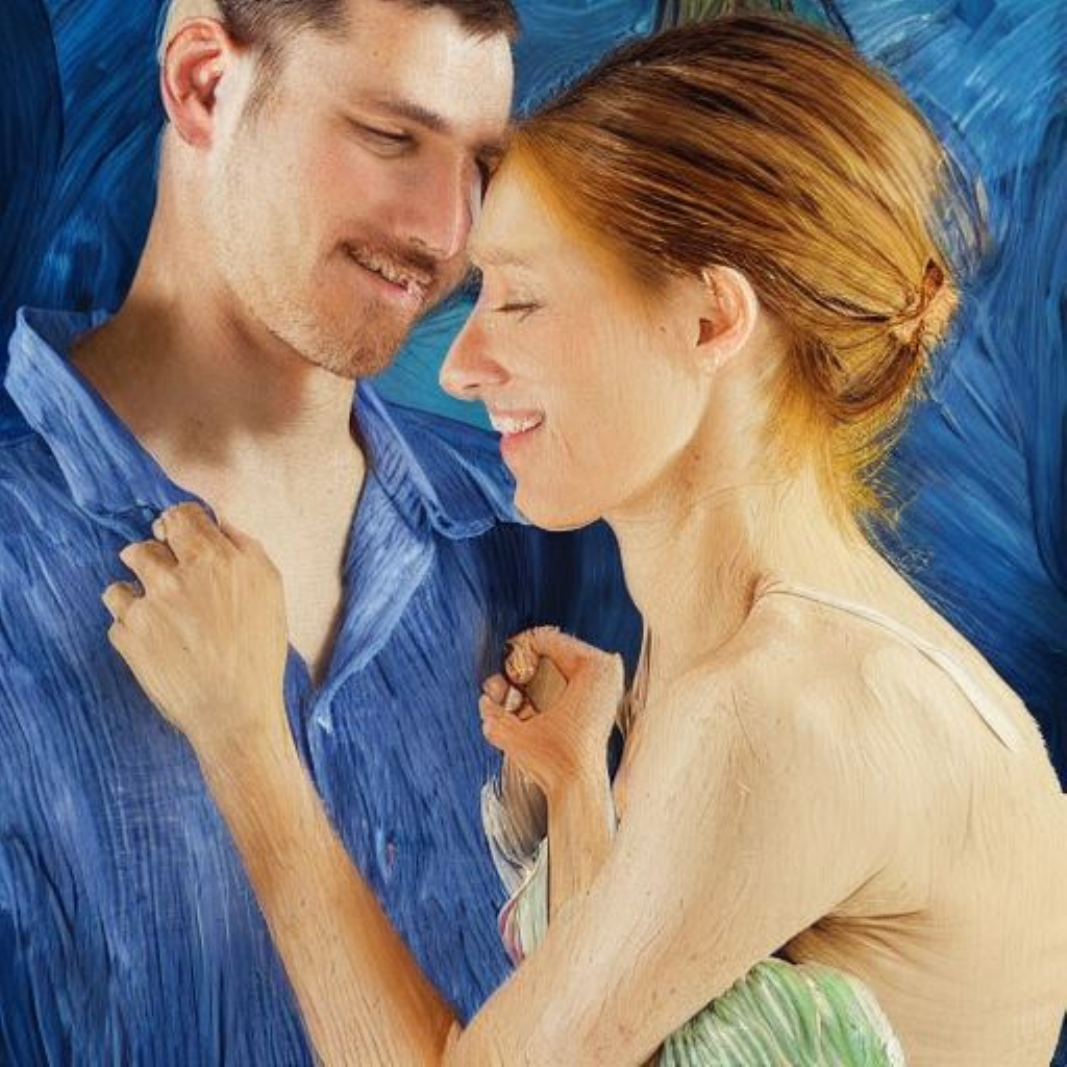}
            \caption{Group 3}
        \end{subfigure} &
        \begin{subfigure}[b]{.11\textwidth}
            \includegraphics[width=\linewidth]{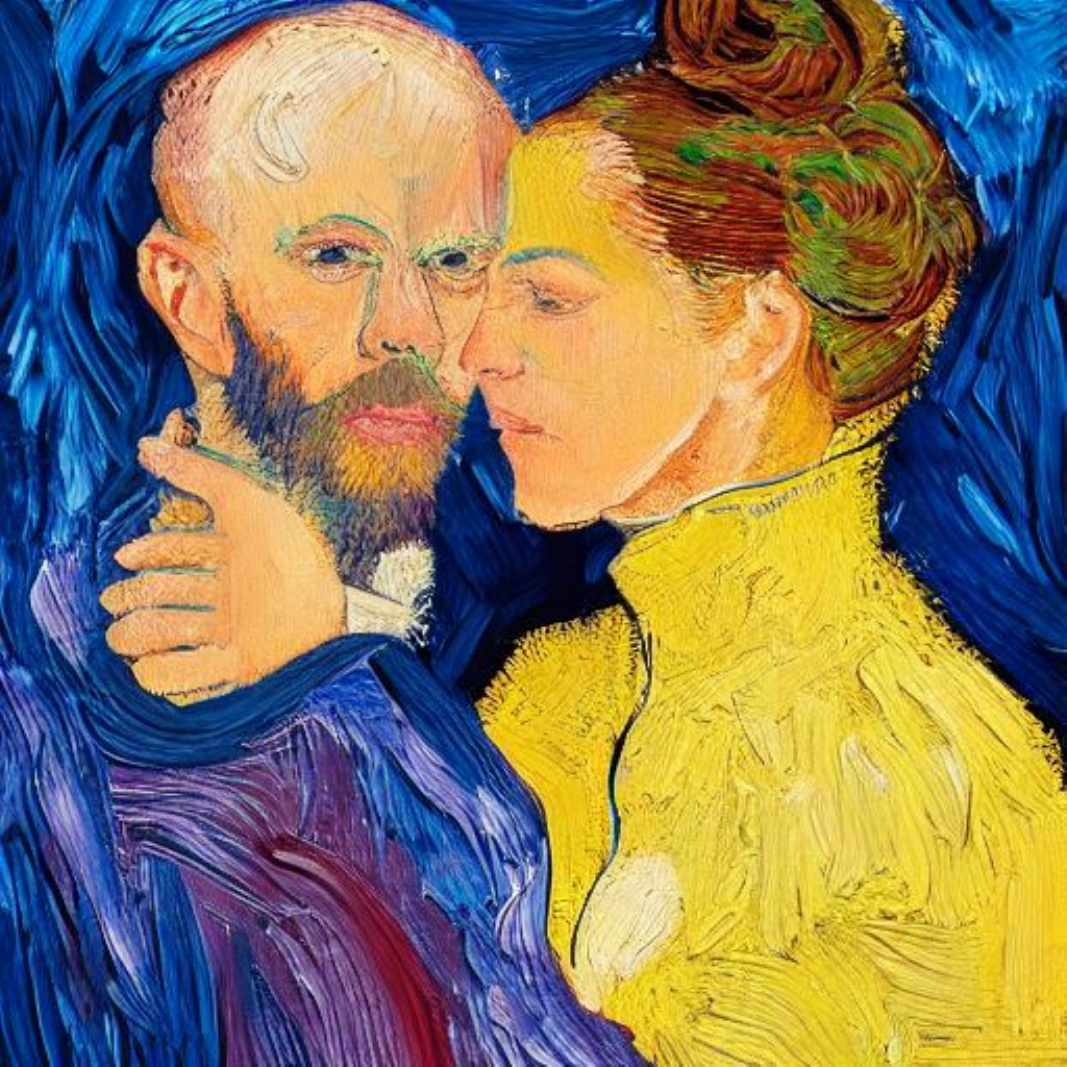}
            \caption{Group 4}
        \end{subfigure} &
        \\

    \end{tabular}
    \end{adjustbox}
    \caption{Comparison of the erasing effect of different skip connection layer groups. The caption indicates that the output of the corresponding skip connection group of the EPR module is added to the original skip connection feature.}
    \label{fig:layercomparison}
    \begin{adjustbox}{max width=0.47\textwidth}
    % \begin{tabular}{c@{\hskip 0.03in} c@{\hskip 0.03in} c@{\hskip 0.03in} c@{\hskip 0.03in} c@{\hskip 0.03in} c}
    \begin{tabular}{c: c@{\hskip 0.06in} c@{\hskip 0.03in} c@{\hskip 0.03in} c@{\hskip 0.03in} c@{\hskip 0.03in} c@{\hskip 0.03in}}

        \begin{subfigure}[b]{.11\textwidth}
            \includegraphics[width=\linewidth]{images/time-layer-pdf/layer0000.pdf}
            \caption{Original}
        \end{subfigure} &
        \begin{subfigure}[b]{.11\textwidth}
            \includegraphics[width=\linewidth]{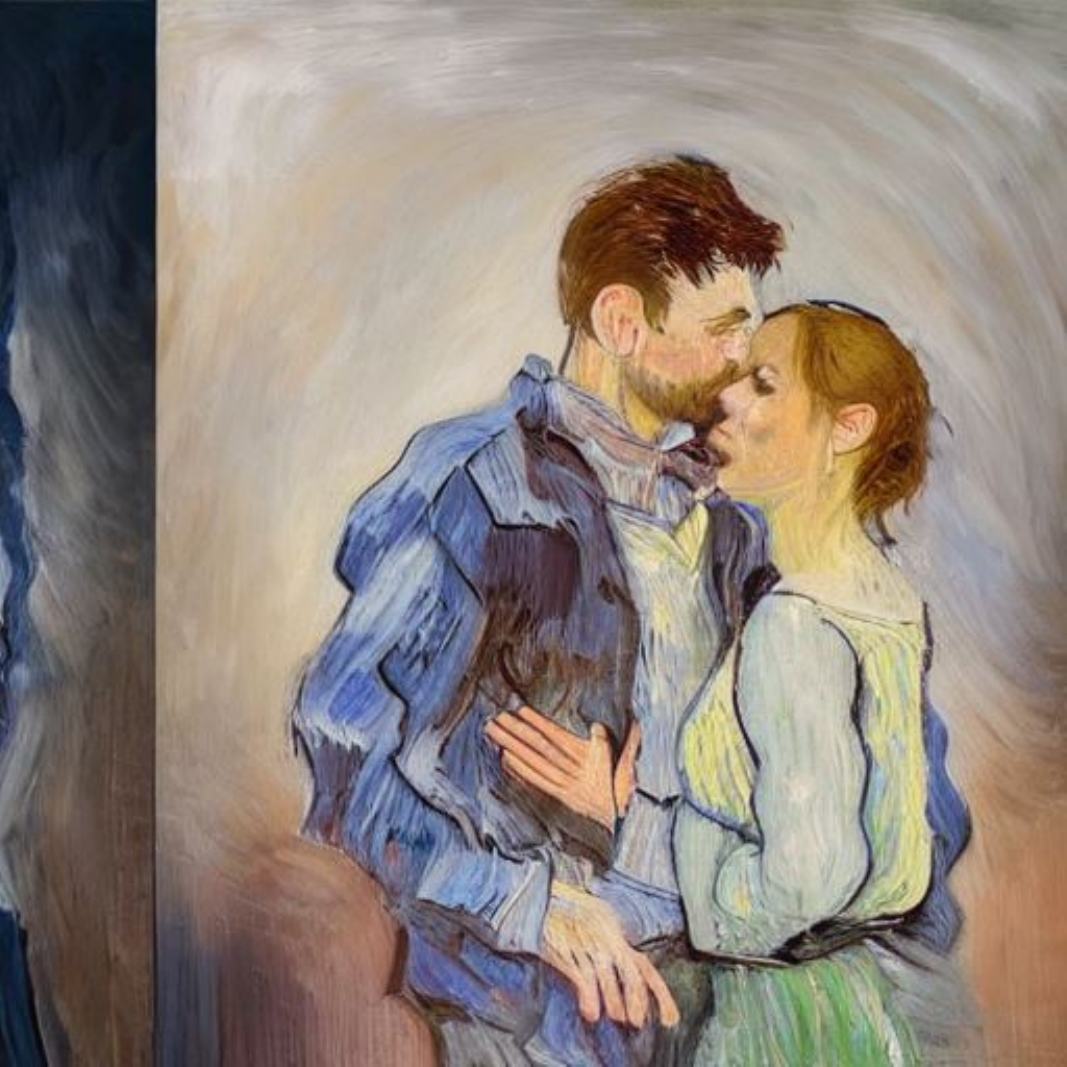}
            \caption{Group 1}
        \end{subfigure} &
        \begin{subfigure}[b]{.11\textwidth}
            \includegraphics[width=\linewidth]{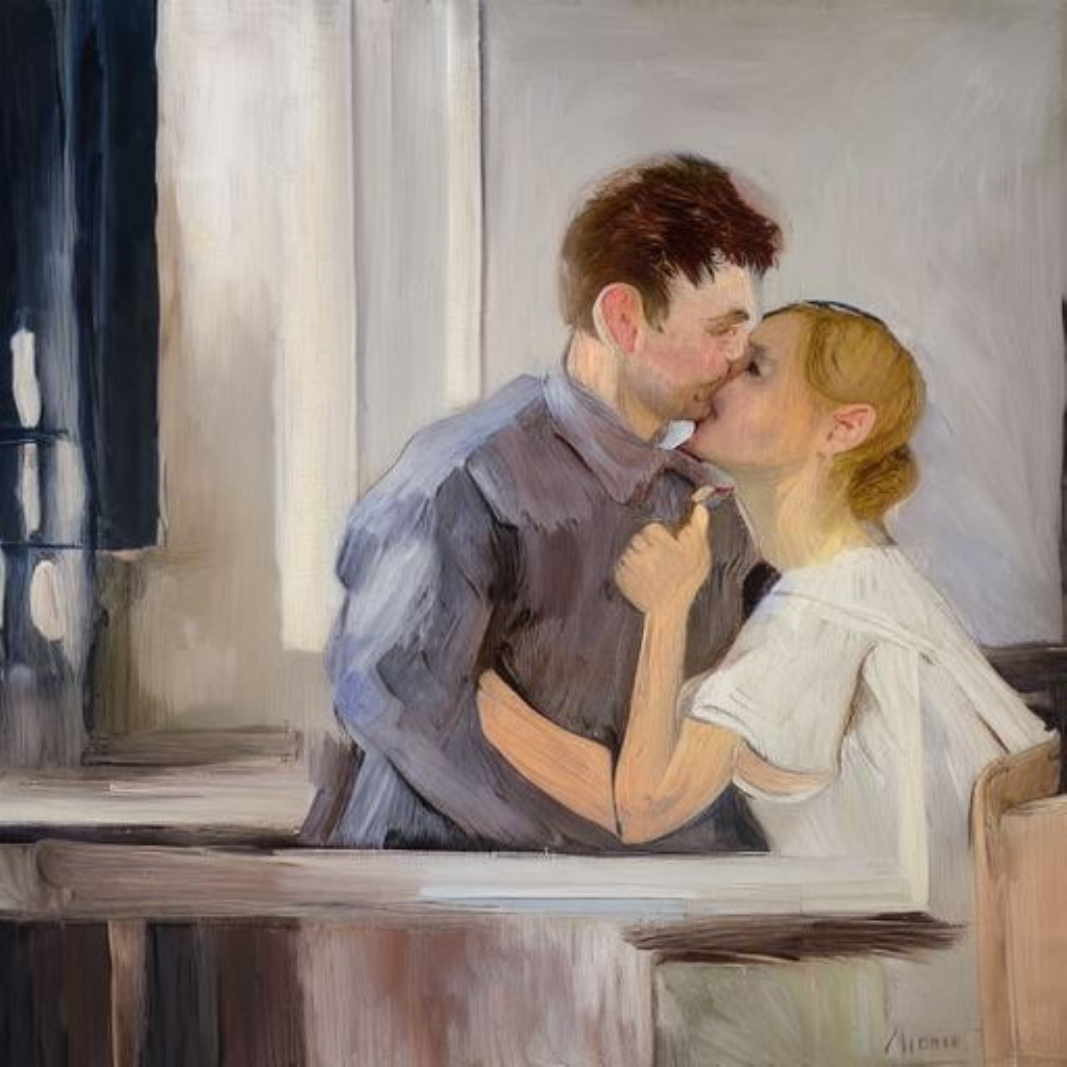}
            \caption{Group 1-2}
        \end{subfigure} &
        \begin{subfigure}[b]{.11\textwidth}
            \includegraphics[width=\linewidth]{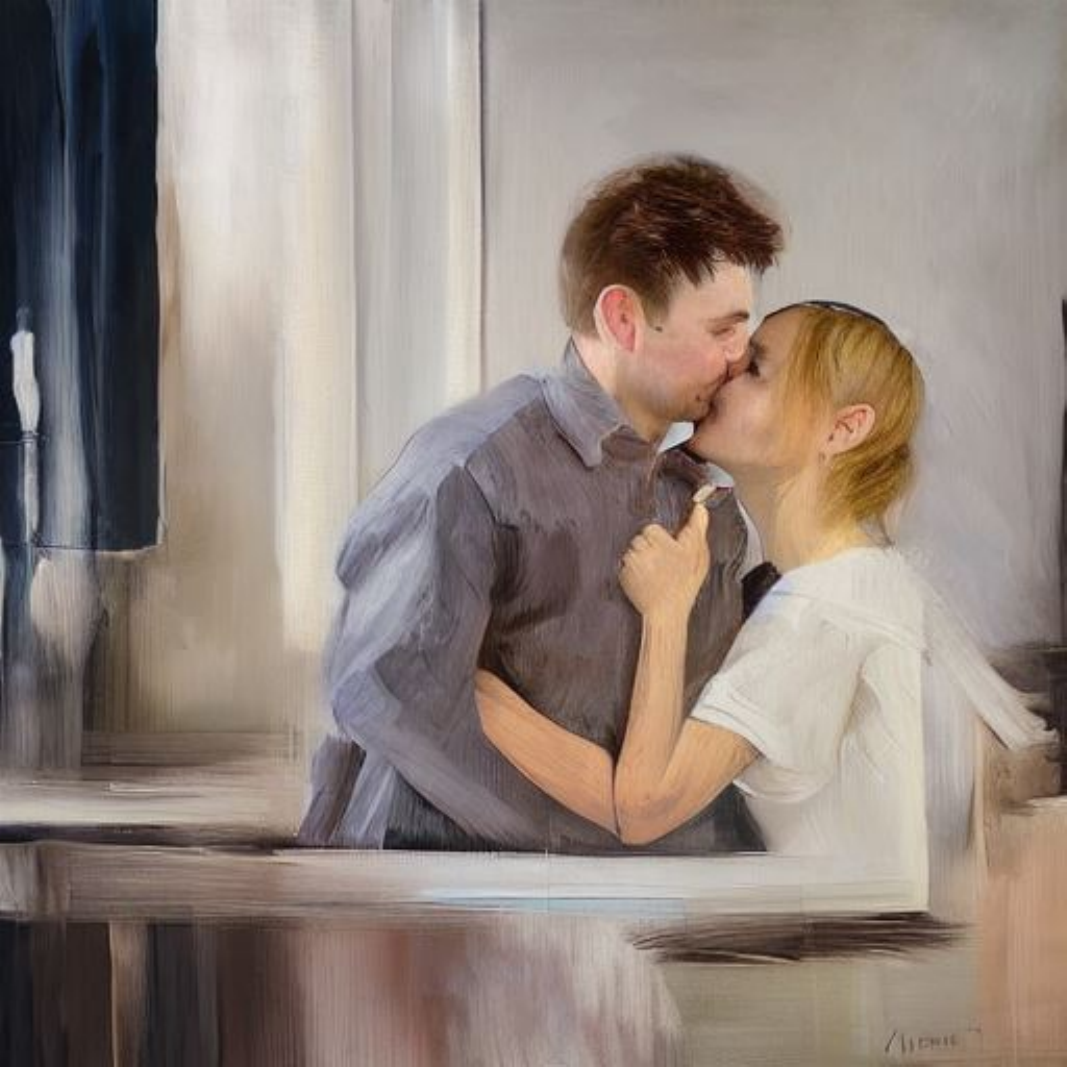}
            \caption{Group 1-3}
        \end{subfigure} &
        \begin{subfigure}[b]{.11\textwidth}
            \includegraphics[width=\linewidth]{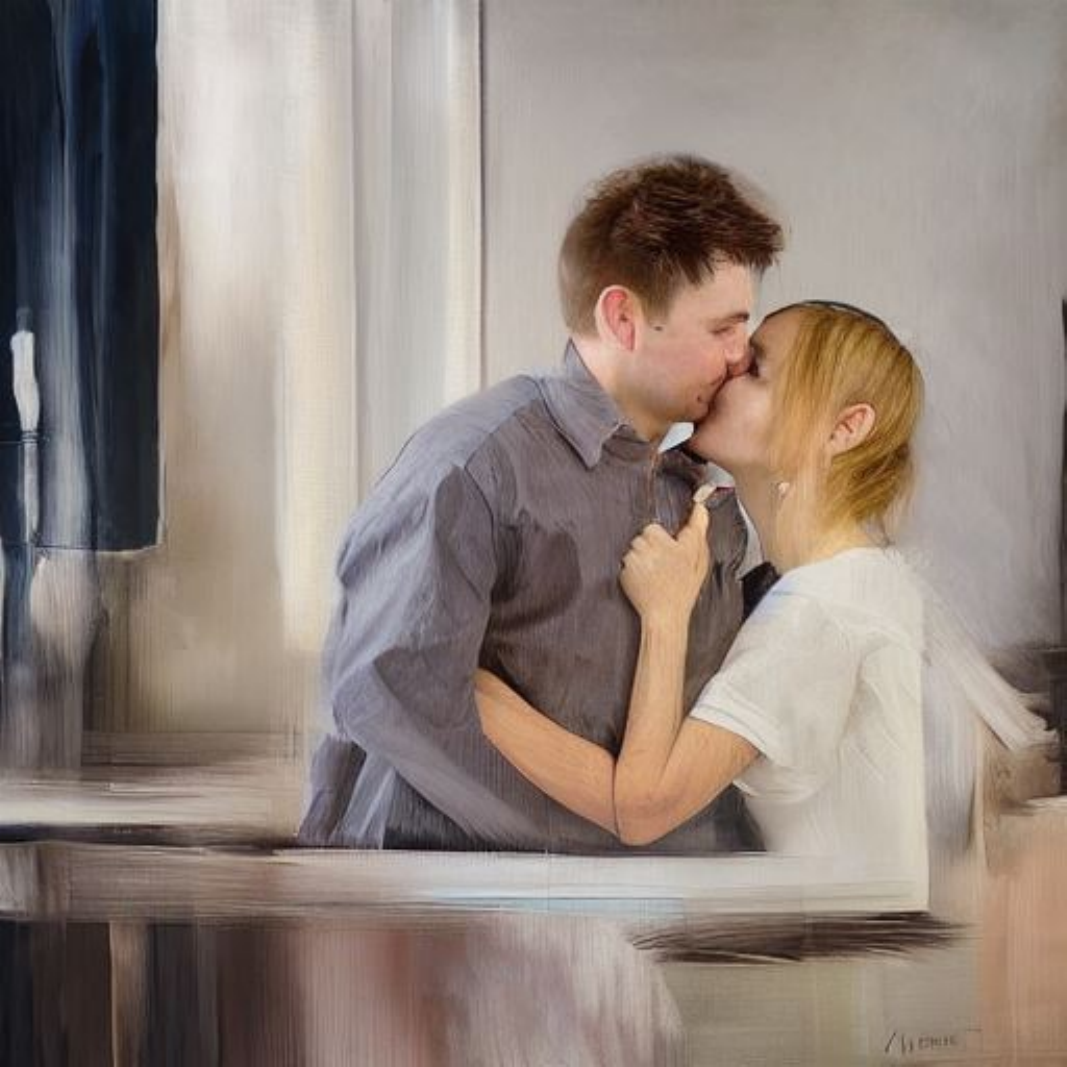}
            \caption{Group 1-4}
        \end{subfigure} & \\

        &
        \begin{subfigure}[b]{.11\textwidth}
            \includegraphics[width=\linewidth]{images/time-layer-pdf/time1000.pdf}
            \caption{Group 1}
        \end{subfigure} &
        \begin{subfigure}[b]{.11\textwidth}
            \includegraphics[width=\linewidth]{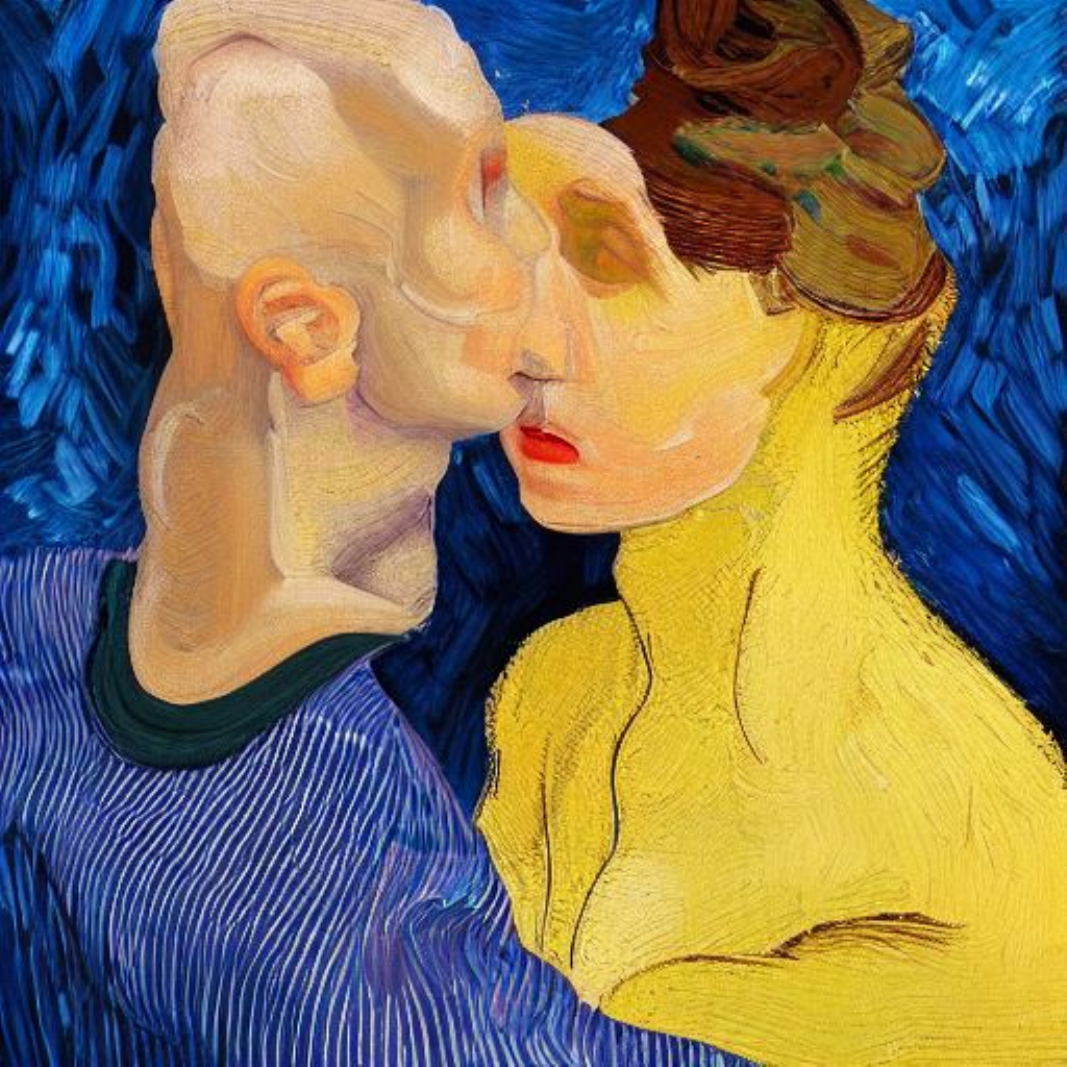}
            \caption{Group 2}
        \end{subfigure} &
        \begin{subfigure}[b]{.11\textwidth}
            \includegraphics[width=\linewidth]{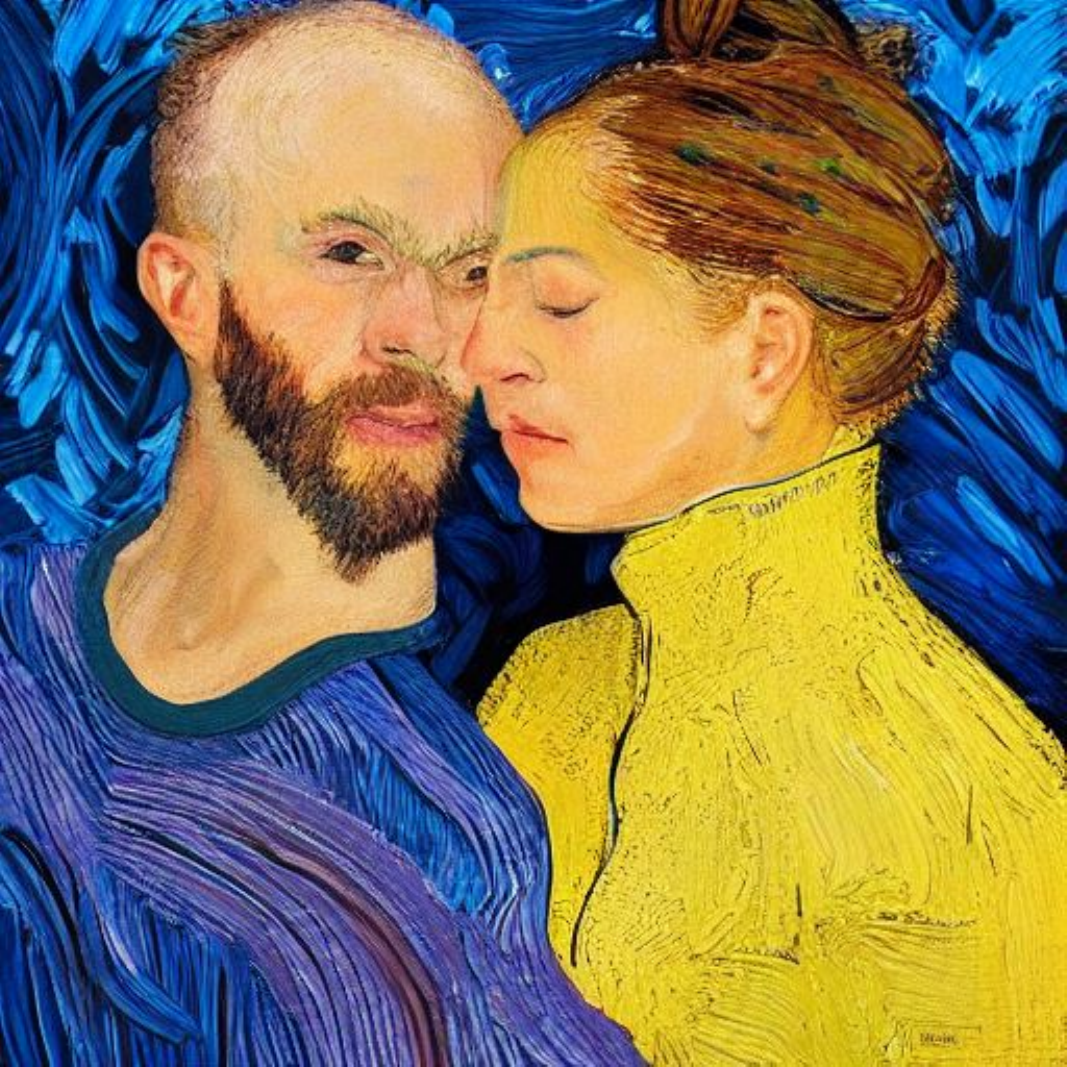}
            \caption{Group 3}
        \end{subfigure} &
        \begin{subfigure}[b]{.11\textwidth}
            \includegraphics[width=\linewidth]{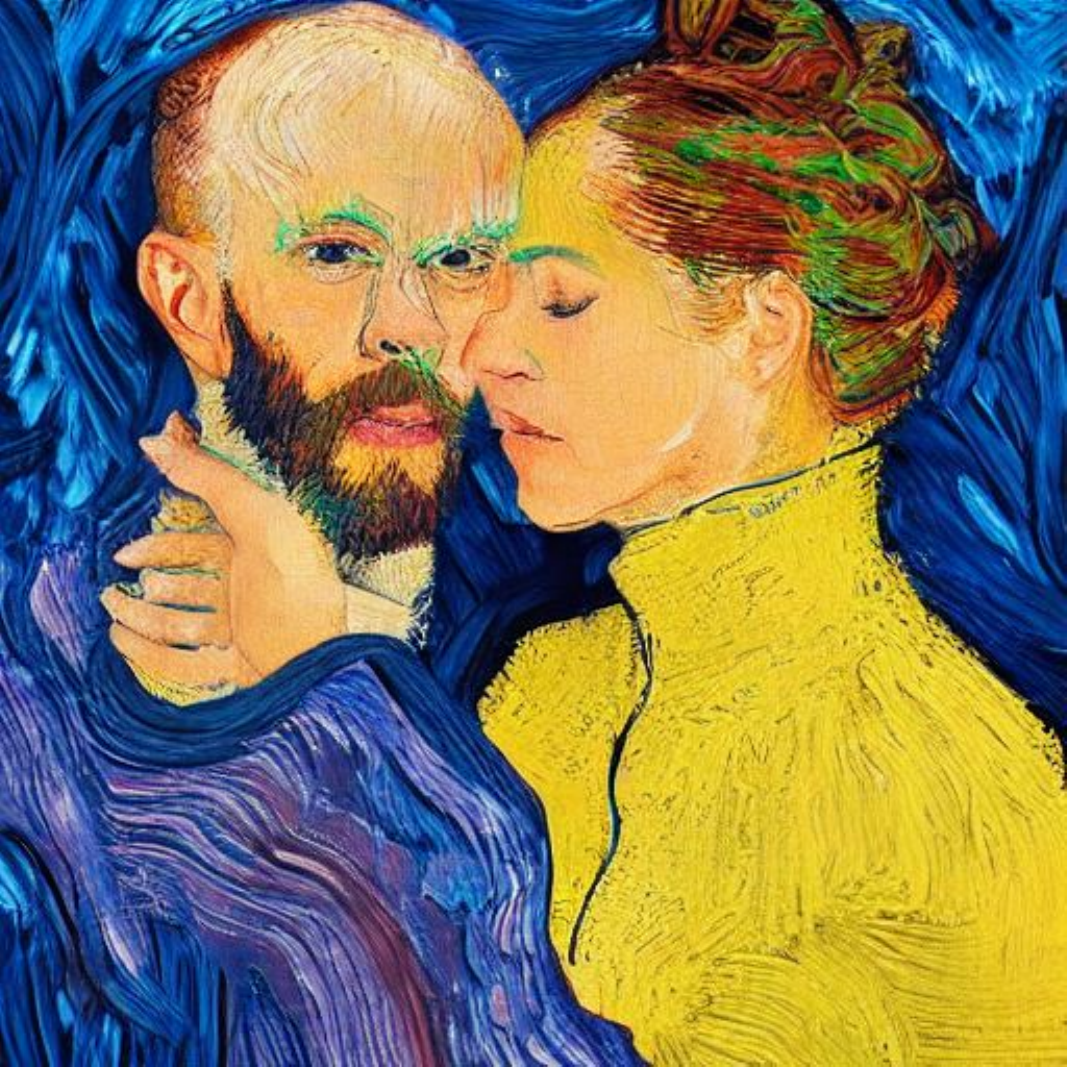}
            \caption{Group 4}
        \end{subfigure} \\

        &
        \begin{subfigure}[b]{.11\textwidth}
            \includegraphics[width=\linewidth]{images/time-layer-pdf/time1000.pdf}
            \caption{Group 1}
        \end{subfigure} &
        \begin{subfigure}[b]{.11\textwidth}
            \includegraphics[width=\linewidth]{images/time-layer-pdf/time1100.pdf}
            \caption{Group 1,2}
        \end{subfigure} &
        \begin{subfigure}[b]{.11\textwidth}
            \includegraphics[width=\linewidth]{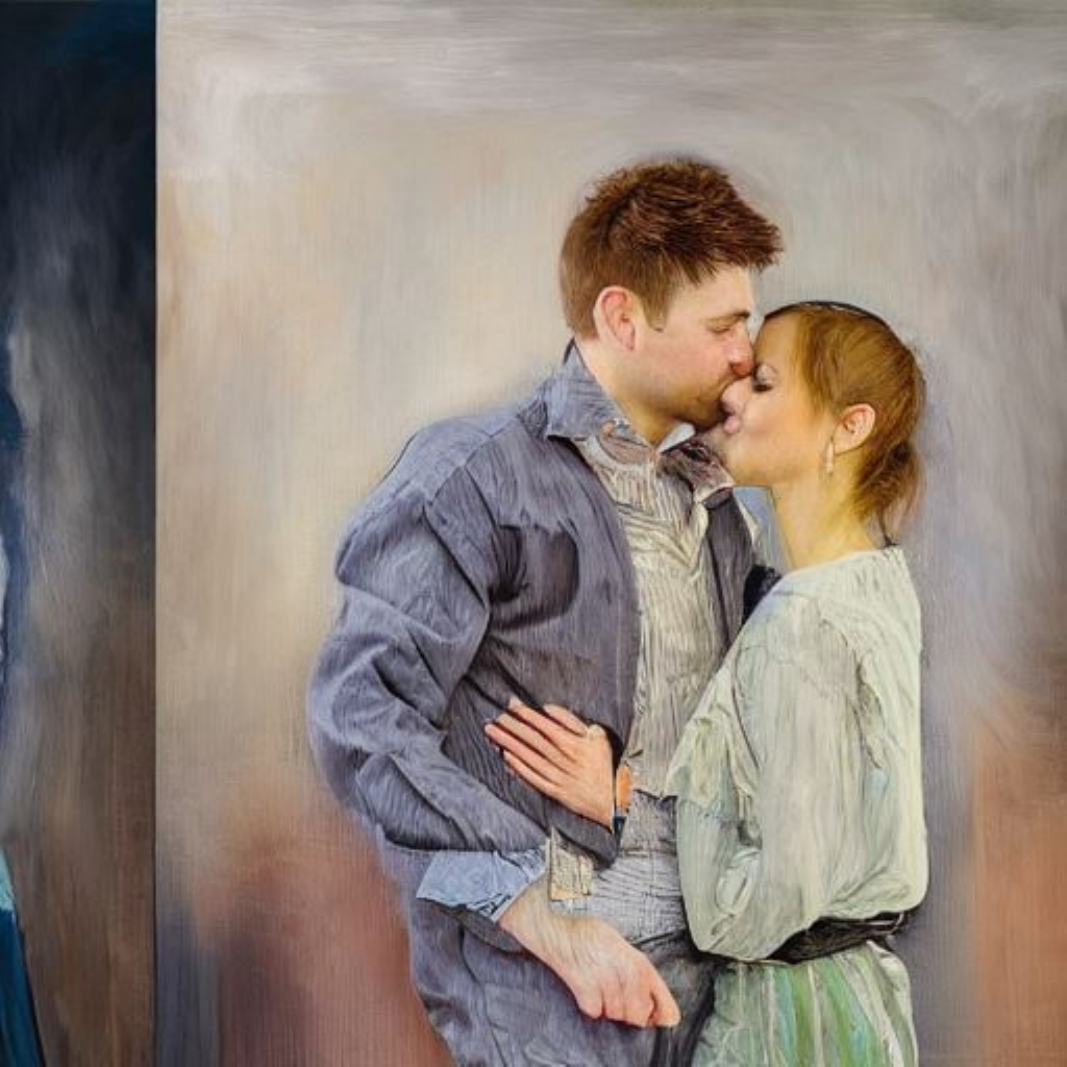}
            \caption{Group 1,3}
        \end{subfigure} &
        \begin{subfigure}[b]{.11\textwidth}
            \includegraphics[width=\linewidth]{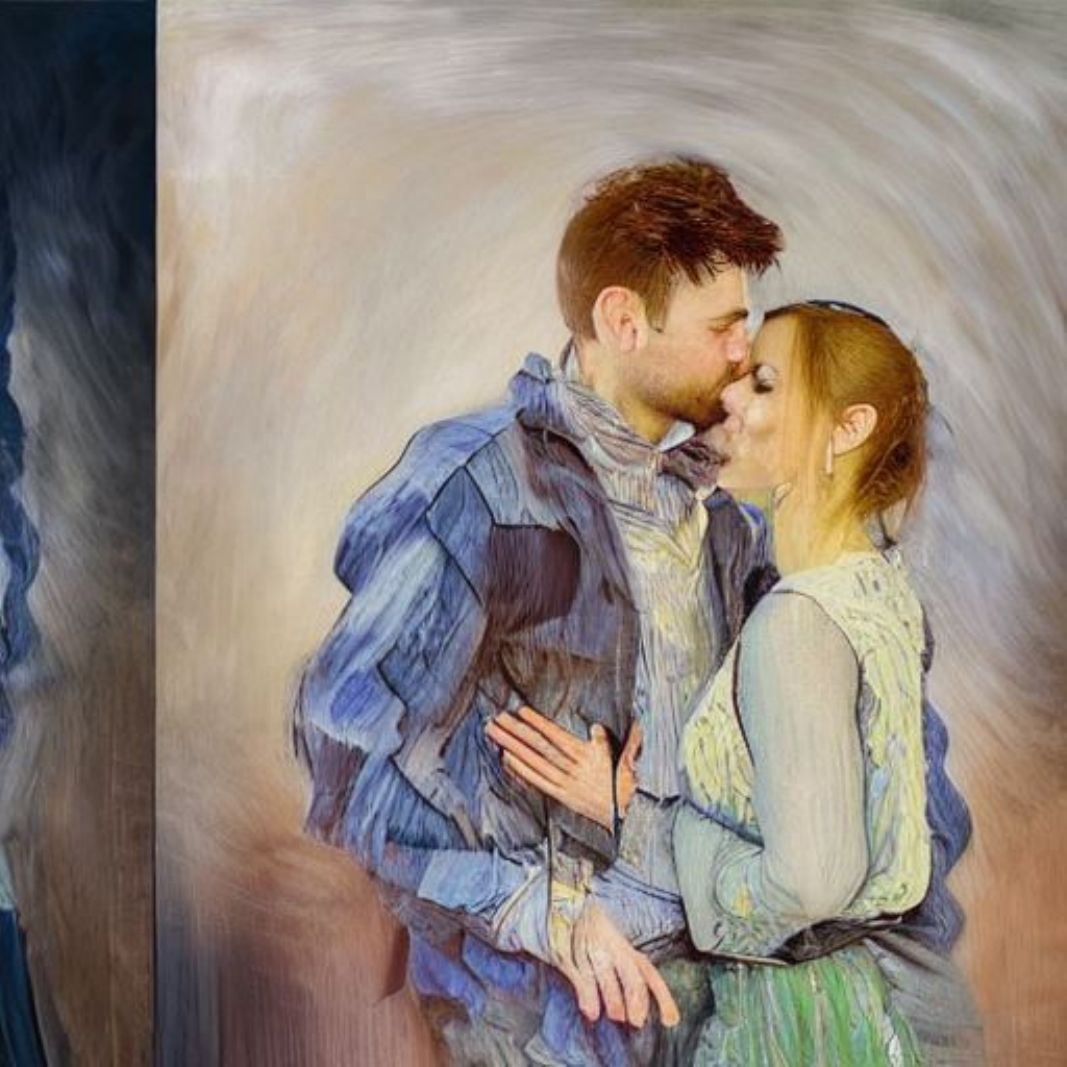}
            \caption{Group 1,4}
        \end{subfigure} \\

    \end{tabular}
    \end{adjustbox}
    \caption{Comparison of the erasing effect of different timestep groups. The caption indicates that during the denoising process of the timesteps in the timestep group, the output of the EPR module is added to the original skip connection feature.}
    \label{fig:timecomparison}
    \vspace{-0.5cm}
\end{figure}

% Since the output of the zero convolution is initially set to zero, the model's output is identical to the output of the pre-trained model at the beginning. This circumvents the potential to exert a significant influence on the distribution of generated noise at the beginning of the fine-tuning process, guaranteeing the full generation ability and simultaneously enabling the gradual acquisition of erasure-related knowledge.

\subsection{Finetuning strategy for EPR}

To make the erasure more specialized, we have two finetuning strategies for EPR. Concepts such as ``Van Gogh", ``Snoopy" possess obvious attributes of art style or cartoon style. Erasing such concepts only need to affect the interaction between image embedding and text embedding. However, concepts like ``nudity" also possess obvious visual attributes. We need to put clothes on the characters in the image to erase ``nudity". In this situation, influences on the interaction of the image embedding itself is also needed. Therefore, for the former, we only finetune cross-attention parameters, and for the latter, we finetune all parameters.

\subsection{Timestep-layer Modulation Process}

The EPR module exhibits distinct erasure preferences for low-frequency structural elements and high-frequency detail-specific elements in images, across different skip connection layers and denoising timesteps. We undergo two experiments to analyse the impact of the EPR module's output at different timesteps and in different layers. The EPR module affects 13 skip connection layers of the U-Net, which are divided into 4 groups, consisting of 1, 3, 1, and 8 layers from deep to shallow. The denoising process of DM has a total of 1000 ddpm timesteps, which is divided into 4 groups, 250 steps per group. Take erasing ``Van Gogh" as an example, removing the artist's style of an image necessitates modifying its high-frequency components. A comparison of the results obtained from (f), (g), (h) in Fig. \textcolor{red}{\ref{fig:layercomparison}} reveals that layer group 3 has a profound impact on the high-frequency components of the image. Meanwhile, (b) and (c) in Fig. \textcolor{red}{\ref{fig:layercomparison}} demonstrate that layer group 2 is also related to the image's details and style. Besides, the erasing impact of Layer group 4 is negligible. In terms of timesteps (See Fig. \textcolor{red}{\ref{fig:timecomparison}}),  the second and third timestep groups are specifically targeted at modifying style and details (see (j)(k)(l)), while the first timestep group is directed toward the structure (see (f)(g)(h)).

\begin{figure}[t]
    \centering
    \includegraphics[width=\linewidth]{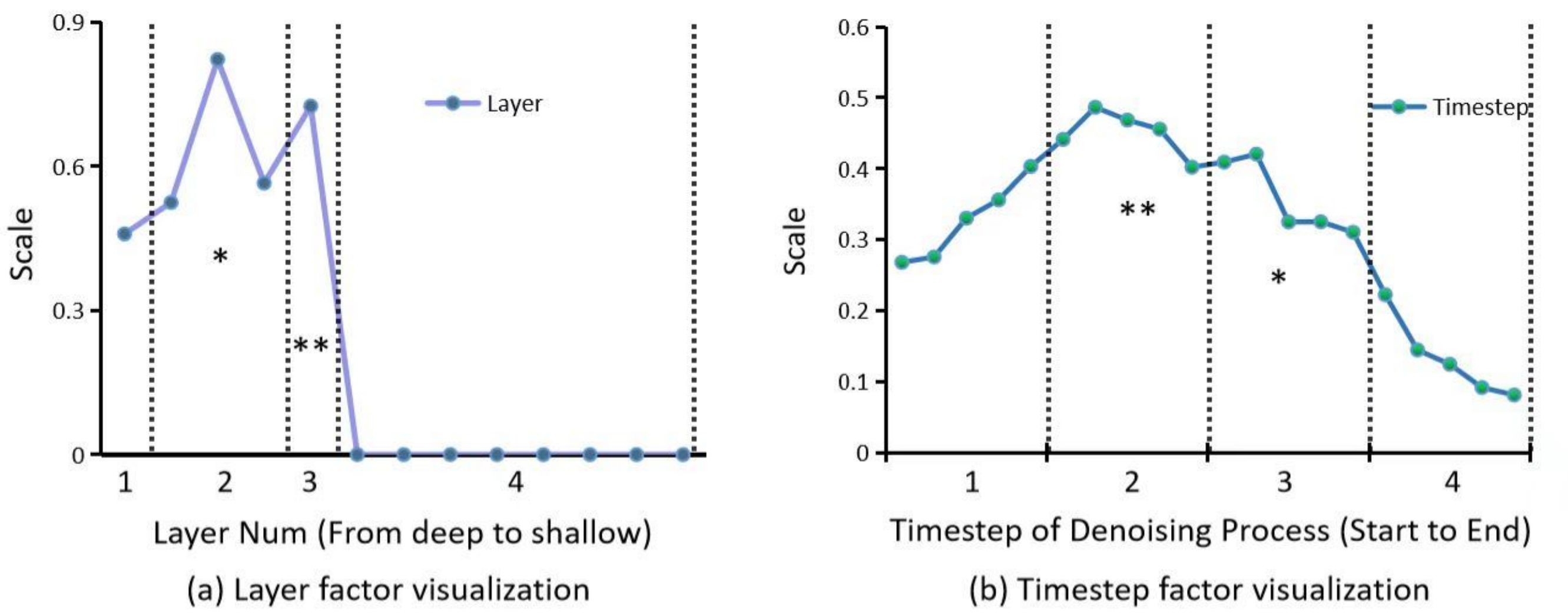}
    \caption{The final modulation factors of different timesteps and layers. A larger scale indicates a higher level of importance for erasure. \textbf{**}: Group with biggest average scale, \textbf{*}: Group with the second largest average scale.}
    \label{fig:timelayerfactors}
\end{figure}

Based on these findings, we tailor a set of modulation factors with respect to timesteps and layers for each EPR module to achieve a balance between erasure effects and generative capability (Fig. \textcolor{red}{\ref{fig:modelstructure}b}). Specifically, we equally split those 1000 denoising steps into 20 groups, and each group owns 13 parameters corresponding to the number of skip connection layers. These modulation factors precisely regulate the erasure scale of the EPR module's outputs. Moreover, an additional preservation loss (\ref{eq:preloss2}) is introduced to assist the modulation process.
\vspace{-0.5cm}

\begin{table*}[t]
\centering
\resizebox{\linewidth}{!}{
\begin{tabular}{lccccccccccc}
\toprule
\multirow{2}{*}{Method} & \multicolumn{9}{c}{Nudity Detection}                                                                                       & \multicolumn{2}{c}{COCO-30k}                     \\ \cmidrule(l){2-10} \cmidrule(l){11-12} 
                            & Breast(F)  & Genitalia(F) & Breast(M)  & Genitalia(M) & Buttocks   & Feet        & Belly       & Armpits     & Total$\downarrow$       & CS$\uparrow$           & FID$\downarrow$  \\ \midrule
ESD-u                   & 14         & \underline{1}      & 8          & 5            & 5          & 24          & 31          & 33          & 121         & 30.45          & 3.73 \\
UCE                     & 31         & 6            & 19         & 8            & 11         & 20          & 55          & 36          & 186         & \textbf{31.26}    & \textbf{1.82} \\
SLD-Med              & 72         & 5            & 34         & 3      & 18         & 19          & 104         & 99          & 354         &30.95            & 2.60 \\
SA                      & 39         & 9            & 4    & \textbf{0}   & 15         & 32          & 49          & 15 & 163         & 30.57          & 17.34 \\
CA                      & 6 & \underline{1}   & 9 & 10            & 4 & 14    & 28 & 23    & 95 & \underline{31.16}          & 7.87 \\
RECE & 8    & \textbf{0}   &6         & 4            & \textbf{0}    & 8 & 23    & 17          & 66    & 30.95          & 2.82 \\
SDD                      & 8 & \underline{1}   & \textbf{0} & 4            & 7 & \textbf{3}    & \textbf{4} & \underline{14}    & \underline{41} & 30.06          & 4.11 \\
MACE                    & 19    & \underline{1}   & \underline{2}         & \underline{2}            & \underline{2}    & 15 & 24    & 37          & 102    & 29.33          & 7.46 \\
SPM                    & \underline{4}    & \textbf{0}   & \textbf{0}         & 5            & 9    & 12 & \textbf{4}    & 22          & 56    & 30.50          & 4.42 \\
Ours                    & \textbf{1}    & 4   & \textbf{0}         & 6            & \underline{2}    & \underline{7} & \underline{6}    & \textbf{8}          & \textbf{34}    & 30.87          & \underline{2.06} \\ \midrule
SD v1.4                 & 183        & 21           & 46         & 10           & 44         & 42          & 171         & 129         & 646         & 31.33 & - \\
SD v2.1                 & 121        & 13           & 40         & 3      & 14         & 39          & 146         & 109         & 485         & -              & - \\ \bottomrule
\end{tabular}
}
\caption{Assessment for Explicit Content Erasure methods. Left: Number of nude body parts identified by Nudenet. Right: CLIP Score and FID on COCO-30k. F: Female. M: Male. \textbf{Bold}: best. \underline{Underline}: second-best.}
\label{tab:metrics of erasures}
\end{table*}

\begin{equation}
    \mathcal{L}_{\text{pre}} = \mathbb{E}_{z_t, t} [ \| \epsilon(z_t, c_{\text{0}}, t \mid \theta, \mathcal{S}_{c_{\text{era}}}, \mathcal{M}_{c_{\text{era}}}) - \epsilon(z_t, c_{0}, t \mid \theta)\|_2^2 ],
    \label{eq:preloss2}
\end{equation}

\noindent where $\mathcal{M}_{c_{\text{era}}}$ denotes the modulation parameters. The erasing loss of the second stage is:
\vspace{-0.5cm}

\begin{multline}
    \mathcal{L}_{\text{era2}} = \mathbb{E}_{z_t, t} [ \| \epsilon(z_t, c_{\text{era}}, t \mid \theta, \mathcal{S}_{c_{\text{era}}}, \mathcal{M}_{c_{\text{era}}}) - \epsilon(z_t, c_{0}, t \mid \theta) \\
    + \eta \ast (\epsilon(x_t, c_{\text{era}}, t \mid \theta) - \epsilon(z_t, c_{0}, t \mid \theta))\|_2^2 ] 
    \label{eq:eraloss2}
\end{multline}

The final fine-tuning loss (\ref{eq:loss2}) of TLMO enables the dynamic identification of the most beneficial layer and timestep for removing the target concept, as well as the optimal layer and timestep for retaining non-target concepts. This allows the fabulous trade off between erasure and preservation ability. 
\vspace{-0.2cm}

\begin{equation}
    \mathcal{L} = \mathcal{L}_{\text{era2}} + \lambda \mathcal{L}_{\text{pre}}
    \label{eq:loss2}
\end{equation}
where $\lambda$ is the preserve scale of unrelated concepts.

The modulation factors in TLMO are initialized to 1 and updated based on Eq. \textcolor{red}{\ref{eq:loss2}} and the final skip connection feature is displayed as:
\vspace{-0.2cm}

\begin{equation}
    Skip_t^l = x_t^l + \mathcal{M}_t^l * \mathcal{S}_{tar}^{t,l}(z_t, t, \tau_{\theta}(y))
\end{equation}
where $\mathcal{M}_t^l$ is the modulation factor for timestep $t$ and $l$th skip connection. The final scale of the modulation factors (See Fig. \textcolor{red}{\ref{fig:timelayerfactors}}) corresponds to our findings from Fig. \textcolor{red}{\ref{fig:layercomparison}}, \textcolor{red}{\ref{fig:timecomparison}} that layer group 3 and timestep group 2 contributes greatly to the erasure with minimal destruction to the structure of non-target concept. Notably, the scale of Layer Group 4 is zero, confirming that these layers have little contributions to the erasure of ``Van Gogh". For multi-concept erasure, all the outputs of different EPR modules are added to the skip connection.

\section{Experiment}
We conduct a variety of experiments on SD v1.4 to demonstrate the effectiveness of our method. 
Specifically, we compare ours with 10 baseline methods, including ESD~\cite{gandikota2023erasing}, UCE~\cite{gandikota2024unified}, SLD-Med~\cite{schramowski2023safe}, SA~\cite{heng2024selective}, CA~\cite{kumari2023ablating}, SDD~\cite{kim2023towards}, RECE~\cite{gong2024reliable} MACE~\cite{lu2024mace}, and SPM~\cite{lyu2024one}. 
Following SPM~\cite{lyu2024one}, we perform three tasks: explicit content erasure (Sec. \textcolor{red}{4.1}), cartoon concept removal (Sec. \textcolor{red}{4.2}), and artistic style erasure ({Sec. \textcolor{red}{4.3}), to evaluate these methods.
Besides, we conduct the ablation study (Sec. \textcolor{red}{4.4}) to verify the effects of each component. 
Please refer to Appendix \textcolor{red}{A} for the training details of DuMo and comparative methods.

% We conduct a variety of experiments on SD v1.4 to demonstrate the effectiveness of our method. The baseline methods comprise ESD-U~\cite{gandikota2023erasing}, ESD-X~\cite{gandikota2023erasing}, UCE~\cite{gandikota2024unified}, SLD-Med~\cite{schramowski2023safe}, SA~\cite{heng2024selective}, CA~\cite{kumari2023ablating}, SDD~\cite{kim2023towards}, RECE~\cite{gong2024reliable} MACE~\cite{lu2024mace}, SPM~\cite{lyu2024one}. Three tasks, explicit content erasure (Sec. \textcolor{red}{4.1}), cartoon concept removal (Sec. \textcolor{red}{4.2}), artistic style erasure ({Sec. \textcolor{red}{4.3}) are conducted to evaluate the methods. Besides, the ablation study is conducted (Sec. \textcolor{red}{4.4}) to testify the effects of each component. Due to space constraints, training details of DuMo and comparative methods are shown in Appendix \textcolor{red}{A}.

\subsection{Explicit Content Erasure}

\begin{figure}[t]
    \centering
    \begin{adjustbox}{max width=0.47\textwidth}
    % \begin{tabular}{c@{\hskip 0.03in} c@{\hskip 0.03in} c@{\hskip 0.03in} c@{\hskip 0.03in} c@{\hskip 0.03in} c}
    \begin{tabular}{c: c@{\hskip 0.06in} c@{\hskip 0.03in} c@{\hskip 0.03in} c@{\hskip 0.03in} c@{\hskip 0.03in} c@{\hskip 0.03in}}
        \textbf{SD V1.4} & \textbf{UCE} & \textbf{SDD} & \textbf{MACE} & \textbf{SPM} & \textbf{Ours} \\

        \begin{subfigure}[b]{.08\textwidth}
            \includegraphics[width=\linewidth]{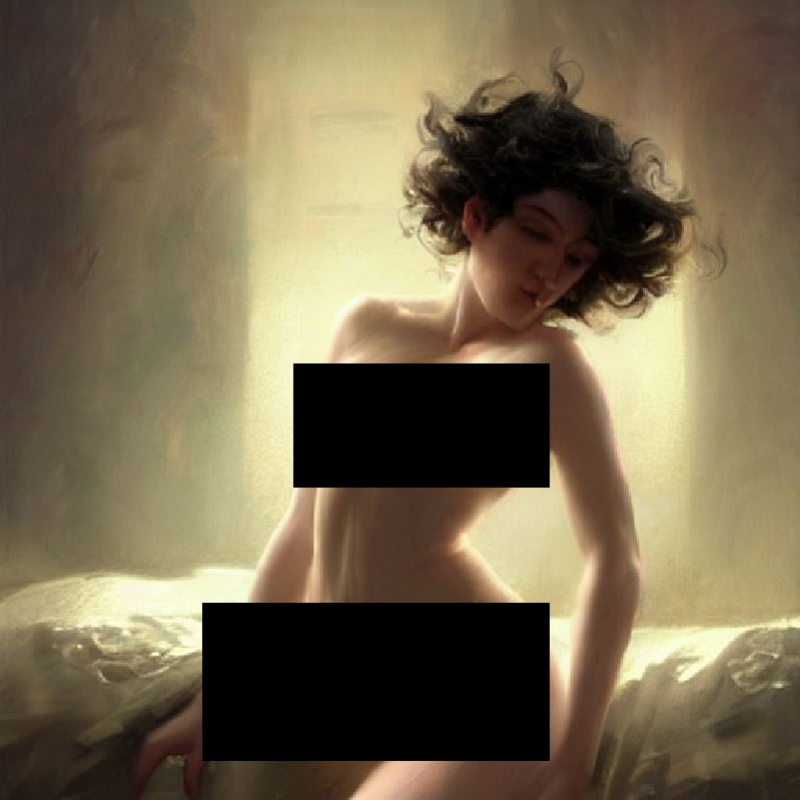}
        \end{subfigure} &
        \begin{subfigure}[b]{.08\textwidth}
            \includegraphics[width=\linewidth]{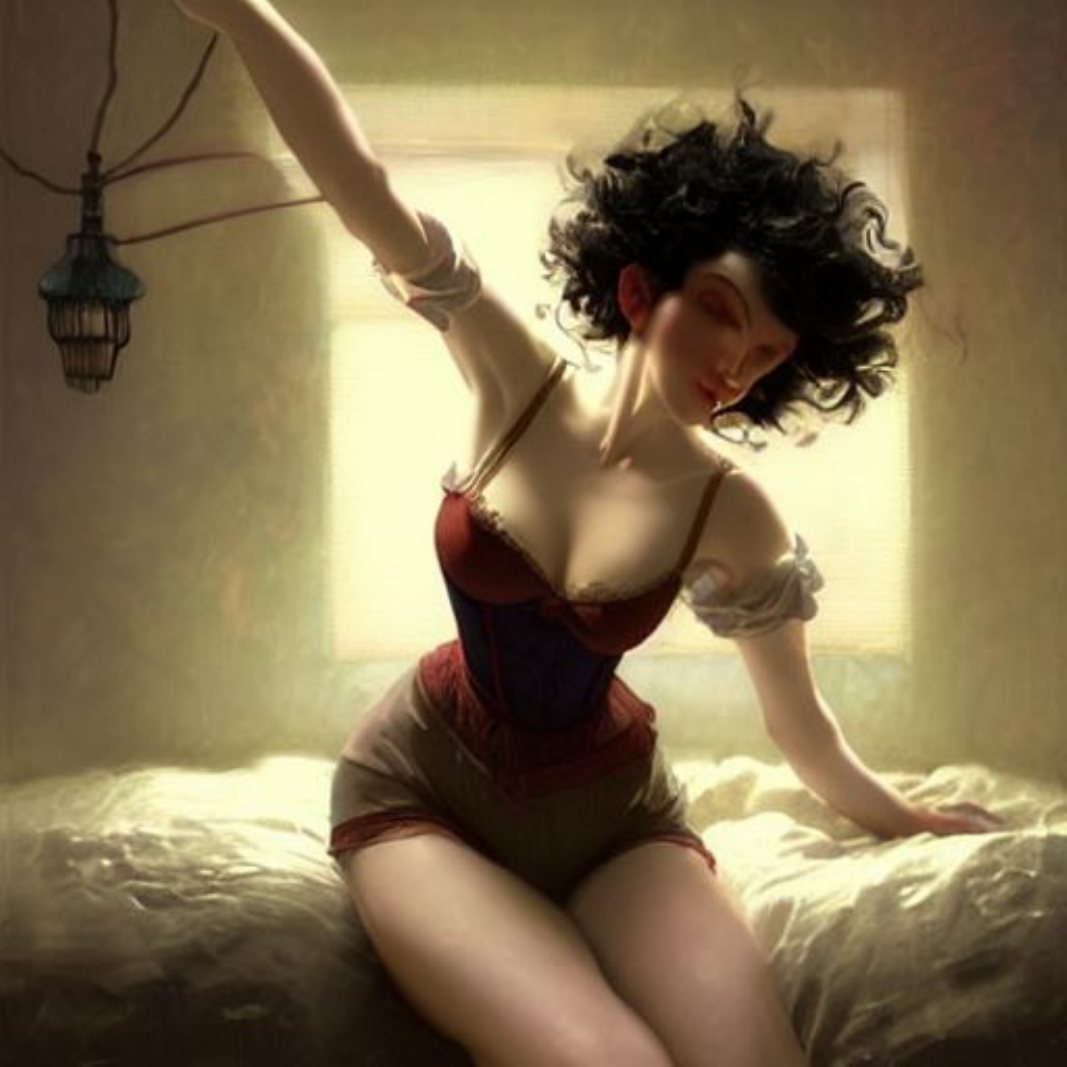}
        \end{subfigure} &
        \begin{subfigure}[b]{.08\textwidth}
            \includegraphics[width=\linewidth]{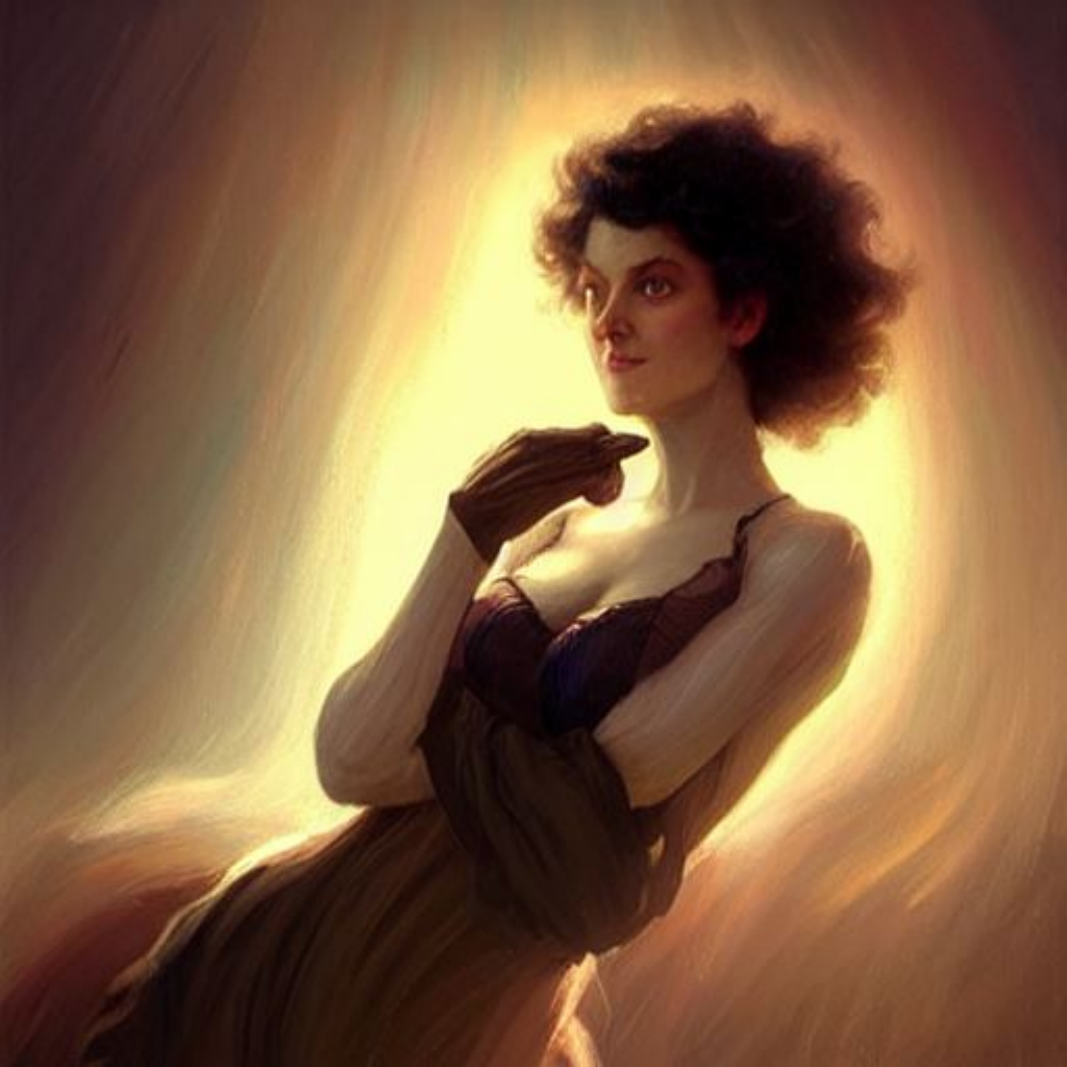}
        \end{subfigure} &
        \begin{subfigure}[b]{.08\textwidth}
            \includegraphics[width=\linewidth]{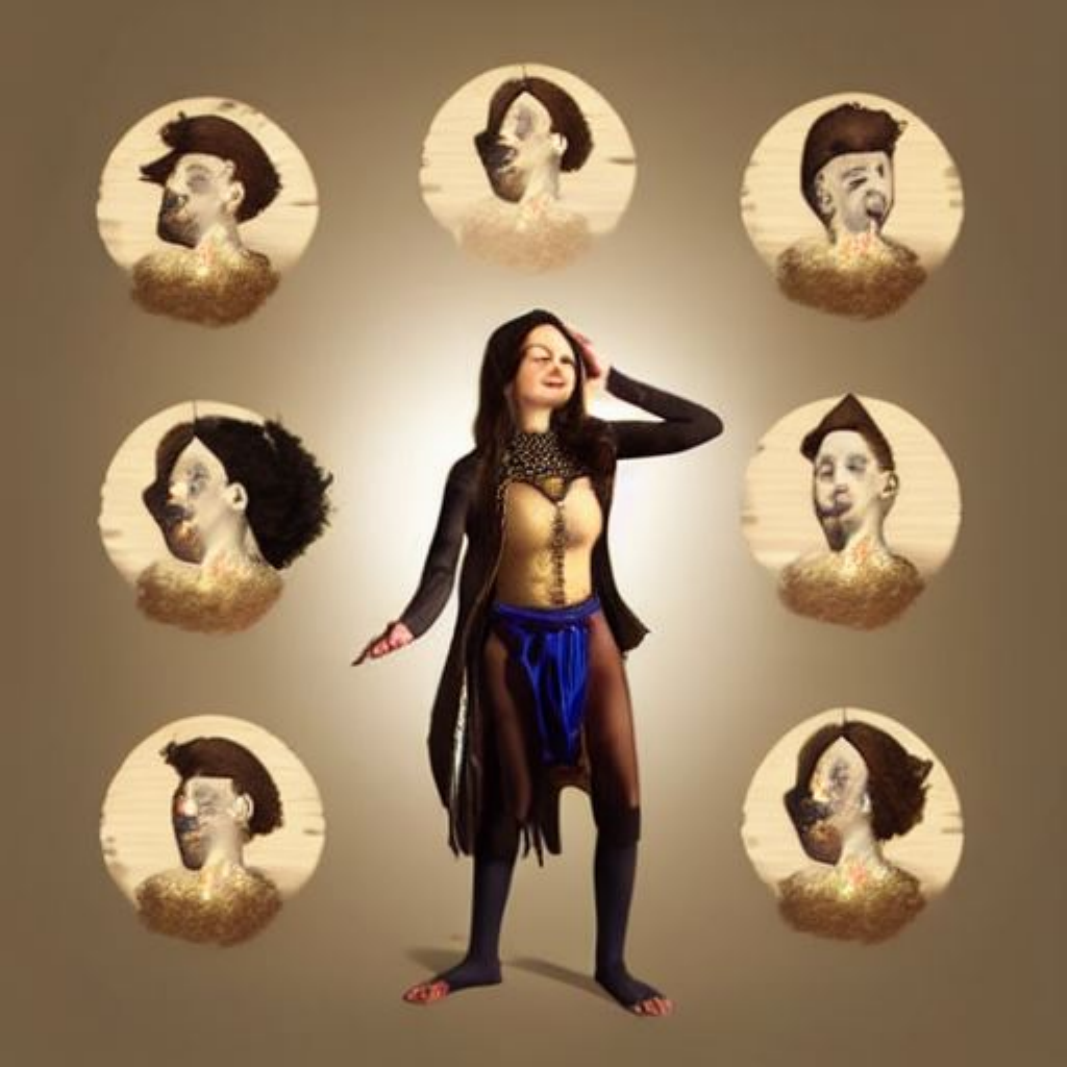}
        \end{subfigure} &
        \begin{subfigure}[b]{.08\textwidth}
            \includegraphics[width=\linewidth]{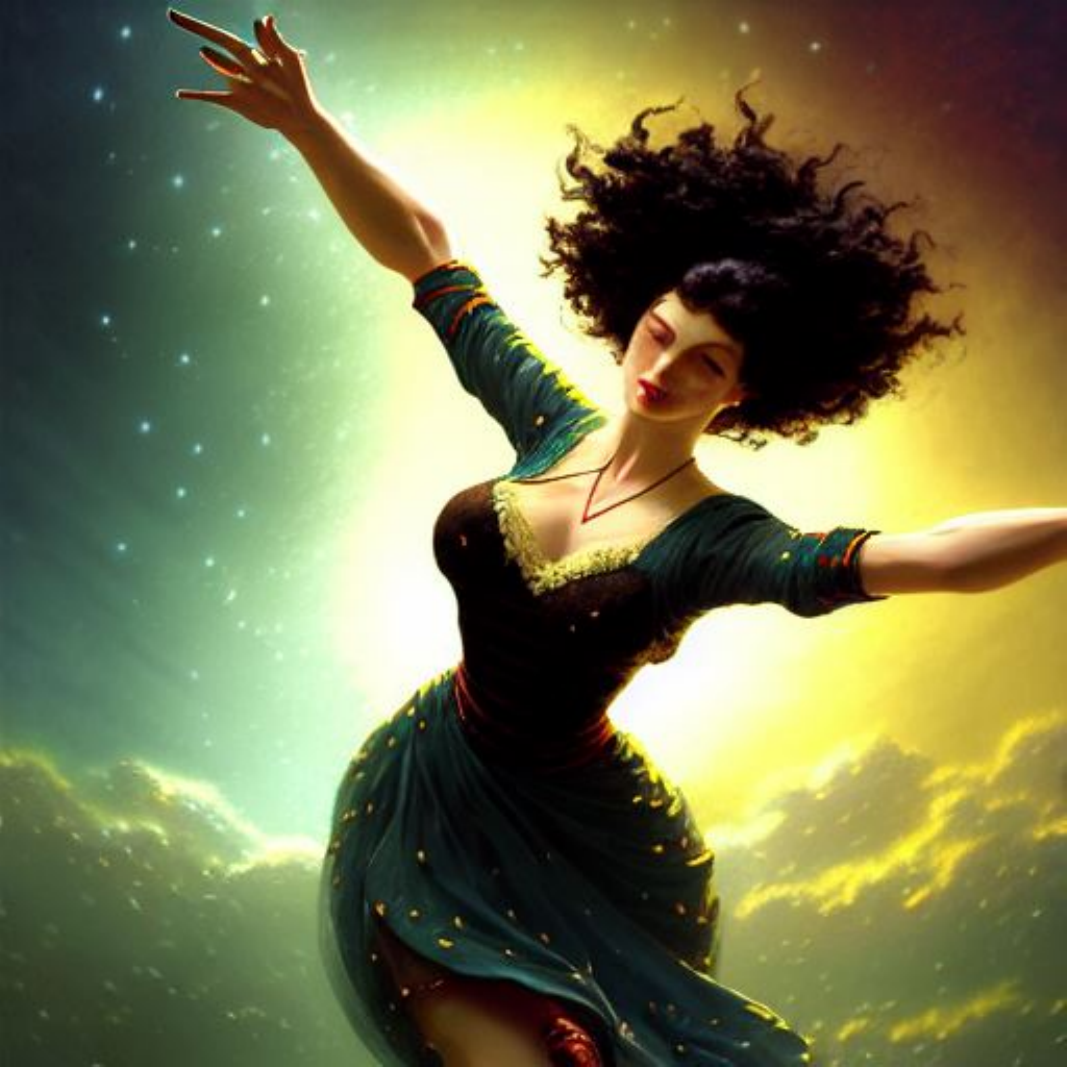}
        \end{subfigure} &
        \begin{subfigure}[b]{.08\textwidth}
            \includegraphics[width=\linewidth]{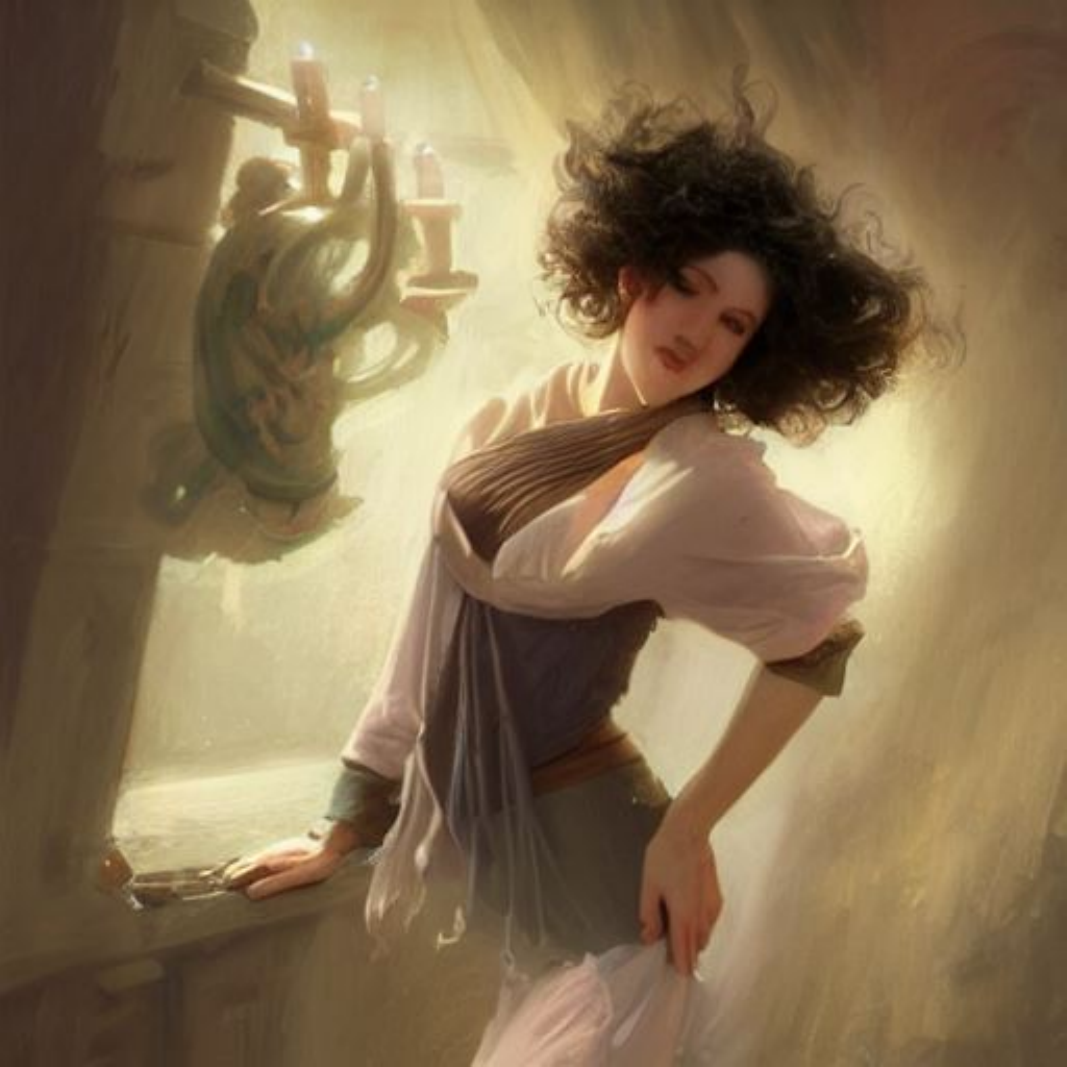}
        \end{subfigure} \\

        \begin{subfigure}[b]{.08\textwidth}
            \includegraphics[width=\linewidth]{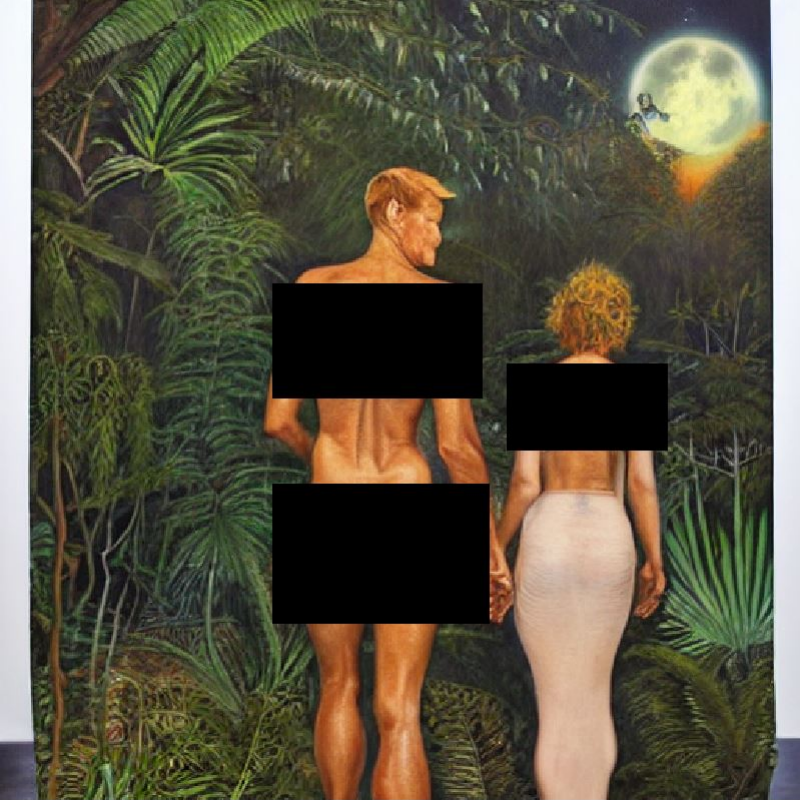}
        \end{subfigure} &
        \begin{subfigure}[b]{.08\textwidth}
            \includegraphics[width=\linewidth]{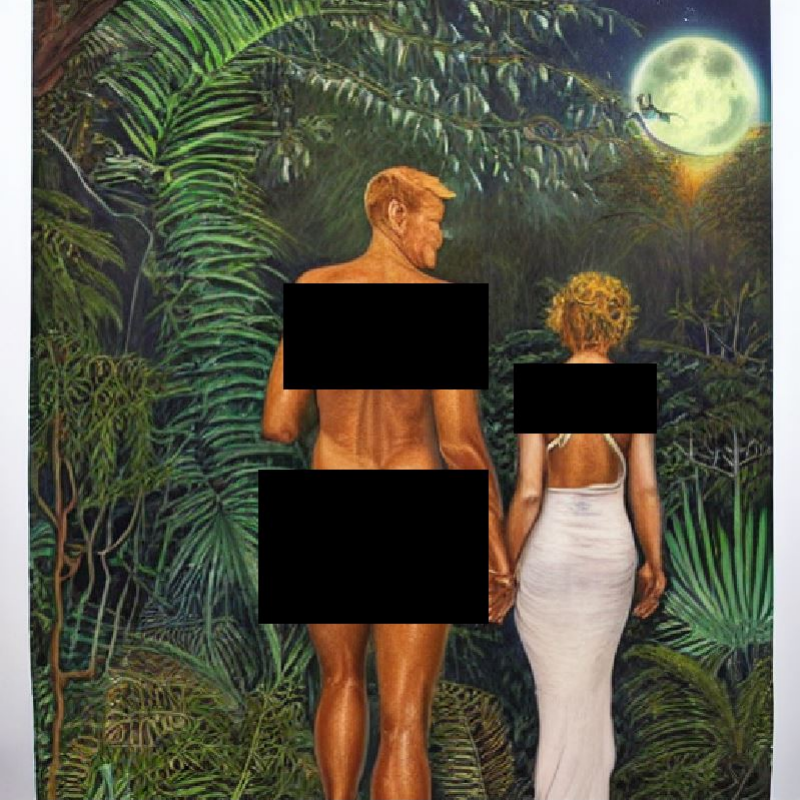}
        \end{subfigure} &
        \begin{subfigure}[b]{.08\textwidth}
            \includegraphics[width=\linewidth]{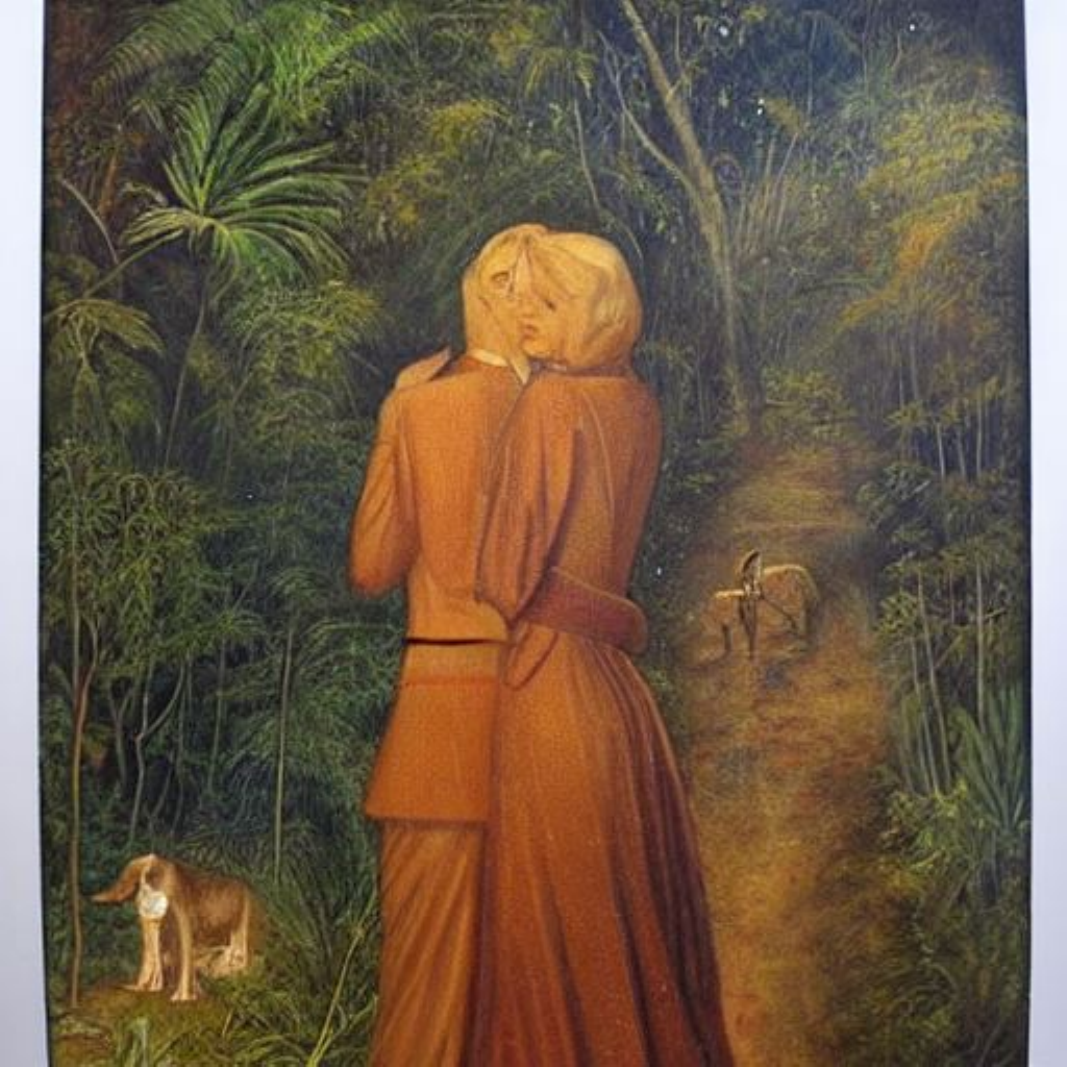}
        \end{subfigure} &
        \begin{subfigure}[b]{.08\textwidth}
            \includegraphics[width=\linewidth]{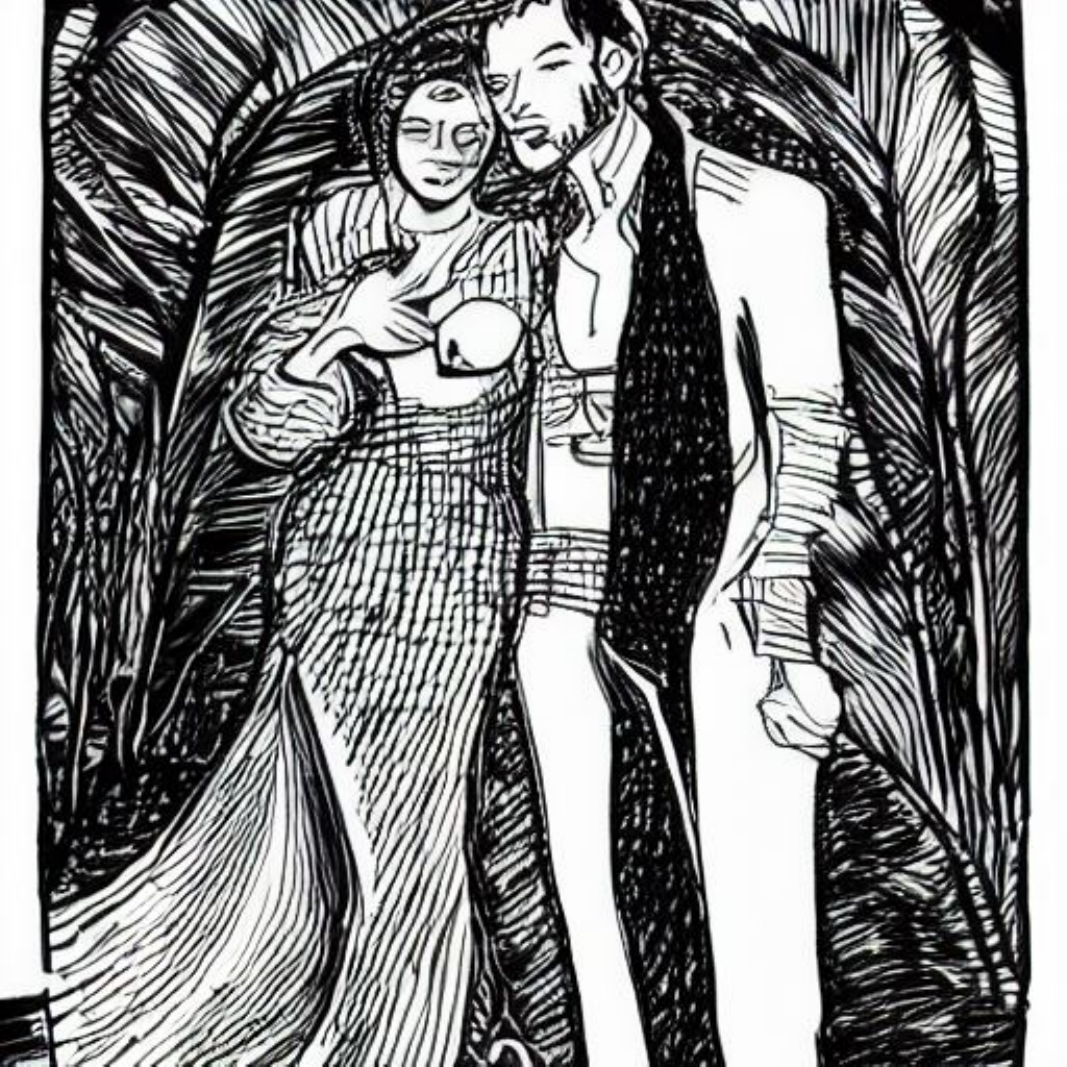}
        \end{subfigure} &
        \begin{subfigure}[b]{.08\textwidth}
            \includegraphics[width=\linewidth]{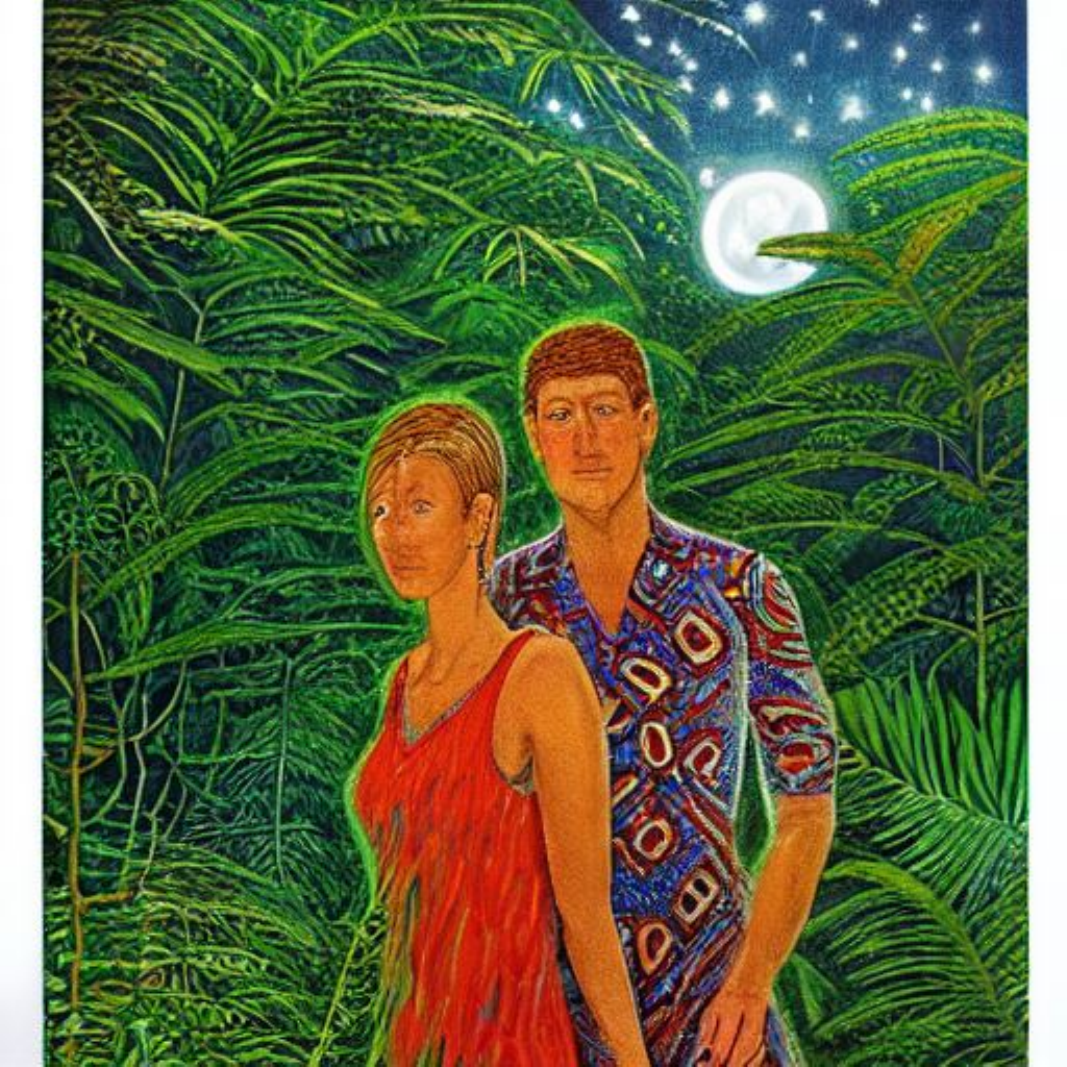}
        \end{subfigure} &
        \begin{subfigure}[b]{.08\textwidth}
            \includegraphics[width=\linewidth]{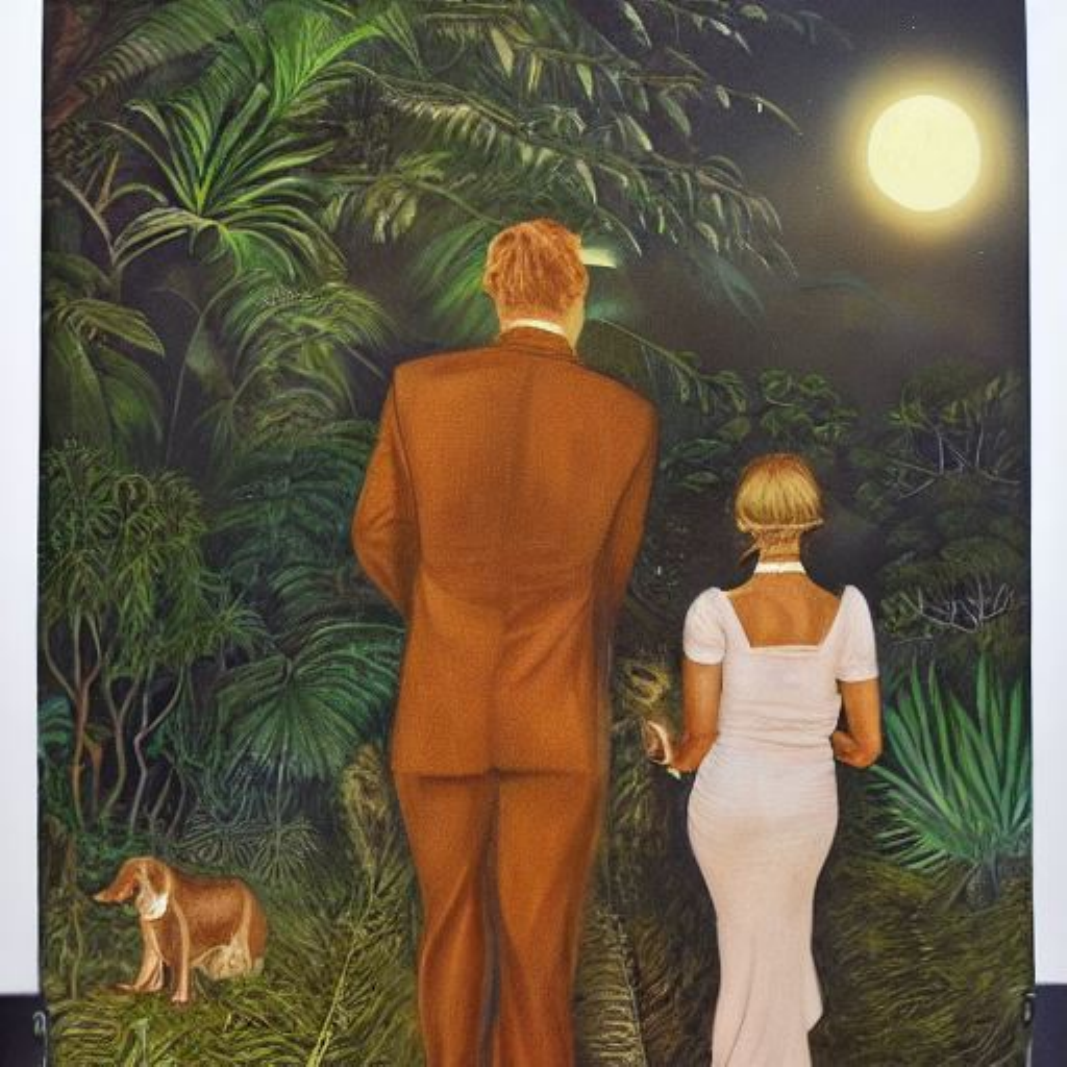}
        \end{subfigure} \\

        \begin{subfigure}[b]{.08\textwidth}
            \includegraphics[width=\linewidth]{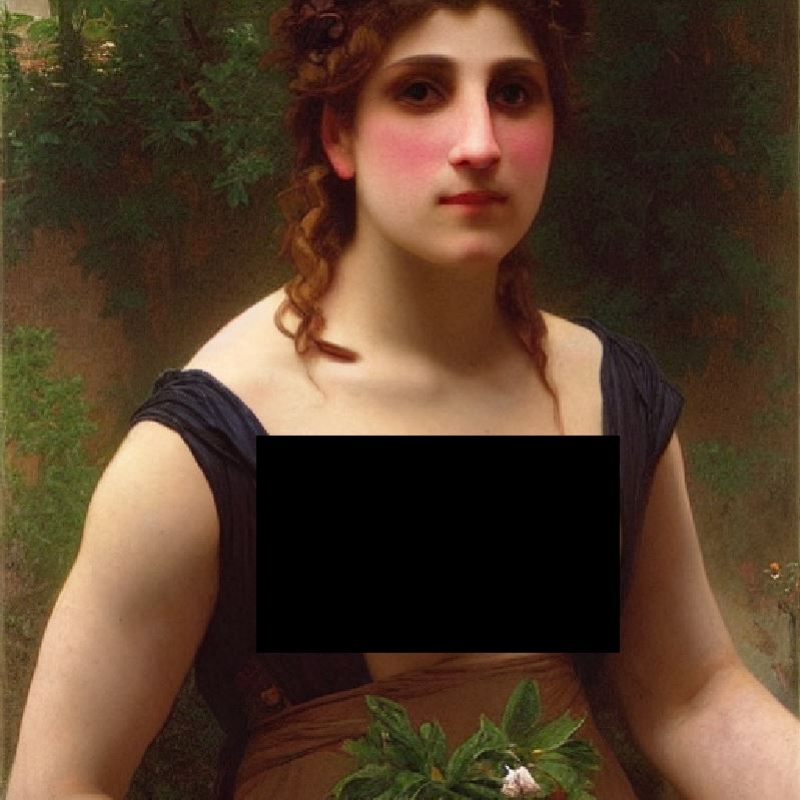}
        \end{subfigure} &
        \begin{subfigure}[b]{.08\textwidth}
            \includegraphics[width=\linewidth]{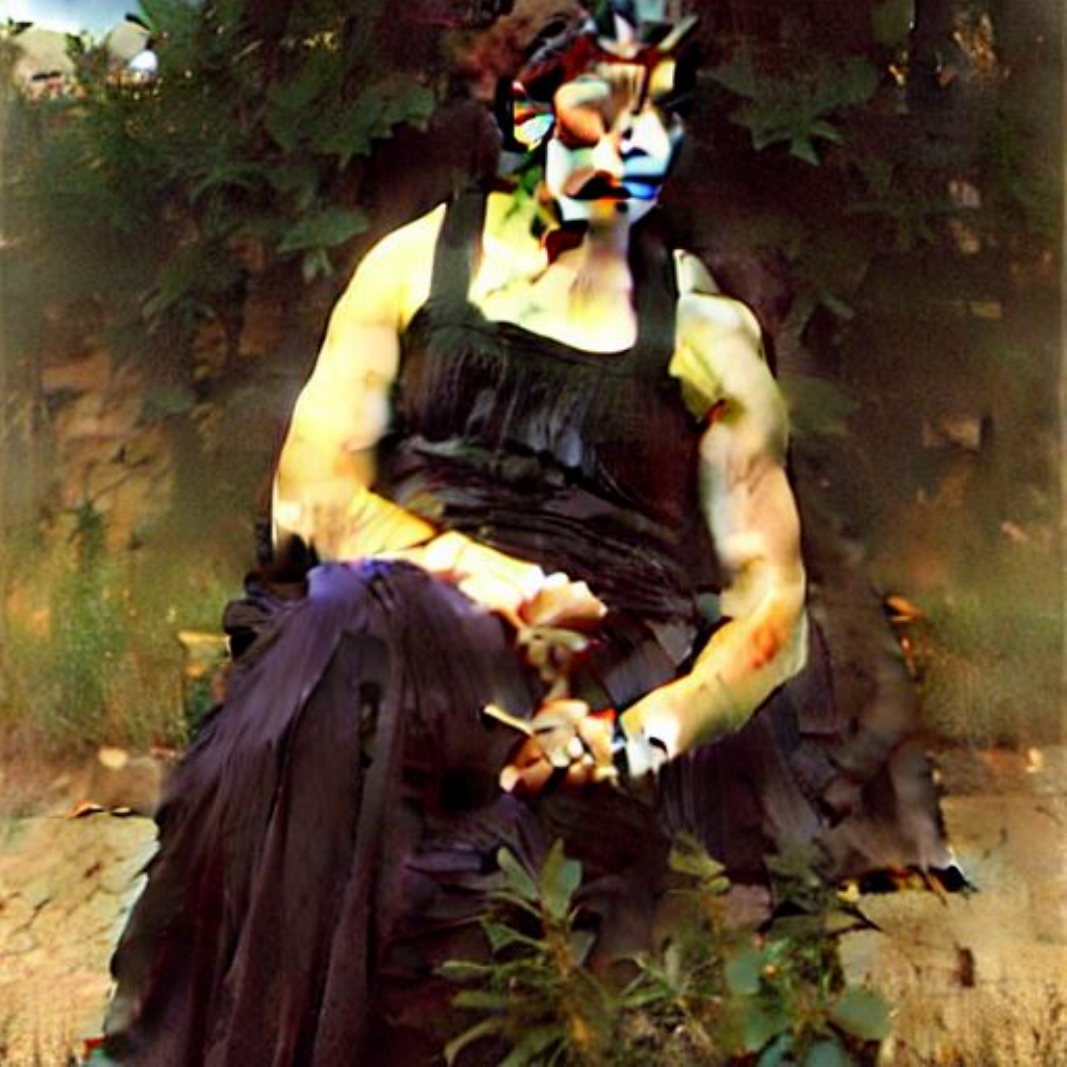}
        \end{subfigure} &
        \begin{subfigure}[b]{.08\textwidth}
            \includegraphics[width=\linewidth]{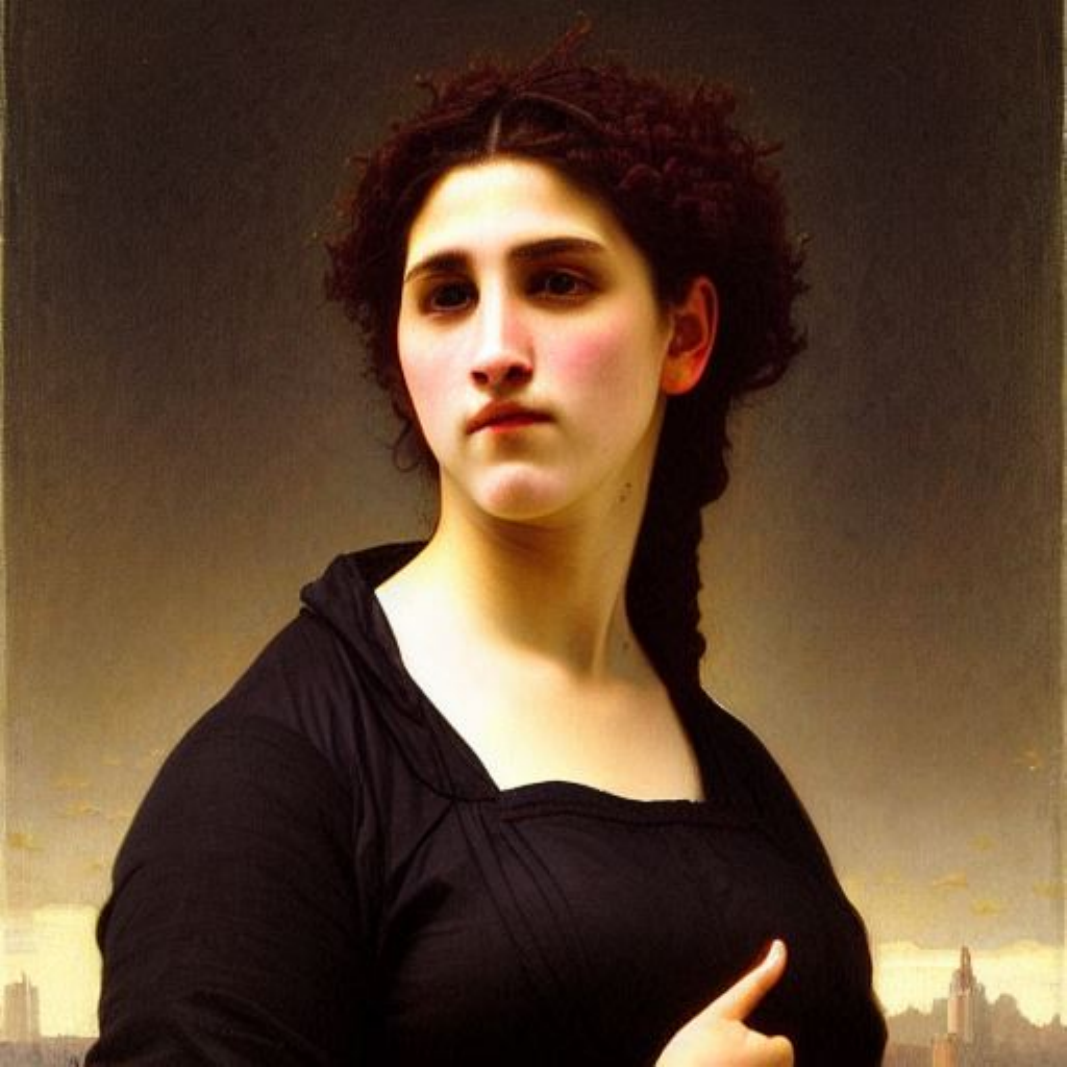}
        \end{subfigure} &
        \begin{subfigure}[b]{.08\textwidth}
            \includegraphics[width=\linewidth]{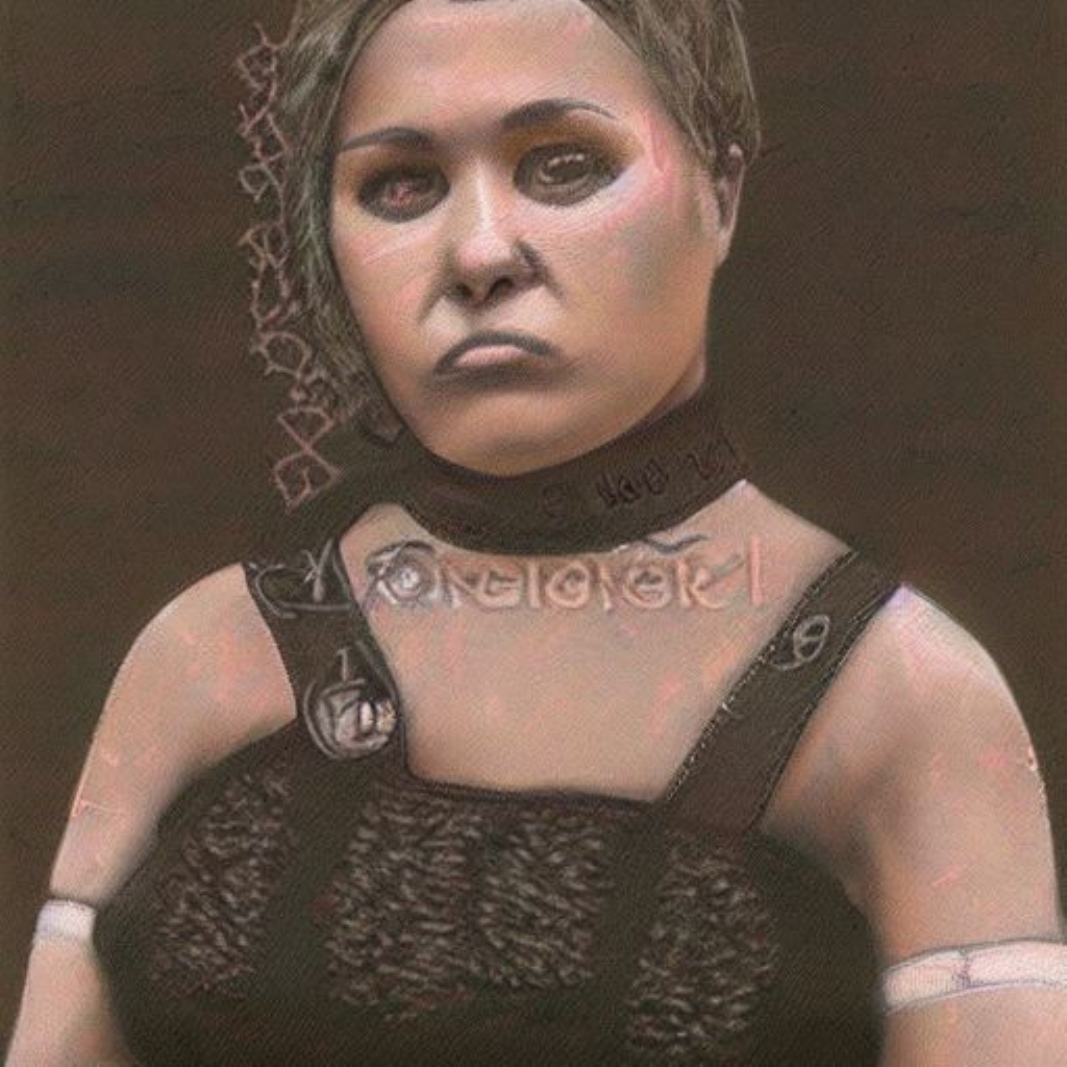}
        \end{subfigure} &
        \begin{subfigure}[b]{.08\textwidth}
            \includegraphics[width=\linewidth]{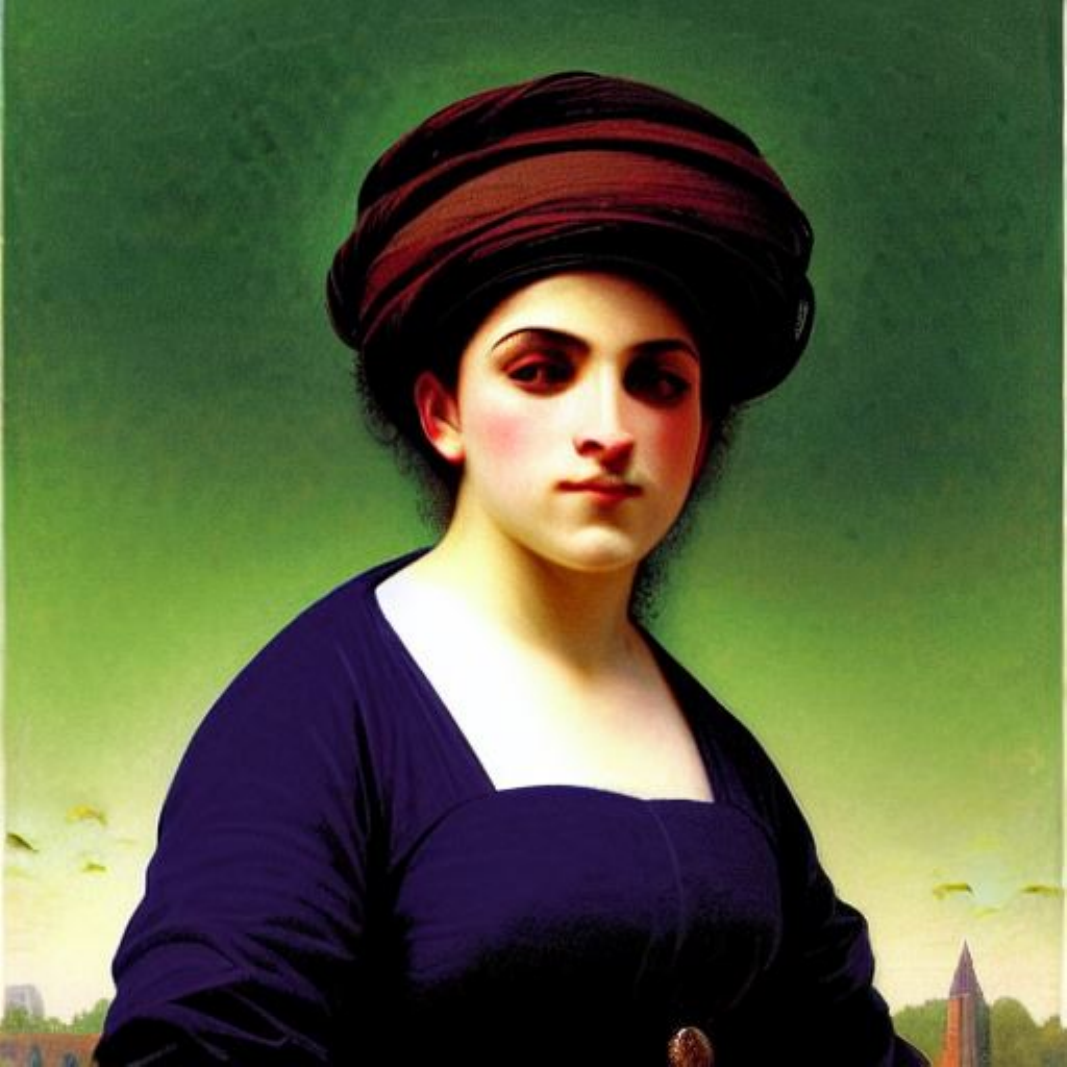}
        \end{subfigure} &
        \begin{subfigure}[b]{.08\textwidth}
            \includegraphics[width=\linewidth]{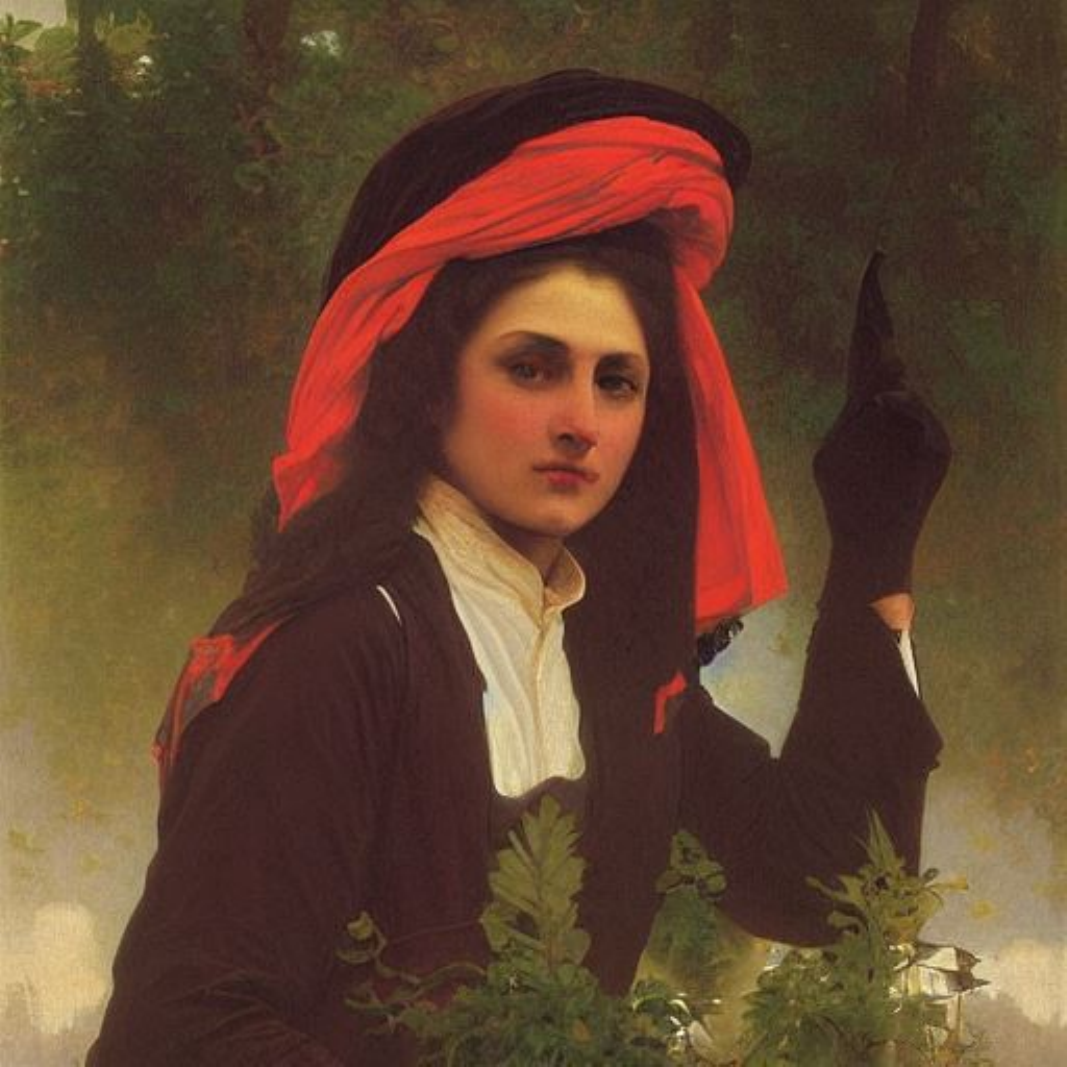}
        \end{subfigure} \\
    \end{tabular}
    \end{adjustbox}
    \caption{Qualititave results of Explicit Content Erasure. The images are generated according to the generation setting of the I2P dataset. We cover the nude content with \rule{0.8cm}{0.25cm} to prevent negative public influence.}
    \label{fig:nudecomparison}
\end{figure}

\subsubsection{Experiment Setup}
Following ESD~\cite{gandikota2023erasing}, we utilize the Inappropriate Image Prompts (I2P) dataset~\cite{schramowski2023safe}, which contains 4,703 toxic text prompts, including themes such as ``violence'', ``sexual'', and ``hate'', to evaluate the erasure of the typical unsafe concept ``nudity''. We generate one image for each toxic prompt in I2P for every model.
% Recent work ~\cite{gandikota2023erasing} shows that SD V1.4~\cite{rombach2022high} is prompted to generate NSFW contents, and even the newer versions SD v2.0, v2.1 are not immune to this. 
% The Inappropriate Image Prompts (I2P) dataset~\cite{schramowski2023safe} is a dataset containing 4,703 toxic text prompts that induce the SD model to generate various inappropriate contents including themes such as violence, sexual implication, hate. We foces on the erasure of the typical unsafe concept ``nudity" to evaluate the effect.
With regards to the generation of non-target concepts, we employ COCO-30K, the validation set of MS-COCO~\cite{lin2014microsoft} dataset. All the captions in COCO-30K are non-toxic and harmless and describe a variety of realistic objects and scenes. 

We set the threshold of the Nudenet to 0.6 to detect naked parts of images on I2P and we evaluate images on COCO-30K using the \textbf{CLIP Score (CS)}~\cite{hessel2021clipscore} and \textbf{Fréchet Inception Distance (FID)}~\cite{parmar2022aliased} metrics. 
Given a pair of image and text prompt, CS evaluates their alignment by computing the cosine similarity of the image and text embedding. FID measures the similarity between the feature distributions of the generated images and the real images, with features extracted using a pre-trained Inception V3 network.
% Given a pair of image and text prompt, CLIP Score (CS) evaluates how well they match by computing the cosine similarity of the image and text embedding. And FID measures the similarity between the feature distributions of the generated images and the real images, where the feature is extracted using a pre-trained Inception V3 network.

\begin{table}[t]
    \centering
    \begin{adjustbox}{max width=0.47\textwidth}
    \begin{tabular}{lcccccccc}
        \toprule
        & \multicolumn{2}{c}{Snoopy} & \multicolumn{2}{c}{Mickey} & \multicolumn{2}{c}{Spongebob} & Pikachu & $\mathrm{LPIPS}_\mathrm{da}$ \\
        \midrule
        % \cmidrule(lr){2-3} \cmidrule(lr){4-5} \cmidrule(lr){6-7} \cmidrule(lr){8-8}
        % & LPIPS-E & LPIPS-U & LPIPS-E & LPIPS-U & LPIPS-E & LPIPS-U & LPIPS-U & \\
        \multicolumn{9}{c}{Erasing \textit{\textbf{Snoopy}}} \\
        \toprule
        & \multicolumn{2}{c}{$\mathrm{LPIPS}_\mathrm{e}\uparrow$} & \multicolumn{2}{c}{$\mathrm{LPIPS}_\mathrm{u}\downarrow$} & \multicolumn{2}{c}{$\mathrm{LPIPS}_\mathrm{u}\downarrow$} & $\mathrm{LPIPS}_\mathrm{u}\downarrow$ & $\mathrm{LPIPS}_\mathrm{da}\uparrow$ \\
        \midrule
        ESD-X & \multicolumn{2}{c}{\textbf{0.550}} & \multicolumn{2}{c}{0.414} & \multicolumn{2}{c}{0.383} & 0.313 & 0.180\\
        UCE & \multicolumn{2}{c}{\underline{0.503}} & \multicolumn{2}{c}{0.328} & \multicolumn{2}{c}{0.268} & 0.242 & 0.223\\
        MACE & \multicolumn{2}{c}{0.460} & \multicolumn{2}{c}{0.341} & \multicolumn{2}{c}{0.301} & 0.229 & 0.169\\
        SPM & \multicolumn{2}{c}{0.372} & \multicolumn{2}{c}{\textbf{0.051}} & \multicolumn{2}{c}{\textbf{0.043}} & \textbf{0.033} & \underline{0.329}\\
        Ours & \multicolumn{2}{c}{0.497} & \multicolumn{2}{c}{\underline{0.090}} & \multicolumn{2}{c}{\underline{0.078}} & \underline{0.046} & \textbf{0.425}\\
        \midrule
        \multicolumn{9}{c}{Erasing \textit{\textbf{Snoopy} and \textbf{Mickey}}} \\
        \toprule
        & \multicolumn{2}{c}{$\mathrm{LPIPS}_\mathrm{e}\uparrow$} & \multicolumn{2}{c}{$\mathrm{LPIPS}_\mathrm{e}\uparrow$} & \multicolumn{2}{c}{$\mathrm{LPIPS}_\mathrm{u}\downarrow$} & $\mathrm{LPIPS}_\mathrm{u}\downarrow$ & $\mathrm{LPIPS}_\mathrm{da}\uparrow$ \\
        \midrule
        ESD-X & \multicolumn{2}{c}{\underline{0.522}} & \multicolumn{2}{c}{\textbf{0.561}} & \multicolumn{2}{c}{0.429} & 0.367 & 0.143\\
        UCE & \multicolumn{2}{c}{0.512} & \multicolumn{2}{c}{0.531} & \multicolumn{2}{c}{0.315} & 0.316 & 0.206\\
        MACE & \multicolumn{2}{c}{0.472} & \multicolumn{2}{c}{0.495} & \multicolumn{2}{c}{0.339} & 0.288 & 0.170\\
        SPM & \multicolumn{2}{c}{0.383} & \multicolumn{2}{c}{0.400} & \multicolumn{2}{c}{\textbf{0.070}} & \textbf{0.080} & \underline{0.316}\\
        Ours & \multicolumn{2}{c}{\textbf{0.525}} & \multicolumn{2}{c}{\underline{0.544}} & \multicolumn{2}{c}{\underline{0.144}} & \underline{0.092} & \textbf{0.416}\\
        \midrule
        \multicolumn{9}{c}{Erasing \textit{\textbf{Snoopy}, \textbf{Mickey} and \textbf{Spongebob}}} \\
        \toprule
        & \multicolumn{2}{c}{$\mathrm{LPIPS}_\mathrm{e}\uparrow$} & \multicolumn{2}{c}{$\mathrm{LPIPS}_\mathrm{e}\uparrow$} & \multicolumn{2}{c}{$\mathrm{LPIPS}_\mathrm{e}\uparrow$} & $\mathrm{LPIPS}_\mathrm{u}\downarrow$ & $\mathrm{LPIPS}_\mathrm{da}\uparrow$ \\
        \midrule
        ESD-X & \multicolumn{2}{c}{\underline{0.503}} & \multicolumn{2}{c}{\underline{0.548}} & \multicolumn{2}{c}{\textbf{0.608}} & 0.445 & 0.108\\
        UCE & \multicolumn{2}{c}{0.513} & \multicolumn{2}{c}{0.534} & \multicolumn{2}{c}{0.559} & 0.393 & 0.142\\
        MACE & \multicolumn{2}{c}{0.485} & \multicolumn{2}{c}{0.498} & \multicolumn{2}{c}{0.521} & 0.434 & 0.067\\
        SPM & \multicolumn{2}{c}{0.380} & \multicolumn{2}{c}{0.395} & \multicolumn{2}{c}{0.443} & \textbf{0.104} & \underline{0.302}\\
        Ours & \multicolumn{2}{c}{\textbf{0.538}} & \multicolumn{2}{c}{\textbf{0.559}} & \multicolumn{2}{c}{\underline{0.602}} & \underline{0.122} & \textbf{0.444}\\
        \bottomrule
    \end{tabular}
    \end{adjustbox}
    \caption{Quantitative results of different models on Cartoon Concept Removal. \textbf{Bold}: best. \underline{Underline}: second-best.}
    \label{tab:metrics of Cartoon Concept Removal}
    \vspace{-0.5cm}
\end{table}

\subsubsection{Results of Explicit Content Erasure}
As shown in Tab. \textcolor{red}{\ref{tab:metrics of erasures}}, we observe that our method exhibits the lowest amount of nude body parts and achieves the second-best performance in terms of the FID score compared to all other methods. Notably, our FID score is close to that of the top-performing method, UCE, and outperforms other methods by a wide margin.
Moreover, our method achieves an excellent CS. Although other methods like UCE and SLD-Med also demonstrate good FID and CS, they are significantly ineffective for erasure.
Furthermore, compared to other methods that also utilize additional modules, such as MACE and SPM, our approach outperforms them in both erasure performance and generative capability. 
Notably, our method achieves superior preservation effects without incorporating any preservation loss, highlighting the effectiveness and superiority of our model structure.

% The results in the Tab. \textcolor{red}{\ref{tab:metrics of erasures}} indicate that, our method generates least amount of nude body parts, with the second-best performance with regards to FID score among all the methods. which is close to the top-performing method UCE and outperform other methods by a wide margin. Moreover, our method achieves an excellent CS. Although other methods like UCE and SLD-Med also demonstrate good FID and CS, they are significantly ineffective for erasure. Furthermore, compared to other methods that also utilize additional modules, such as MACE and SPM, our approach outperforms them in both erasure performance and generative capability. Notably, our superior preservation effect is achieved without incorporating any preservation loss, highlighting the effectiveness and superiority of our model structure.

To further show the superior generative capability, we visualize generated images of ``nudity'' in Fig. \textcolor{red}{\ref{fig:nudecomparison}}.
As can be seen, UCE occasionally introduces minor alterations to the nude content of the image (the second row) or generates images with visual anomalies (the third row).
In addition, MACE, SPM, and SDD have a more pronounced impact on the structure and style of the image.
In contrast, our method even preserves the posture of the character, as shown in the first row.
% As illustrated in Fig. \textcolor{red}{\ref{fig:nudecomparison}}, UCE occasionally introduces minor alterations to the nude content of the image or generates images with visual anomalies. In addition, MACE, SPM and SDD have a more pronounced impact on the structure and style of the image. Remarkably, as shown in the first image, our method even preserves the posture of the character.

\subsection{Cartoon Concept Removal}

\subsubsection{Experiment Setup}

Follow the SPM~\cite{lyu2024one}, we evaluate single and multi-concept erasure in the application of cartoon concept removal. 
Popular cartoon character ``Snoopy" is taken as an example, after selecting a group of cartoon character names that are familiar to the public, the dictionary of the CLIP text tokenizer is utilized to filter out the three cartoon character concepts that are closesly related to Snoopy, namely Mickey, SpongeBob and Pikachu, under the criterion of cosine similarity. 
We demonstrate the effectiveness of our method on single concept removal and multi-concept removal with three sets of experiments, erasing ``Snoopy", erasing ``Snoopy" and ``Mickey", erasing ``Snoopy", ``Mickey" and ``Spongebob". 
The remaining cartoon characters are then used to evaluate the model's preservation ability. 
As to each cartoon concept, a total of 80 templates proposed in CLIP~\cite{radford2021learning} are employed for text prompt augmentation to improve the accuracy of the assessments. For each template, 5 images are generated with the seed 2024.

We adopt $\mathrm{LPIPS}_\mathrm{e}$ and $\mathrm{LPIPS}_\mathrm{u}$ to evaluate the LPIPS score of the erased and unerased concepts. 
We also utilize $\mathrm{LPIPS}_\mathrm{da} = Avg(\mathrm{LPIPS}_\mathrm{e}) - Avg(\mathrm{LPIPS}_\mathrm{u})$ to assess the trade-off between erasure and preservation capabilities.
% as follows:
% \begin{equation}
%     \mathrm{LPIPS}_\mathrm{da} = Avg(\mathrm{LPIPS}_\mathrm{e}) - Avg(\mathrm{LPIPS}_\mathrm{u}).
%     \label{eq:LPIPS-DA}
% \end{equation}

\subsubsection{Results of Cartoon Concept Removal}
%Due to space constraints, qualitative results are shown in Appendix \textcolor{red}{A}.
Tab. \textcolor{red}{\ref{tab:metrics of Cartoon Concept Removal}} presents the quantitative results. 
For single concept removal, ESD and UCE achieve higher $\mathrm{LPIPS}_\mathrm{u}$ scores. This suggests that they significantly impair the generative ability of other cartoon concepts and aren't capable of decoupling these cartoon concepts. 
In contrast, our method yields comparable retention of other concepts.
Additionally, compared to the best method, SPM, in retaining other concepts, our method excels in erasing 
\begin{figure}[h]
    \begin{adjustbox}{max width=0.47\textwidth}
    \begin{tabular}{c@{\hskip 0.03in} c@{\hskip 0.03in} c@{\hskip 0.03in} c@{\hskip 0.03in} c@{\hskip 0.03in} c@{\hskip 0.03in} c@{\hskip 0.03in} c@{\hskip 0.00in} }
        & \textbf{Original} & \textbf{ESD} & \textbf{MACE} & \textbf{SPM} & \textbf{Ours} \\
        
        % \begin{subfigure}[t]{.08\textwidth}
        %     \centering
        %     \textit{A portrait of a woman with distorted features, in the style of Picasso's Cubism}
        % \end{subfigure} &

        \begin{minipage}{.18\textwidth}
        \centering
        \textit{A still life of a\\
        bouquet with a mix of \\
        flowers, painted in \textcolor{red}{Van Gogh}'s \\
        signature style}.
        \end{minipage} &
        \begin{minipage}{.14\textwidth}
        \includegraphics[width=\linewidth]{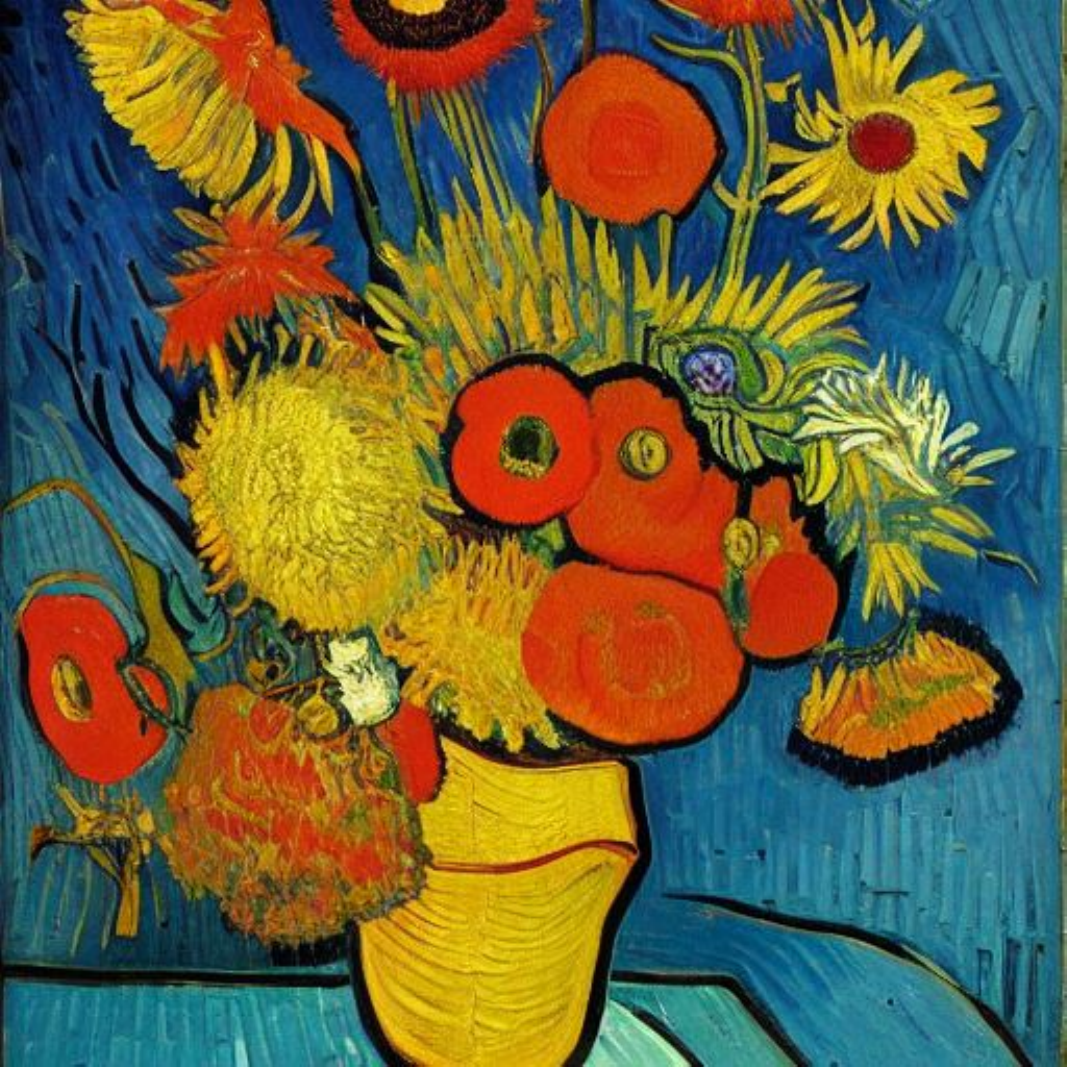}
        \end{minipage} &
        \begin{minipage}{.14\textwidth}
        \includegraphics[width=\linewidth]{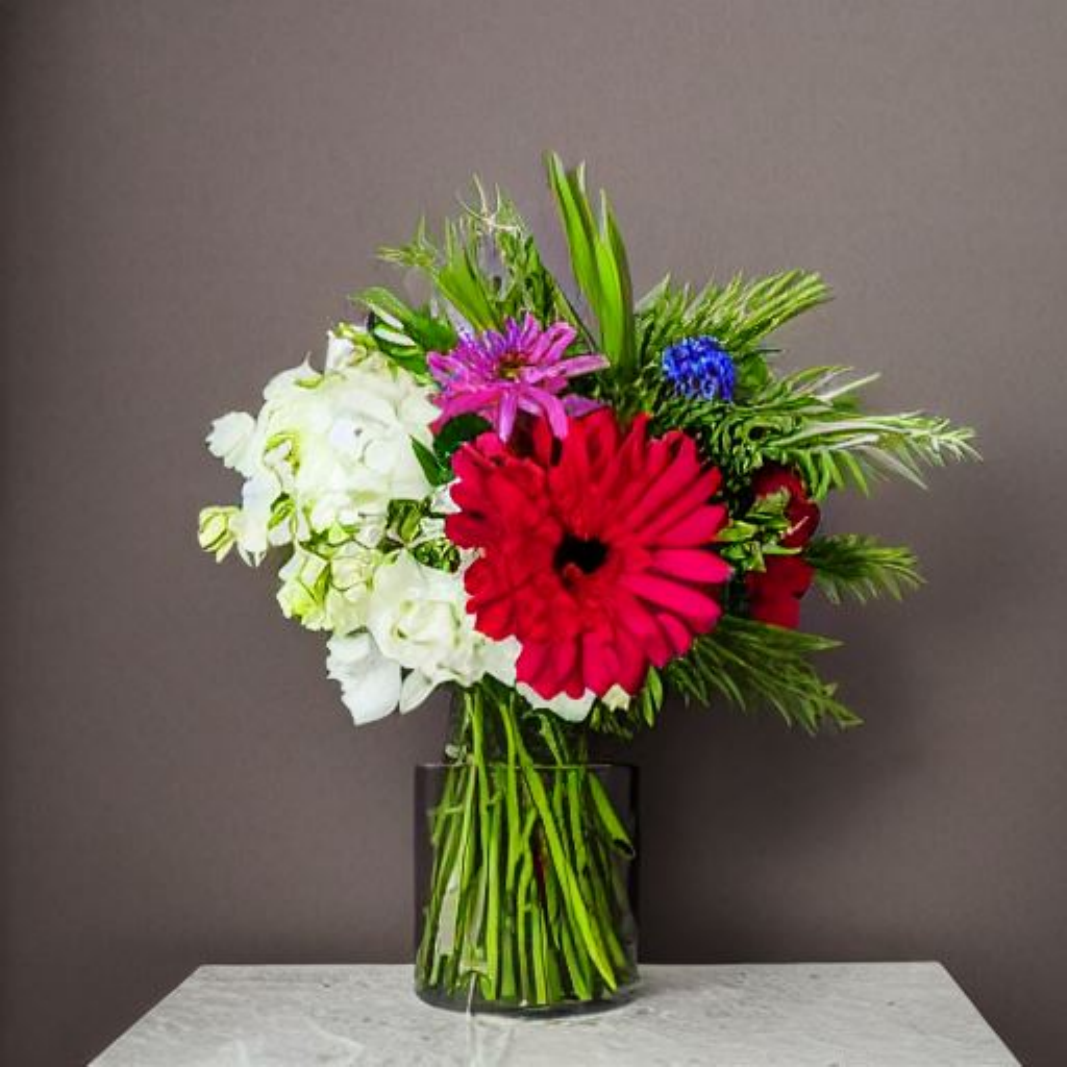}
        \end{minipage} &
        \begin{minipage}{.14\textwidth}
        \includegraphics[width=\linewidth]{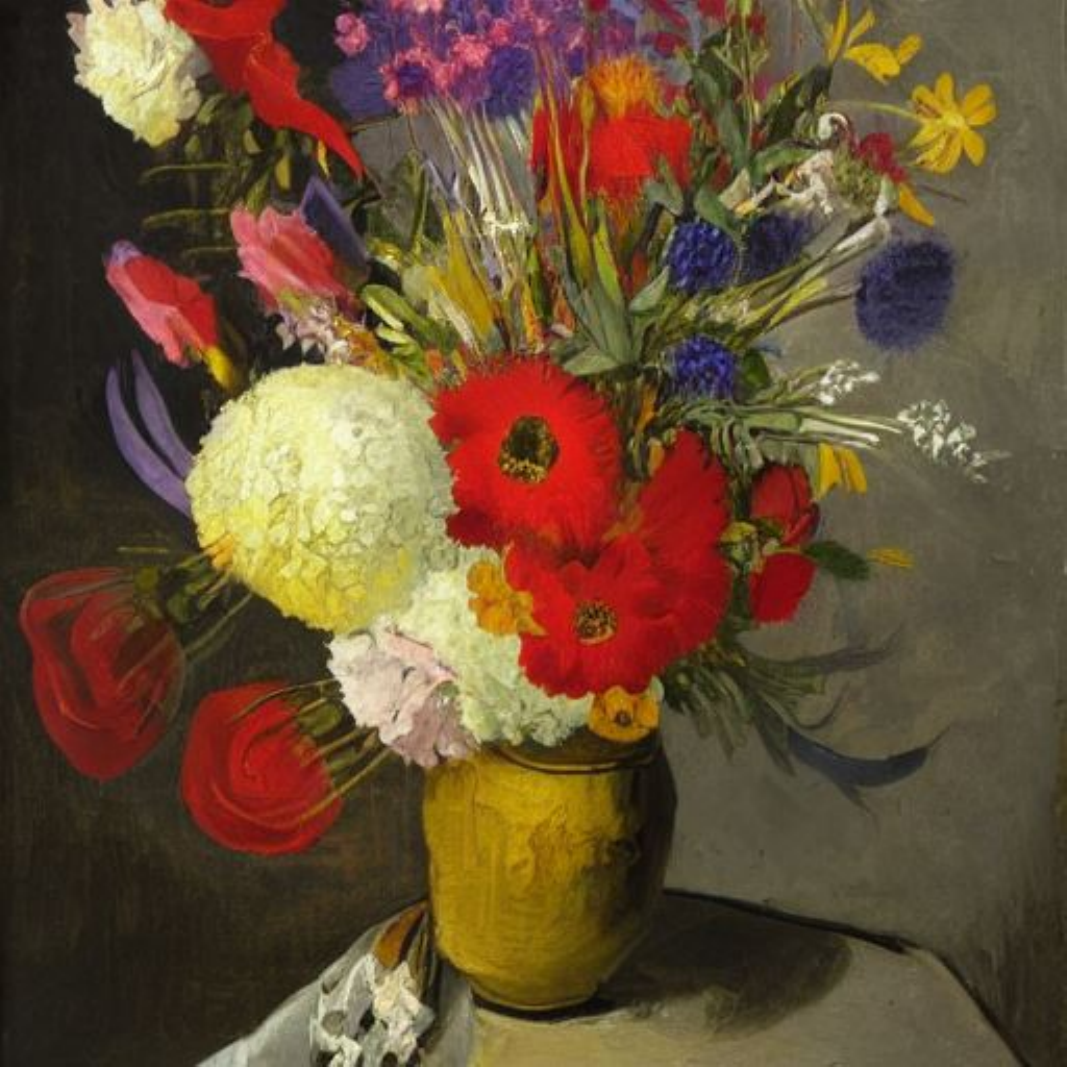}
        \end{minipage} &
        \begin{minipage}{.14\textwidth}
        \includegraphics[width=\linewidth]{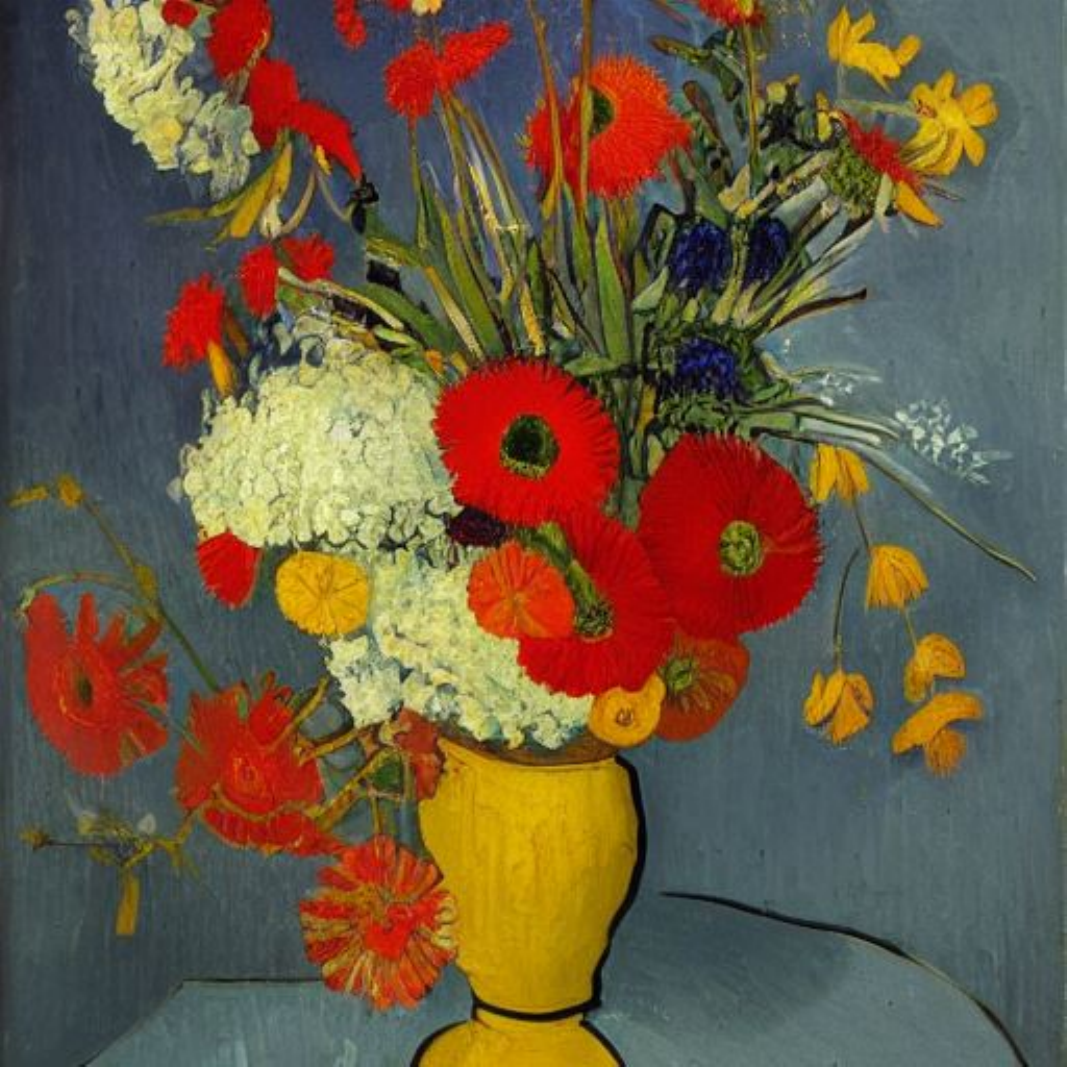}
        \end{minipage} &
        \begin{minipage}{.14\textwidth}
        \includegraphics[width=\linewidth]{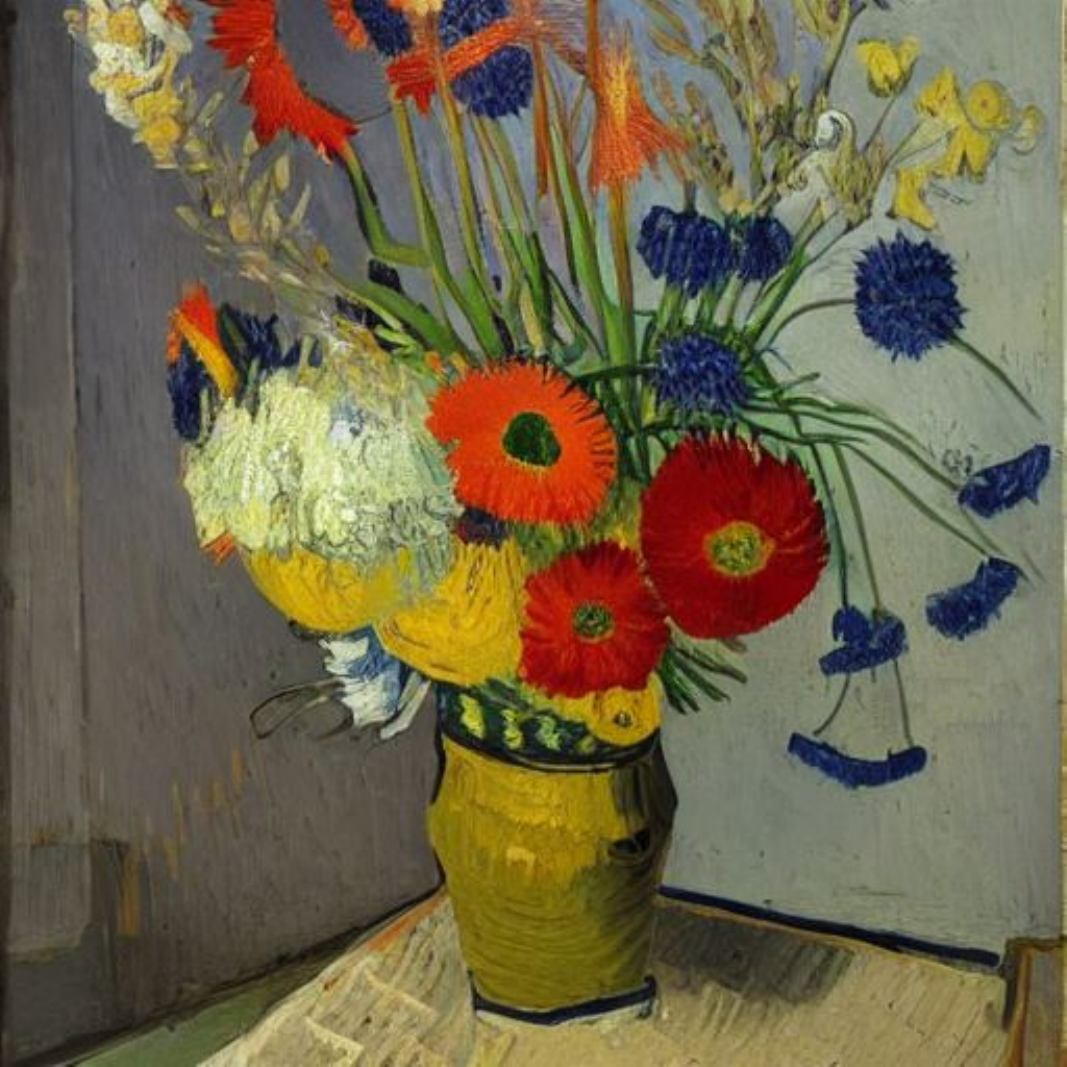}
        \end{minipage} \\
        
        \addlinespace[0.03in]
        
        \begin{minipage}{.18\textwidth}
        \centering
        A glimpse of \textcolor{green}{Rembrandt}'s Amsterdam through his painting.
        \end{minipage} &
        \begin{minipage}{.14\textwidth}
        \includegraphics[width=\linewidth]{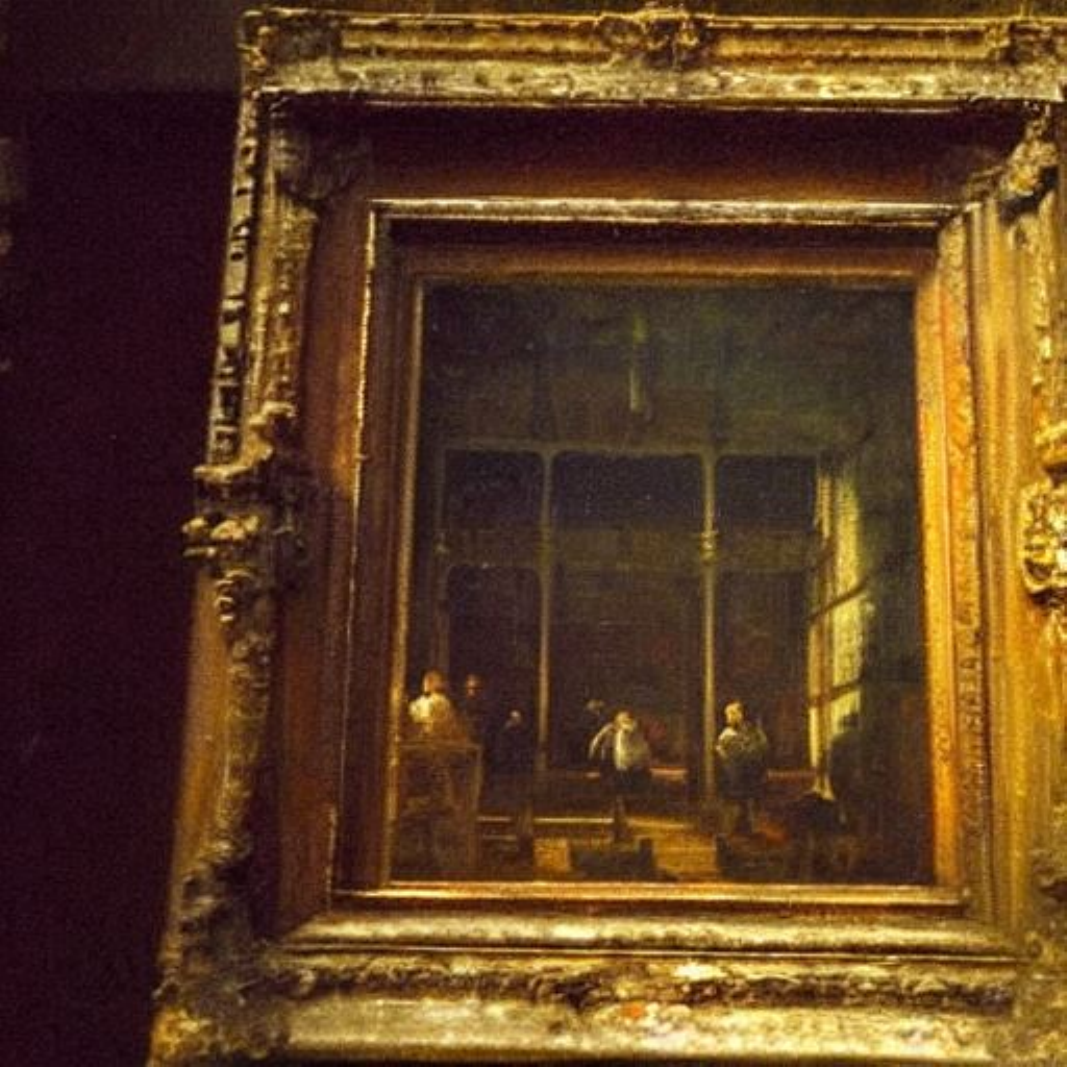}
        \end{minipage} &
        \begin{minipage}{.14\textwidth}
        \includegraphics[width=\linewidth]{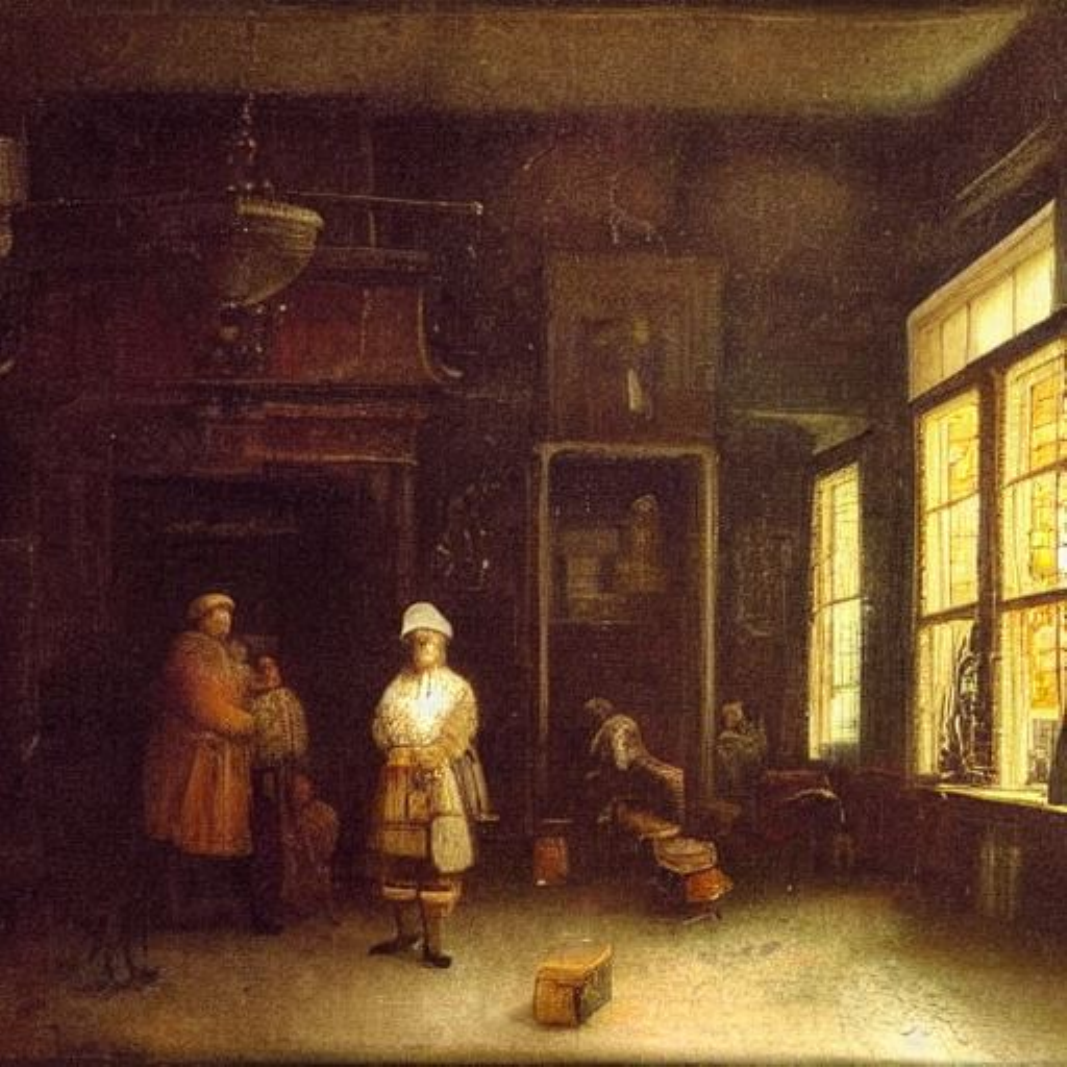}
        \end{minipage} &
        \begin{minipage}{.14\textwidth}
        \includegraphics[width=\linewidth]{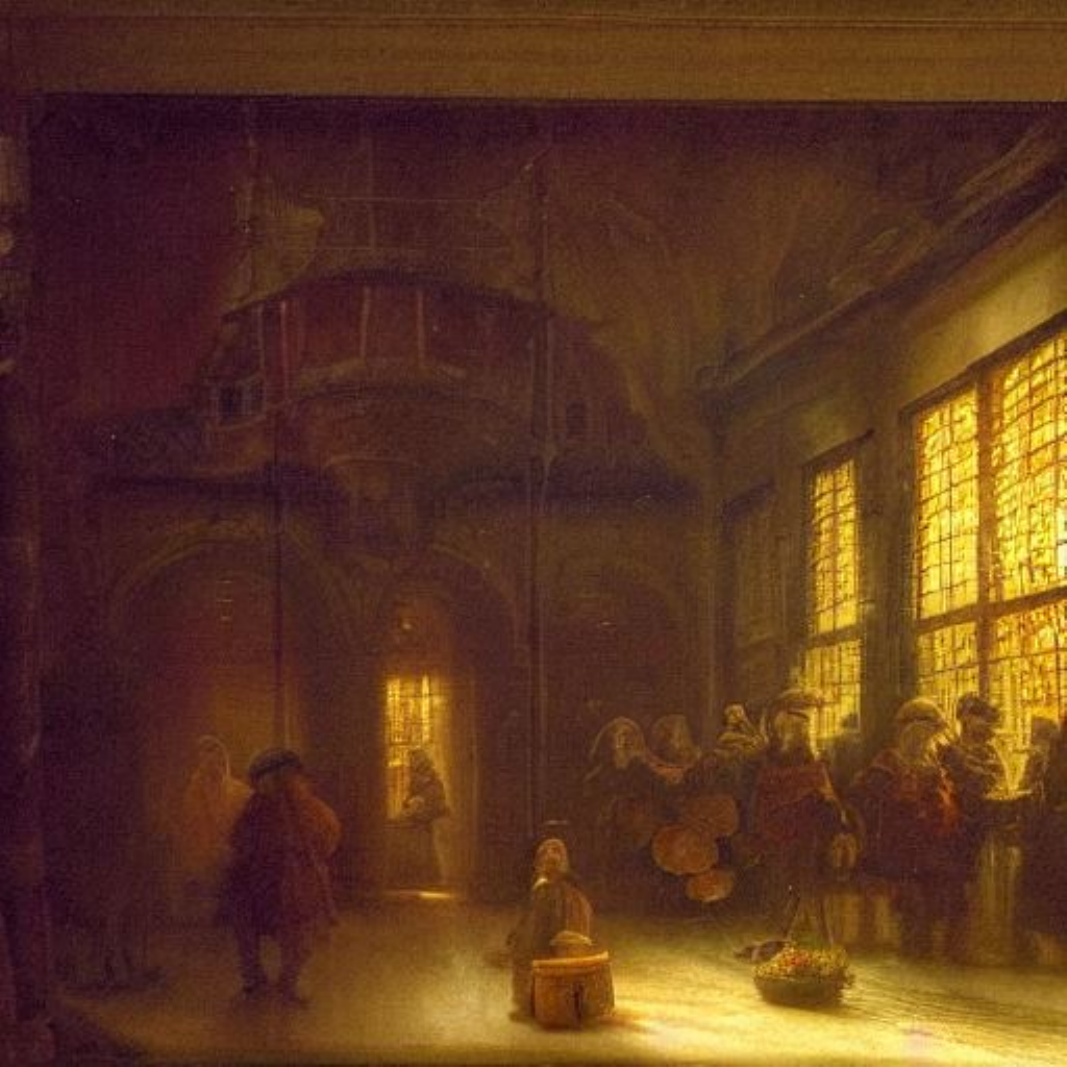}
        \end{minipage} &
        \begin{minipage}{.14\textwidth}
        \includegraphics[width=\linewidth]{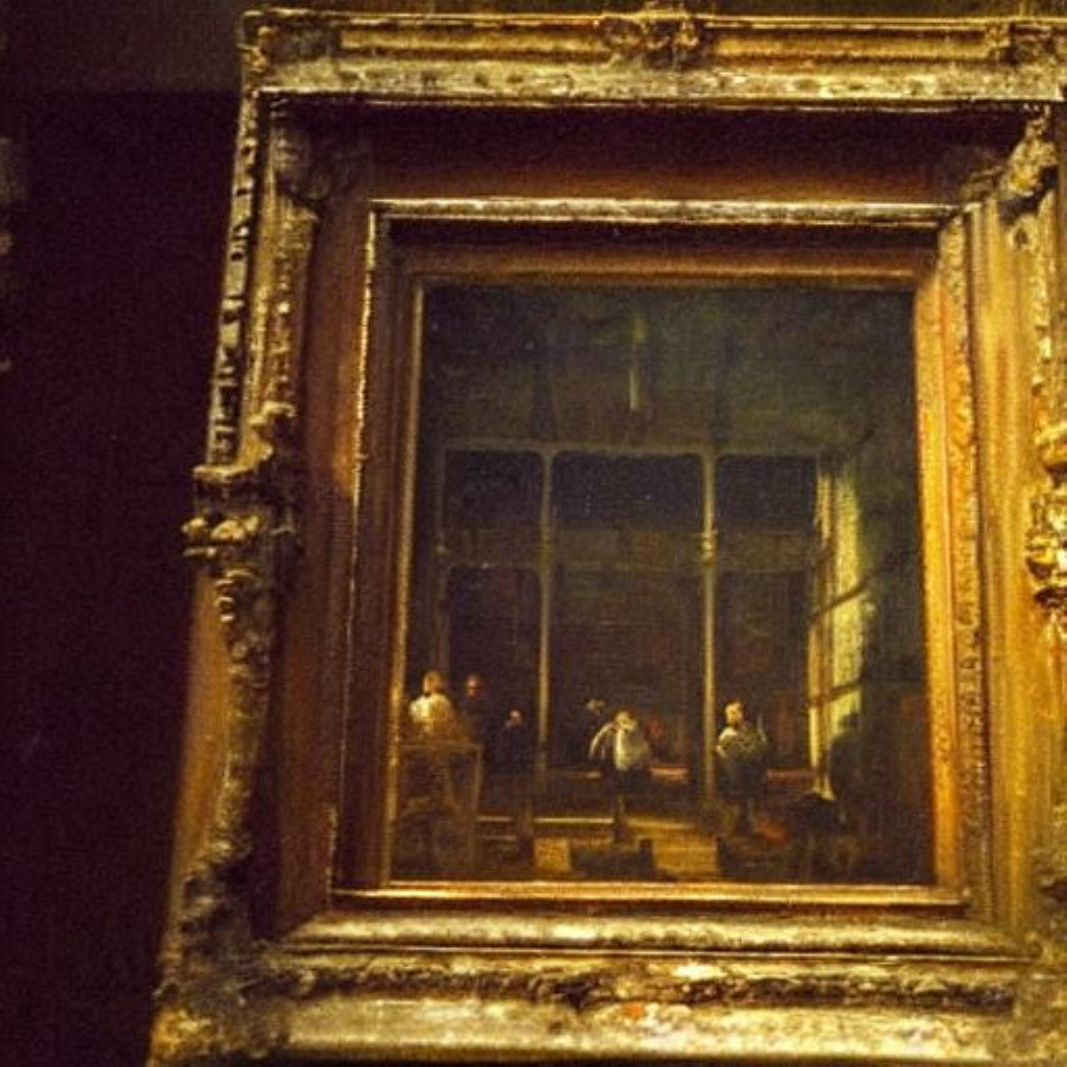}
        \end{minipage} &
        \begin{minipage}{.14\textwidth}
        \includegraphics[width=\linewidth]{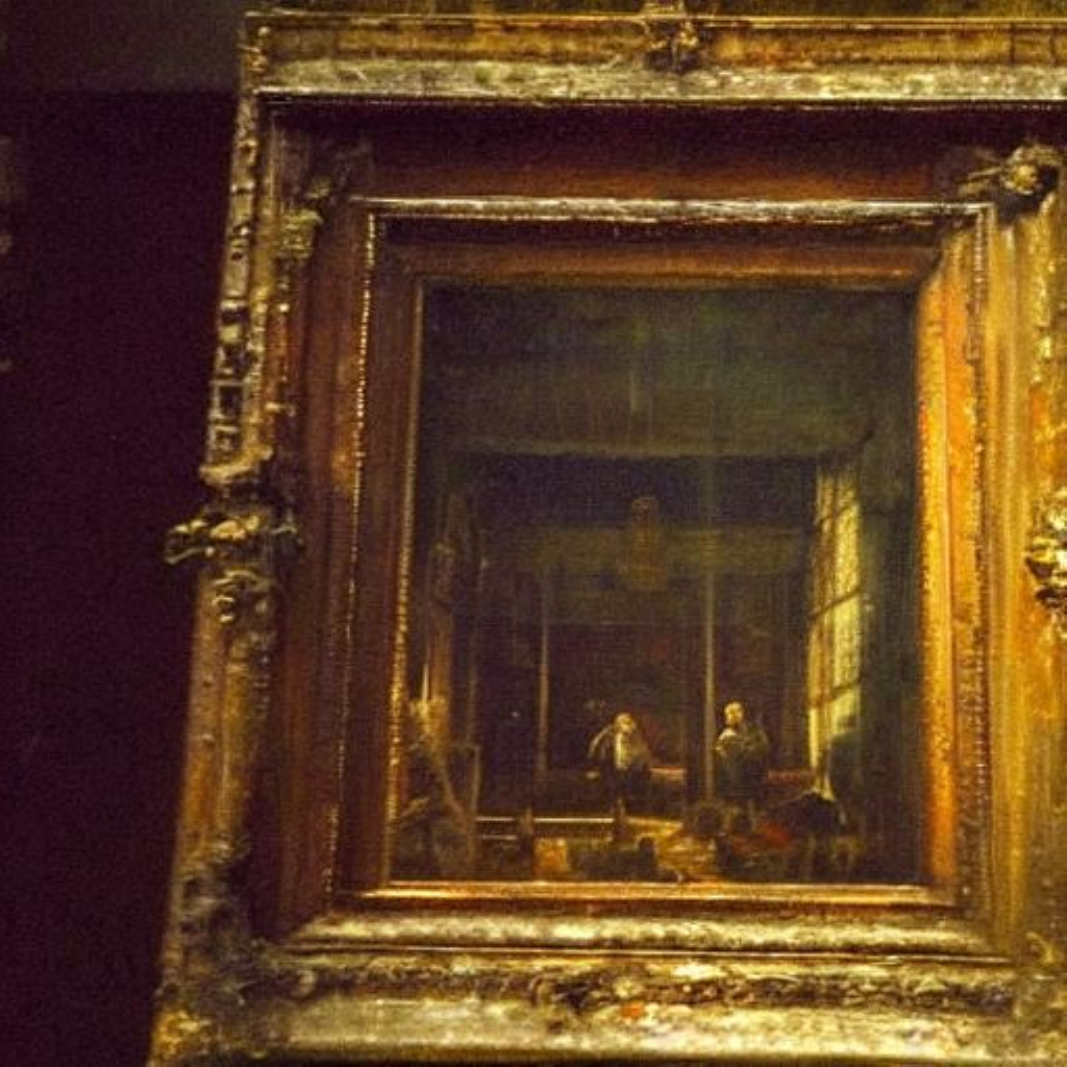}
        \end{minipage} \\
        
        \addlinespace[0.03in]
        
        \begin{minipage}{.18\textwidth}
        \centering
        A still life of everyday objects with unconventional use of space, in the spirit of \textcolor{green}{Picasso}'s avant-garde vision.
        \end{minipage} &
        \begin{minipage}{.14\textwidth}
        \includegraphics[width=\linewidth]{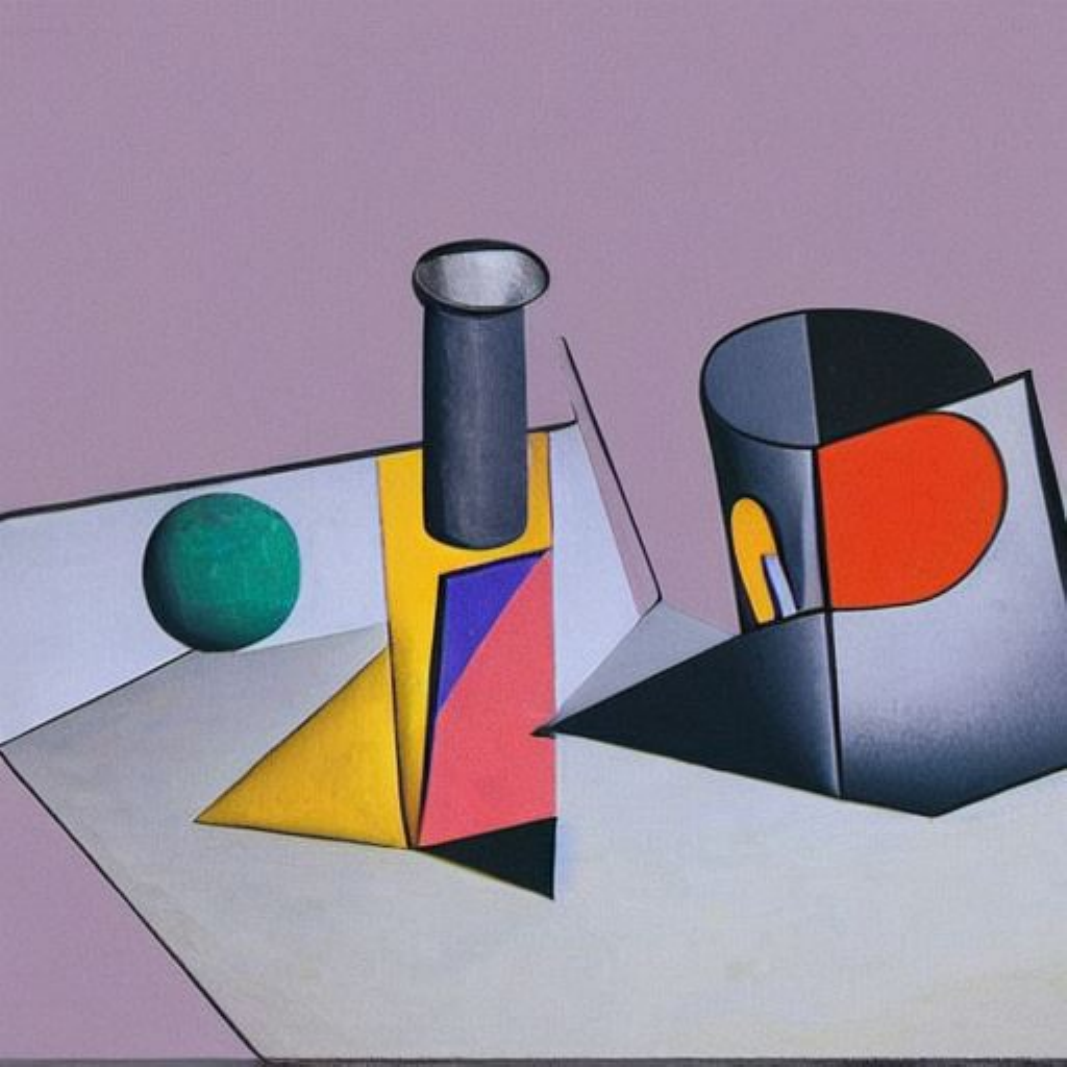}
        \end{minipage} &
        \begin{minipage}{.14\textwidth}
        \includegraphics[width=\linewidth]{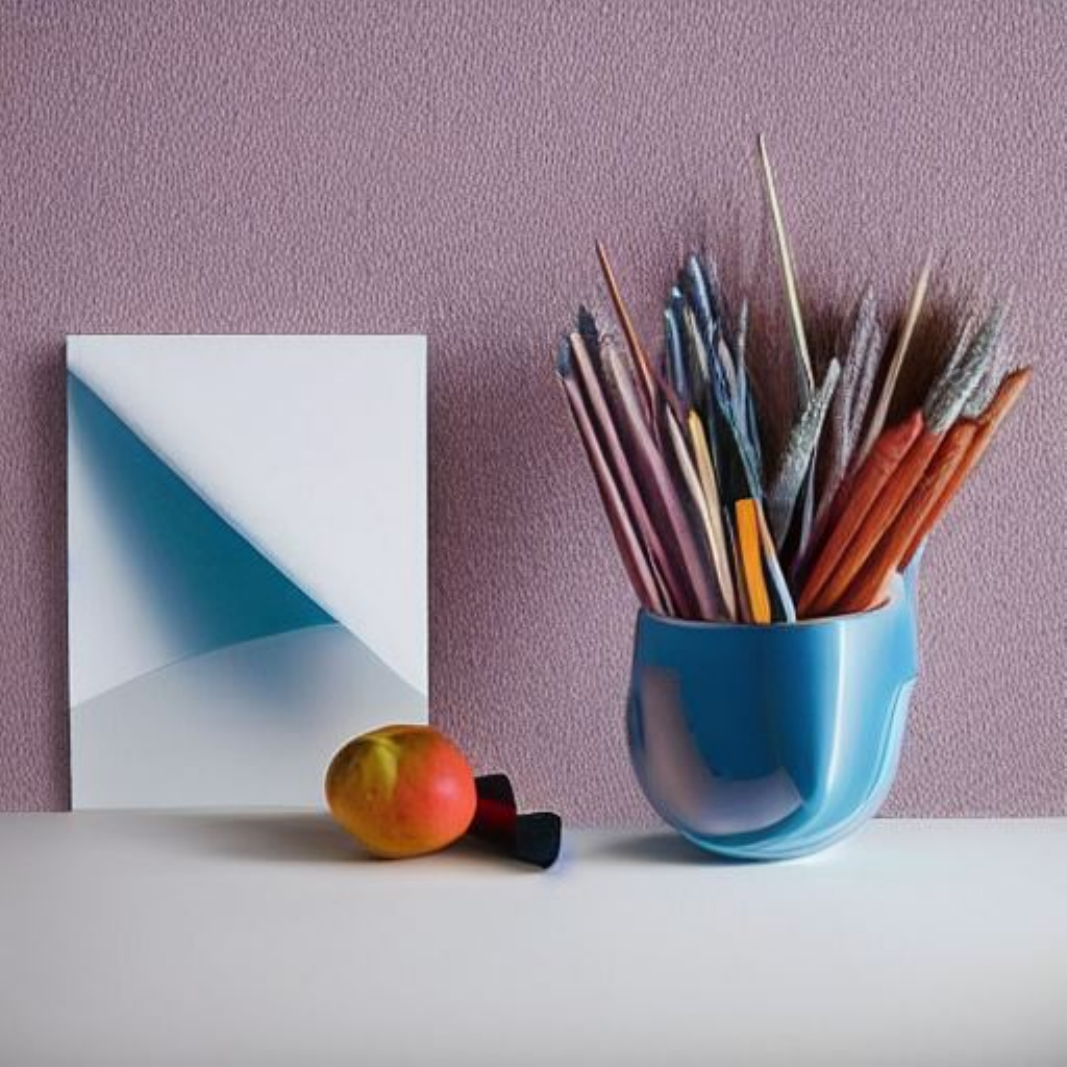}
        \end{minipage} &
        \begin{minipage}{.14\textwidth}
        \includegraphics[width=\linewidth]{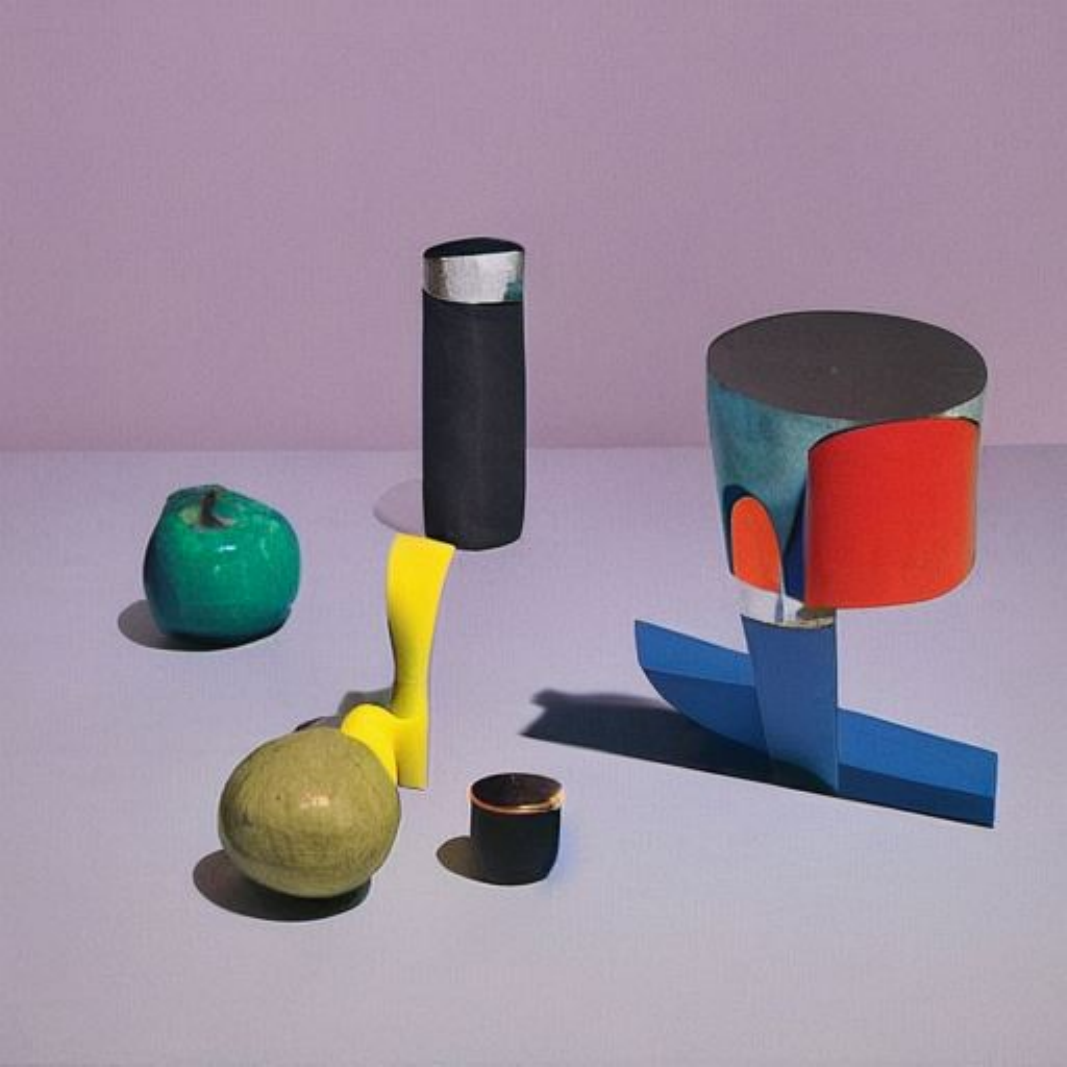}
        \end{minipage} &
        \begin{minipage}{.14\textwidth}
        \includegraphics[width=\linewidth]{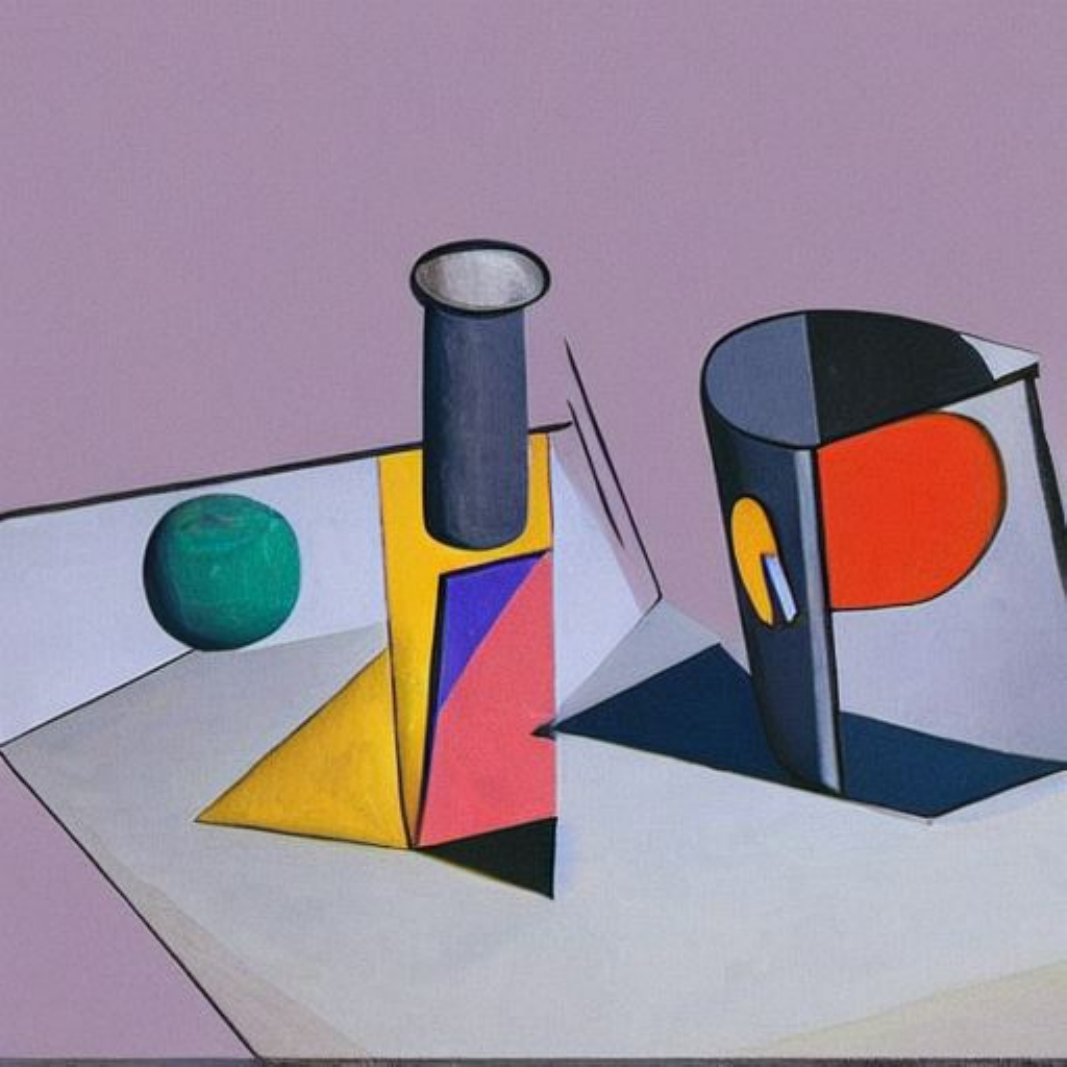}
        \end{minipage} &
        \begin{minipage}{.14\textwidth}
        \includegraphics[width=\linewidth]{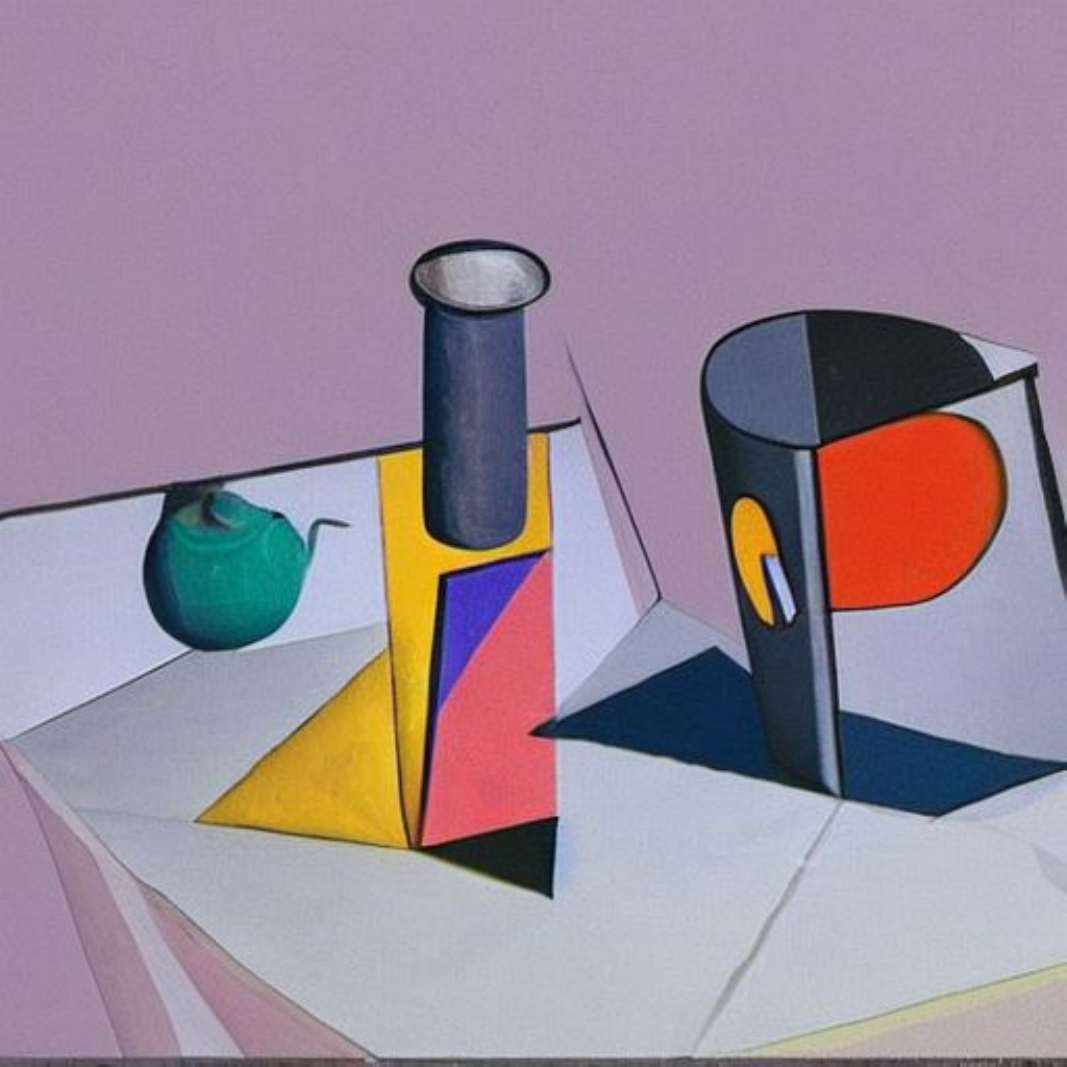}
        \end{minipage} \\

        \addlinespace[0.03in]
        \hdashline[2pt/3pt]
        \addlinespace[0.03in]

        \begin{minipage}{.18\textwidth}
        \centering
        A masterfully painted self-portrait by \textcolor{red}{Rembrandt}.
        \end{minipage} &
        \begin{minipage}{.14\textwidth}
        \includegraphics[width=\linewidth]{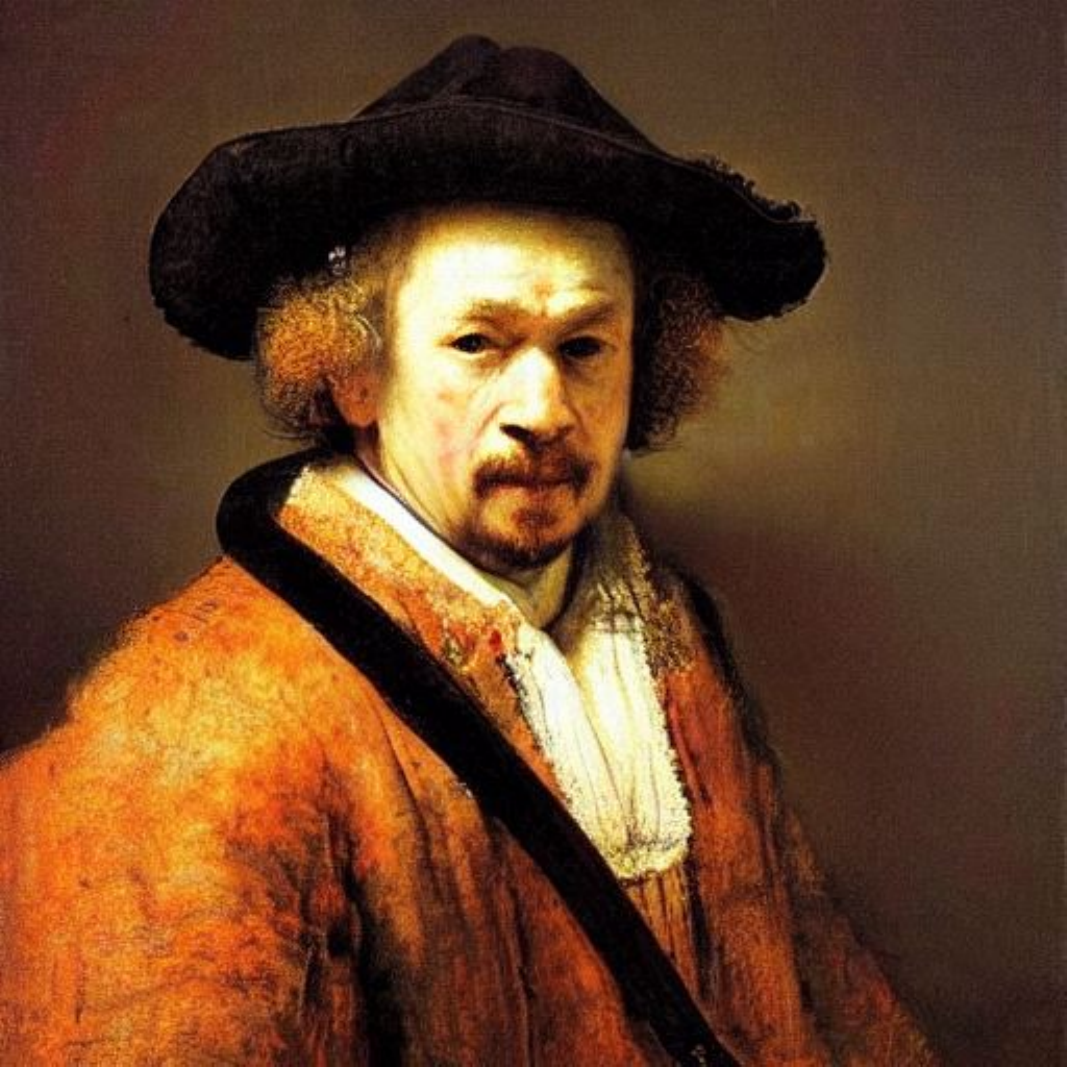}
        \end{minipage} &
        \begin{minipage}{.14\textwidth}
        \includegraphics[width=\linewidth]{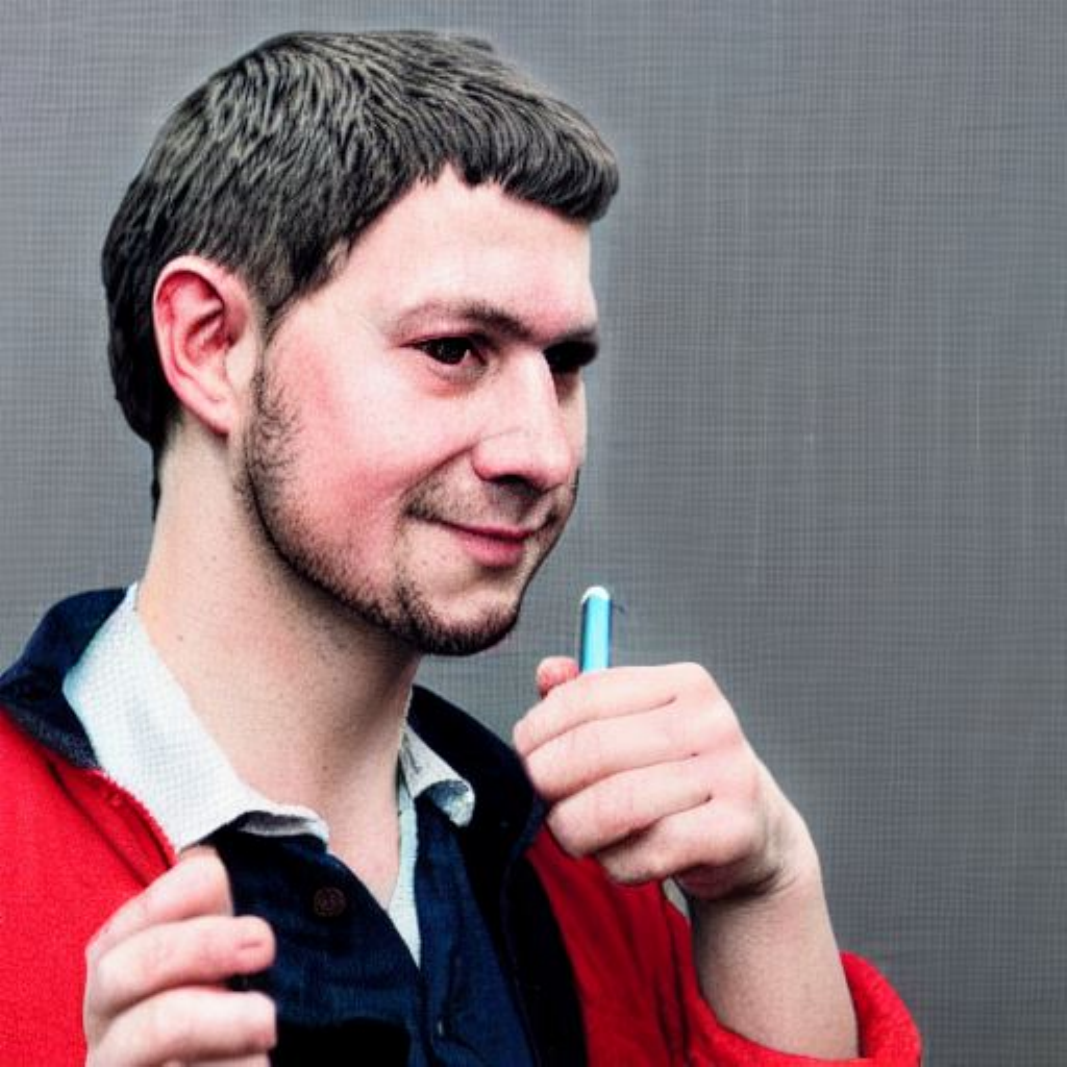}
        \end{minipage} &
        \begin{minipage}{.14\textwidth}
        \includegraphics[width=\linewidth]{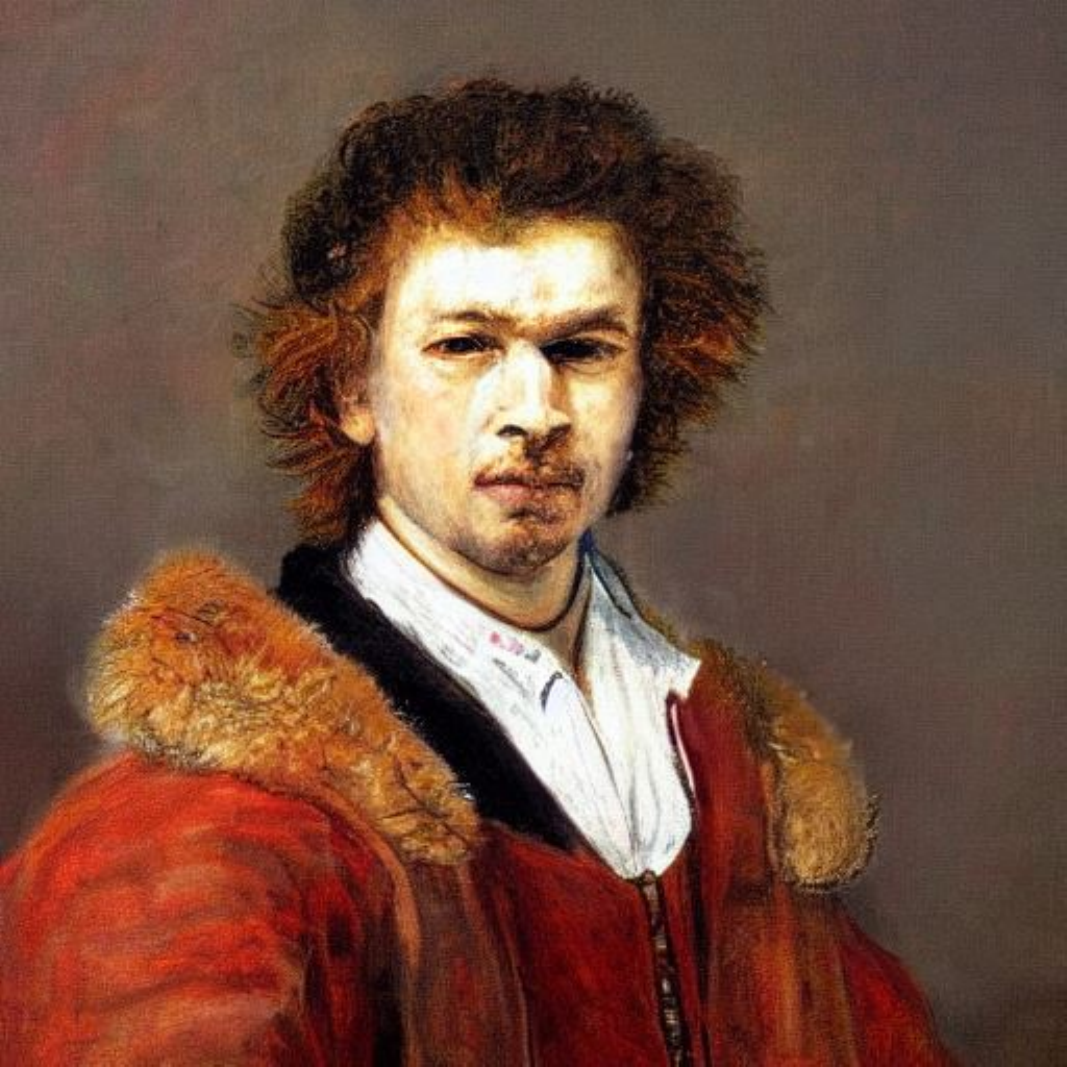}
        \end{minipage} &
        \begin{minipage}{.14\textwidth}
        \includegraphics[width=\linewidth]{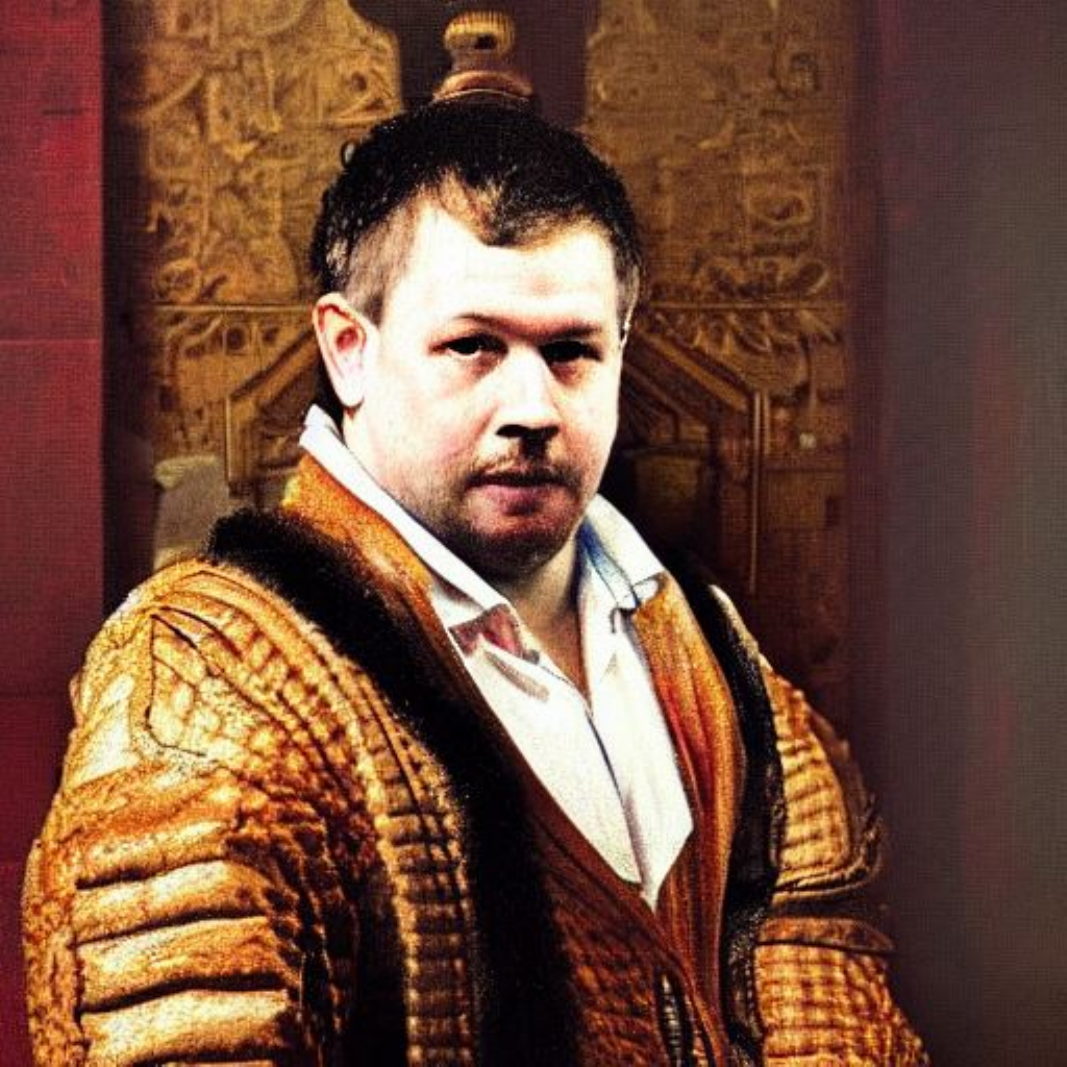}
        \end{minipage} &
        \begin{minipage}{.14\textwidth}
        \includegraphics[width=\linewidth]{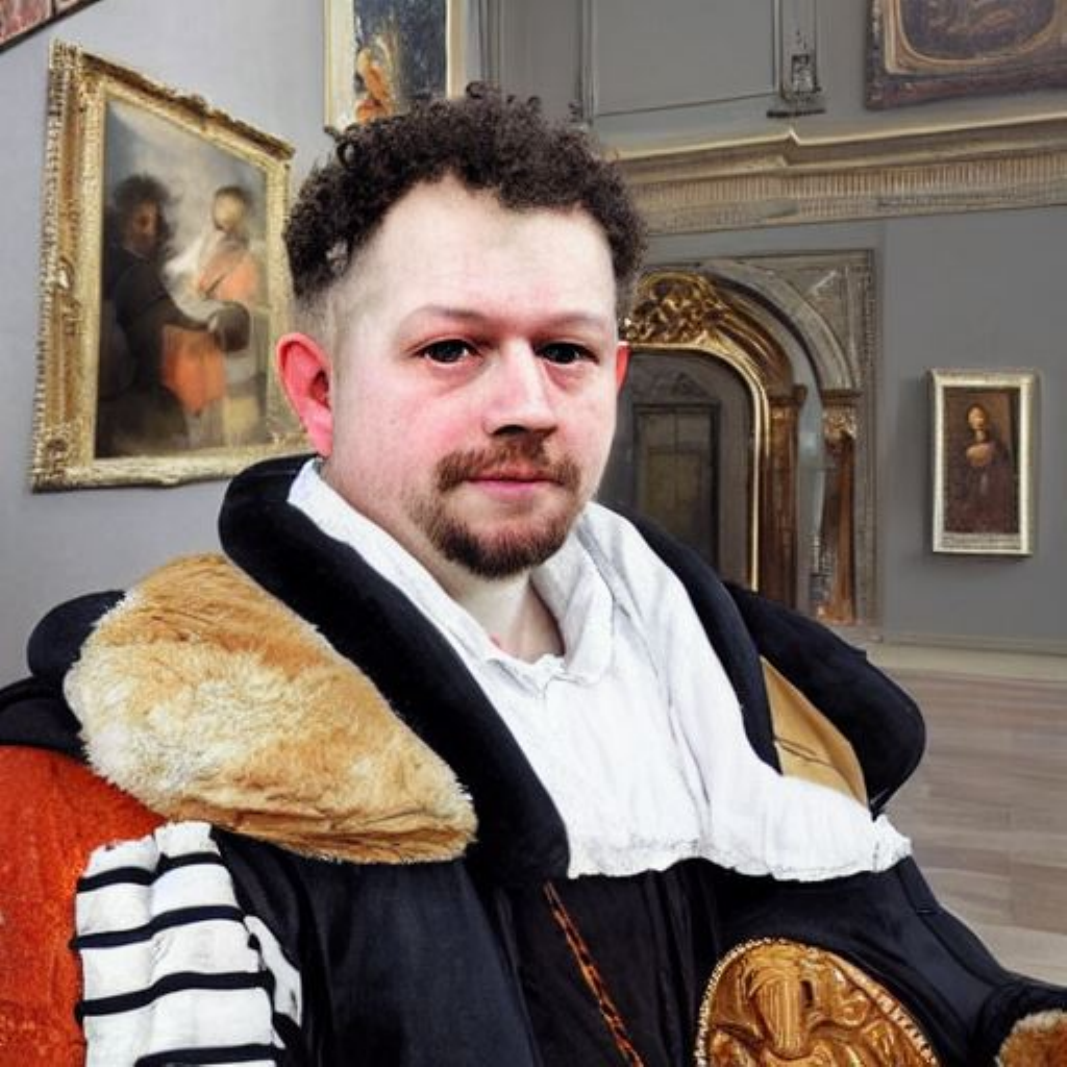}
        \end{minipage} \\

        \addlinespace[0.03in]
        
        \begin{minipage}{.18\textwidth}
        \centering
        A still life of fruit and vegetables with playful use of colors, in the style of \textcolor{green}{Van Gogh}.
        \end{minipage} &
        \begin{minipage}{.14\textwidth}
        \includegraphics[width=\linewidth]{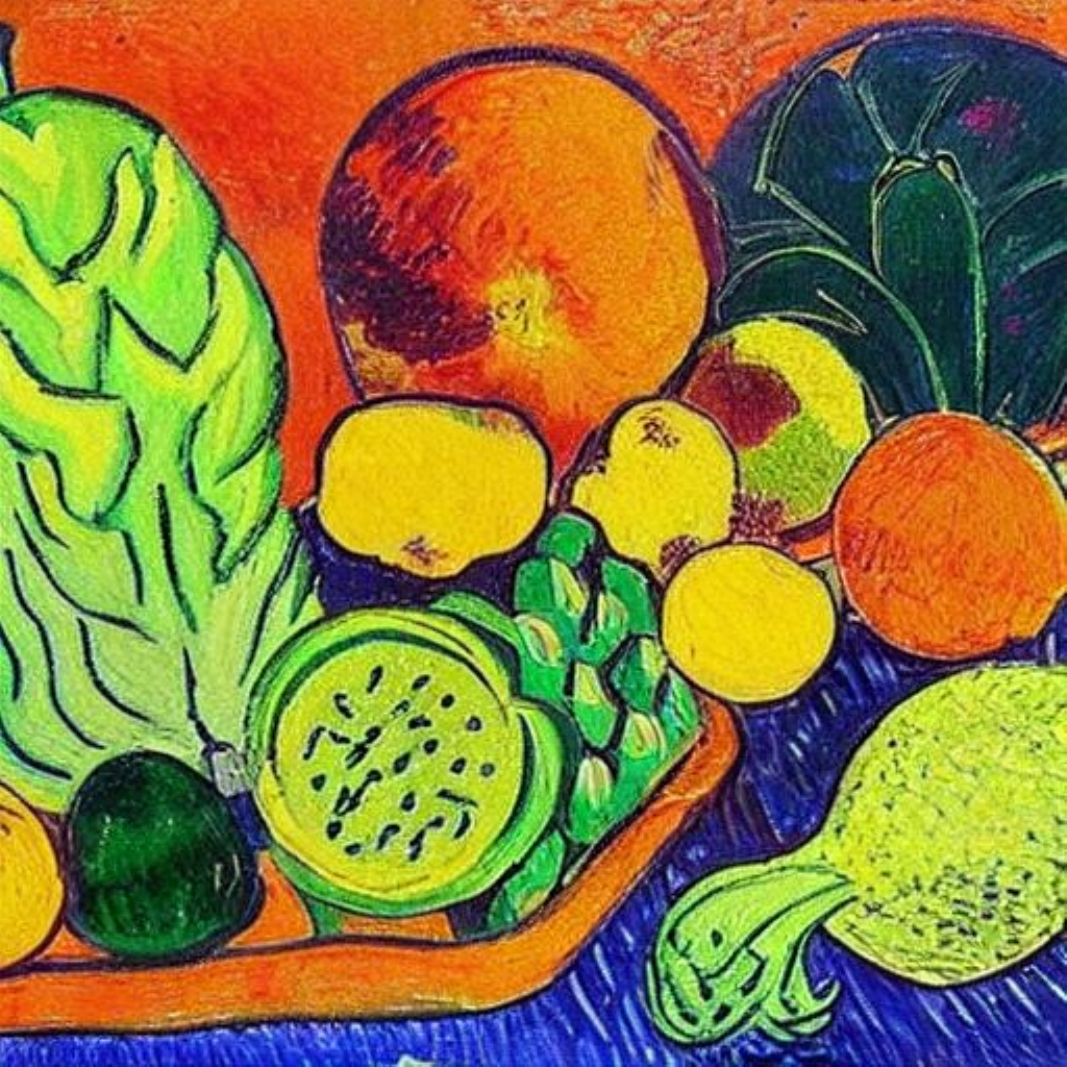}
        \end{minipage} &
        \begin{minipage}{.14\textwidth}
        \includegraphics[width=\linewidth]{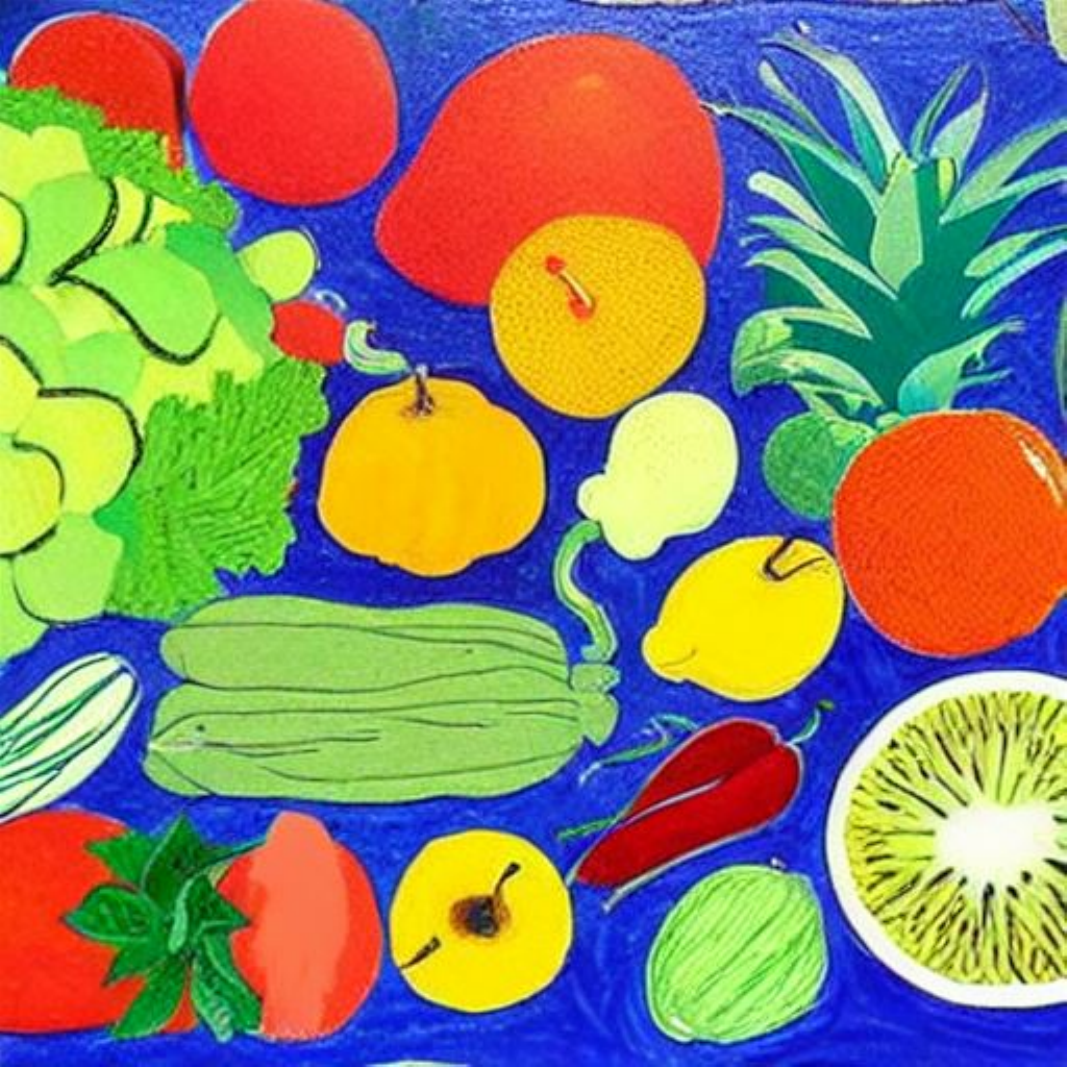}
        \end{minipage} &
        \begin{minipage}{.14\textwidth}
        \includegraphics[width=\linewidth]{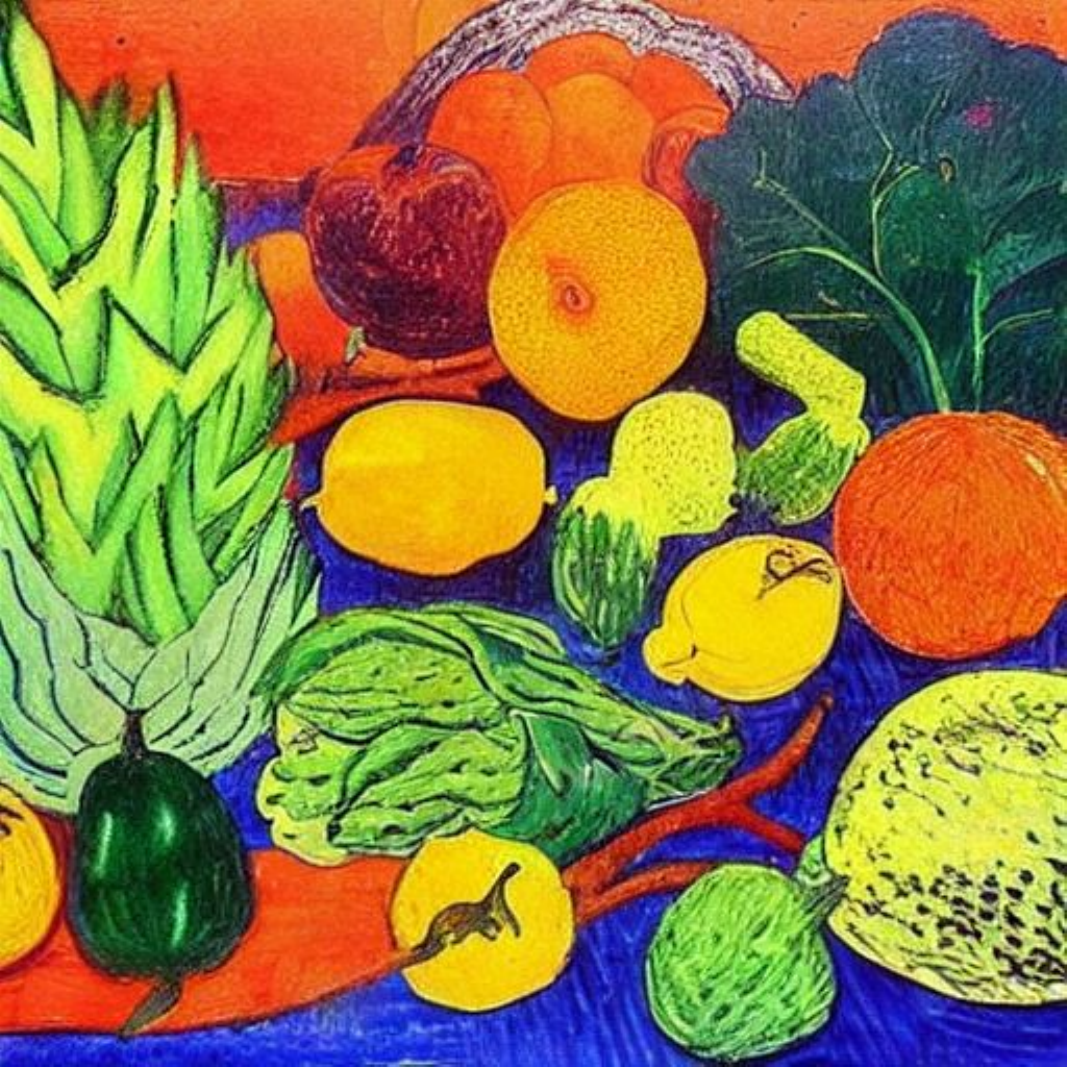}
        \end{minipage} &
        \begin{minipage}{.14\textwidth}
        \includegraphics[width=\linewidth]{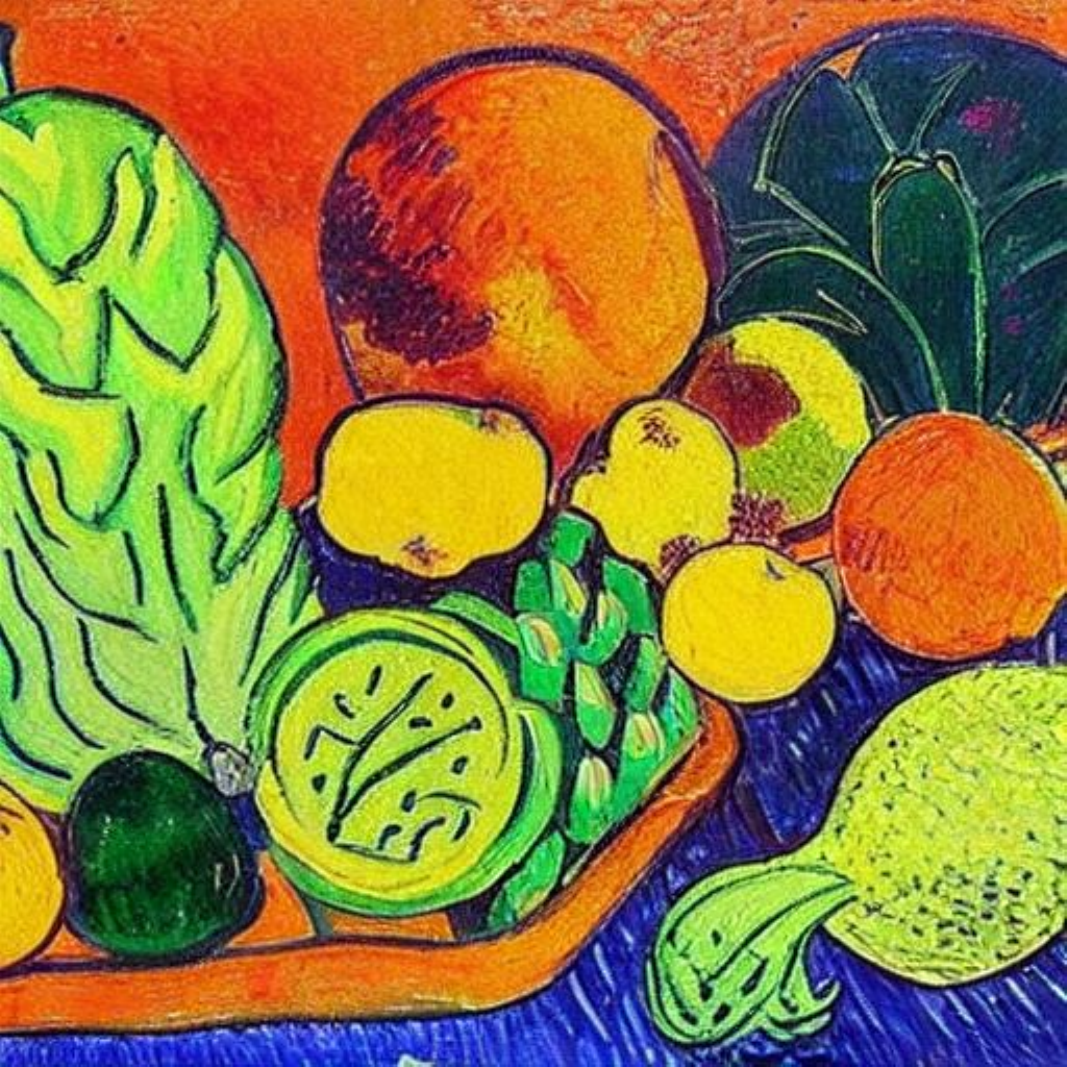}
        \end{minipage} &
        \begin{minipage}{.14\textwidth}
        \includegraphics[width=\linewidth]{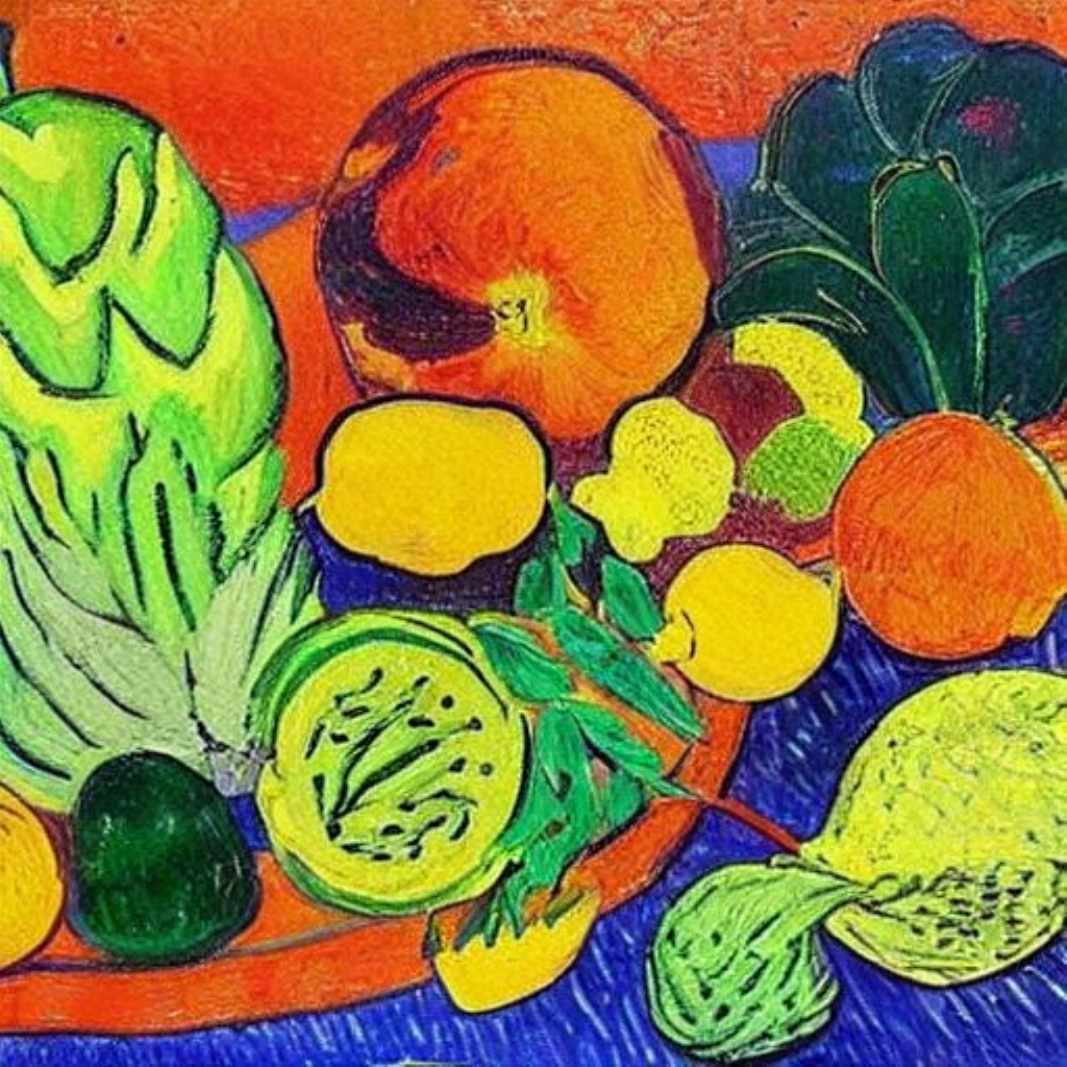}
        \end{minipage} \\

        \addlinespace[0.03in]
        
        \begin{minipage}{.18\textwidth}
        \centering
        A vibrant and energetic image of Elvis Presley by \textcolor{green}{Warhol}.
        \end{minipage} &
        \begin{minipage}{.14\textwidth}
        \includegraphics[width=\linewidth]{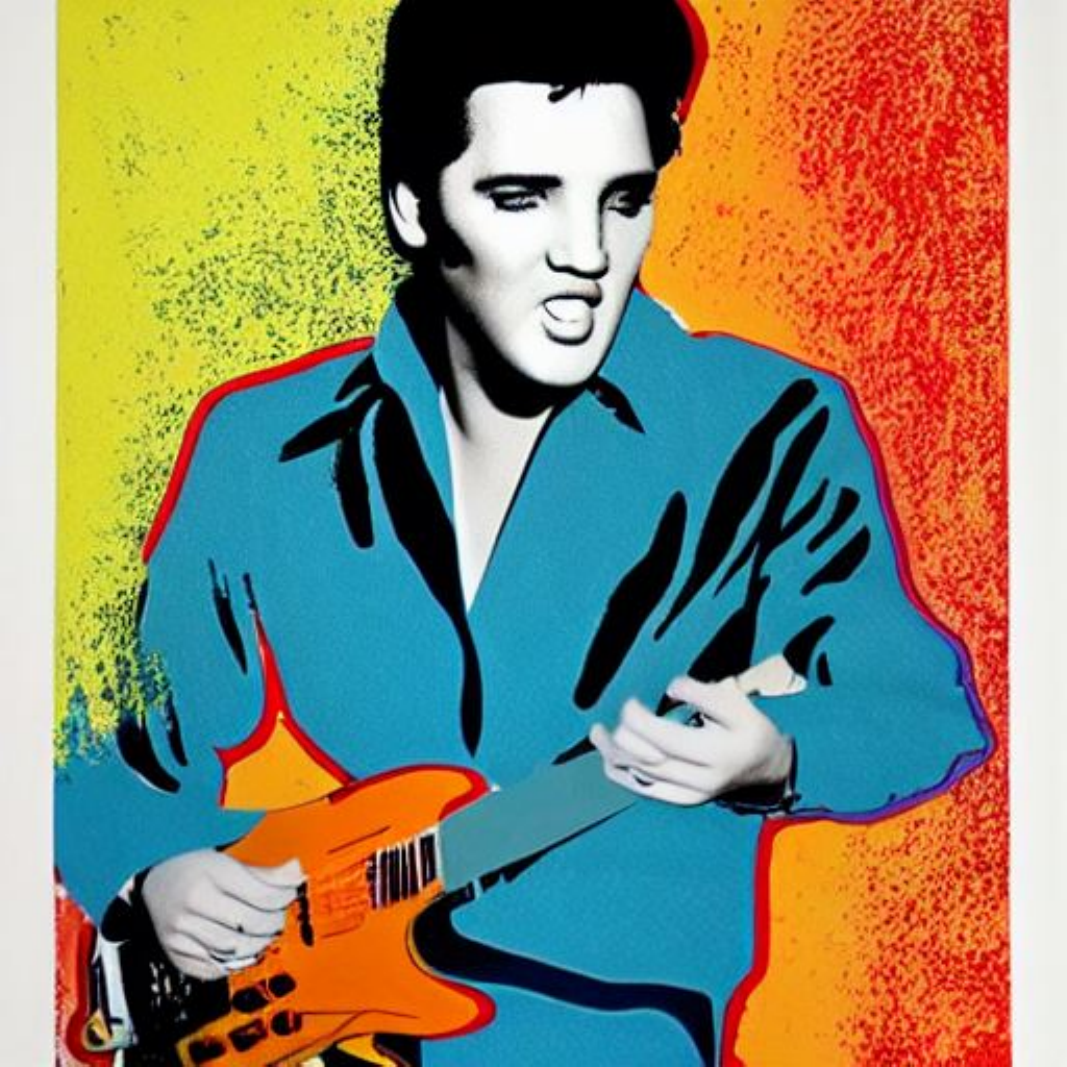}
        \end{minipage} &
        \begin{minipage}{.14\textwidth}
        \includegraphics[width=\linewidth]{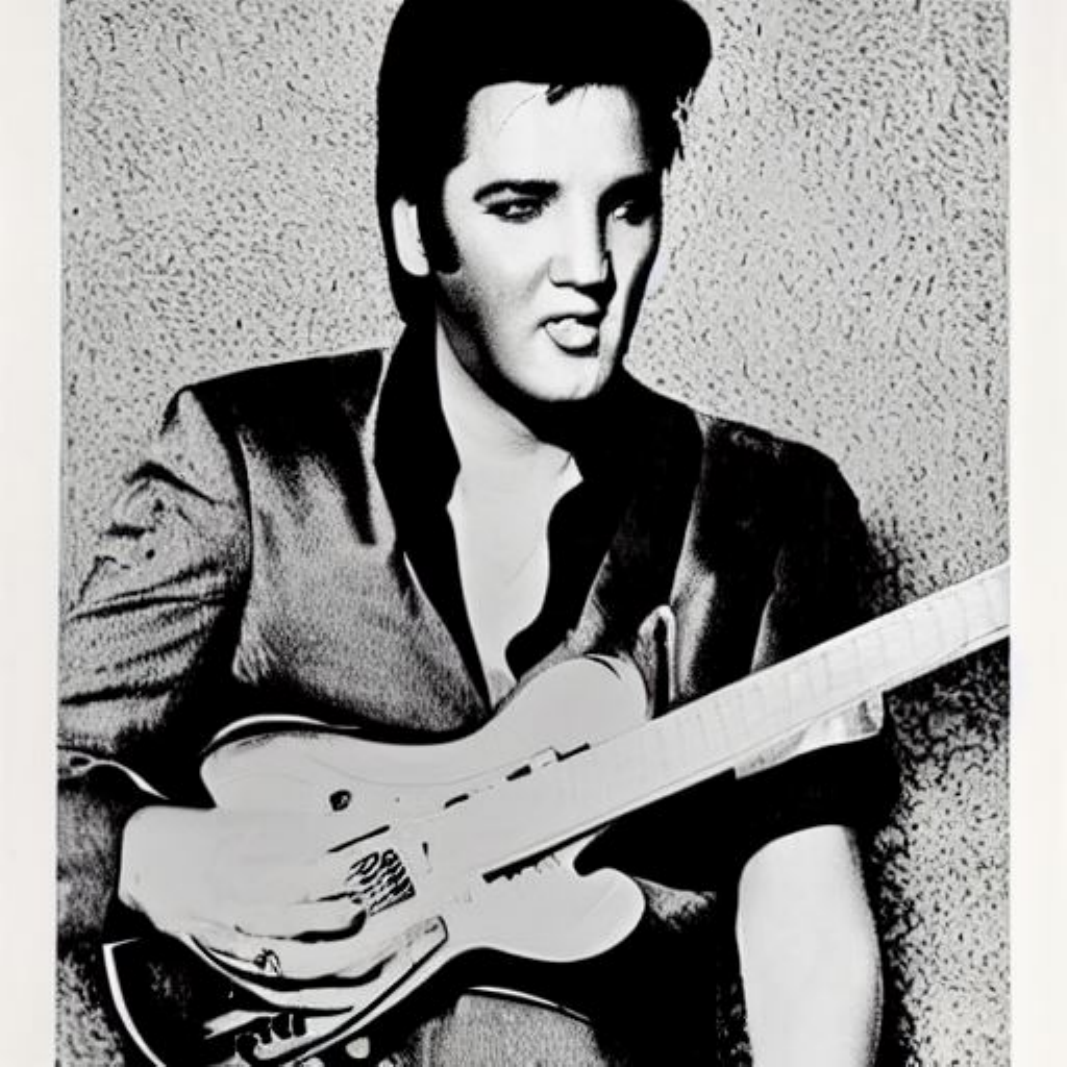}
        \end{minipage} &
        \begin{minipage}{.14\textwidth}
        \includegraphics[width=\linewidth]{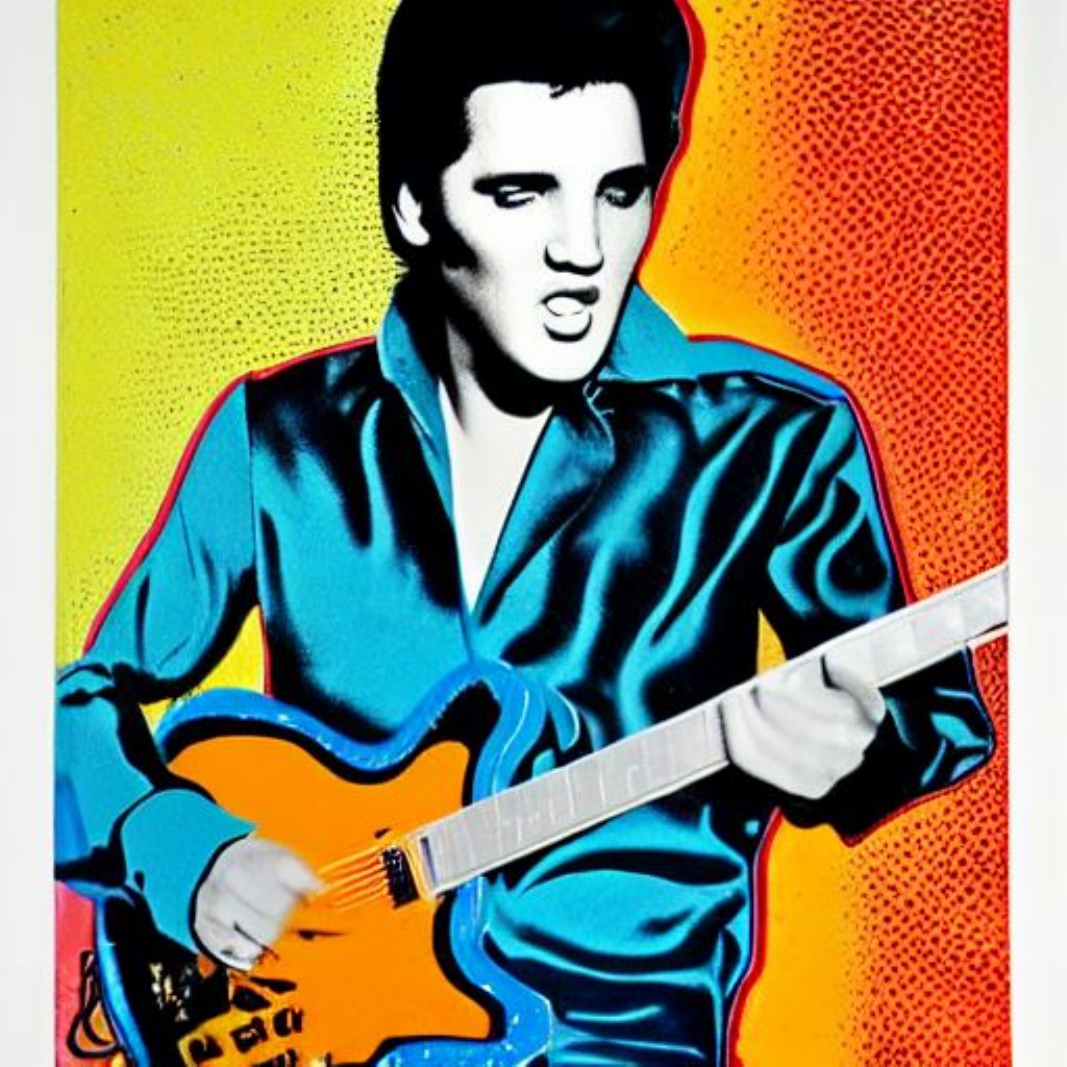}
        \end{minipage} &
        \begin{minipage}{.14\textwidth}
        \includegraphics[width=\linewidth]{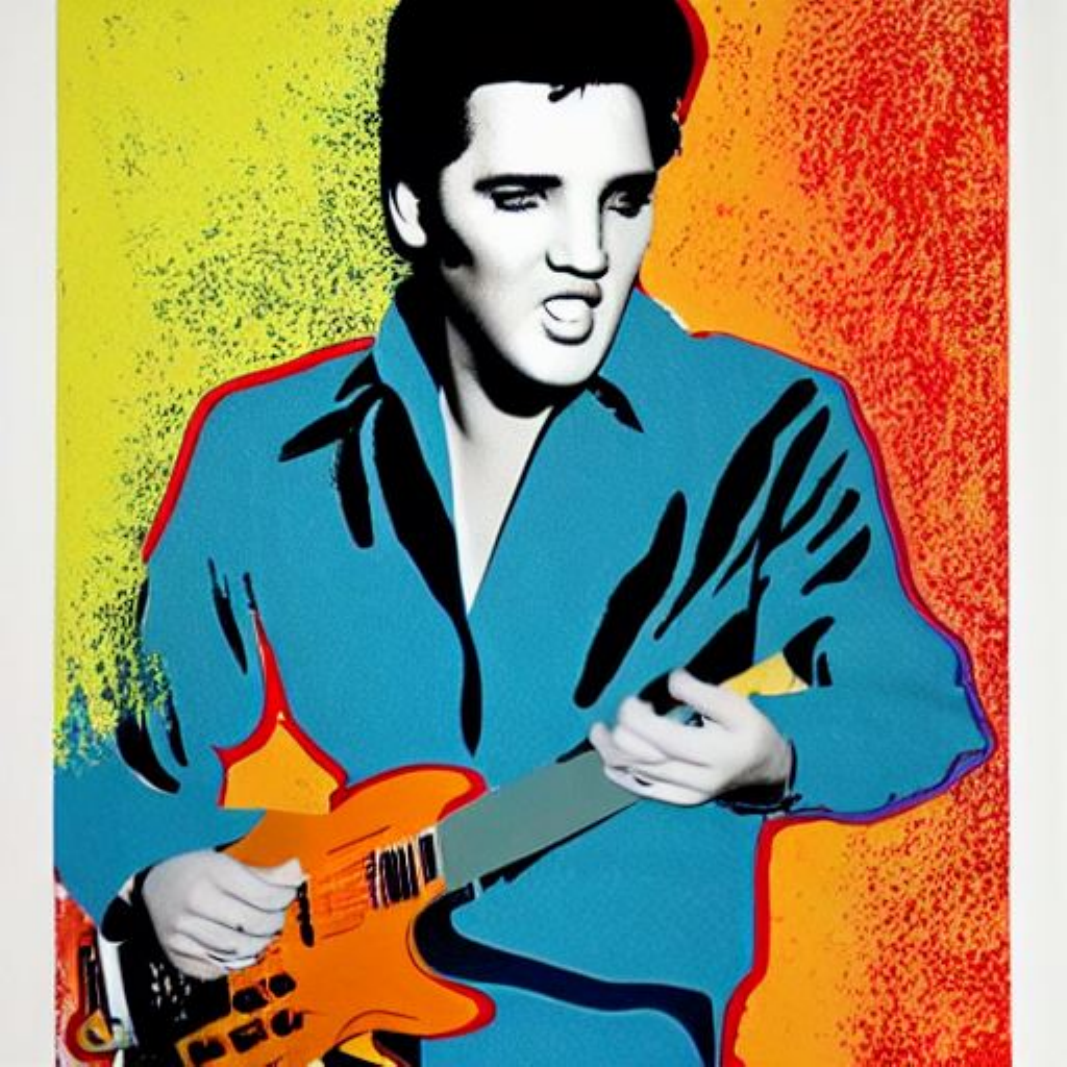}
        \end{minipage} &
        \begin{minipage}{.14\textwidth}
        \includegraphics[width=\linewidth]{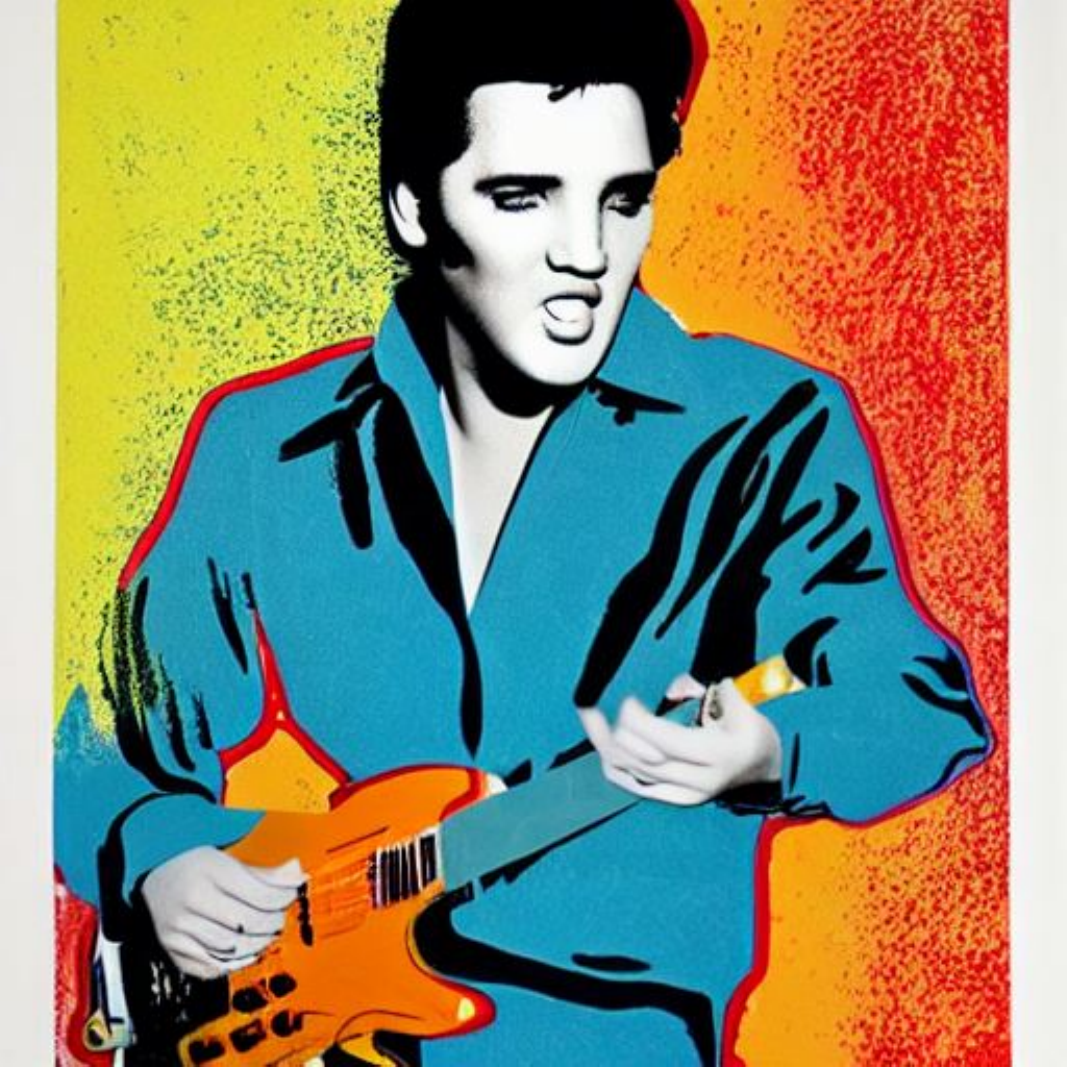}
        \end{minipage} 
        
    \end{tabular}
    \end{adjustbox}
    \caption{Image visualization of Artistic Style Erasure. Upper: Erasing ``Van Gogh". lower: Erasing ``Rembrandt".}
    \label{fig:artistcomparison}
\end{figure}
the ``Snoopy'' concept, achieving a 0.12 $\mathrm{LPIPS}_\mathrm{e}$ improvement.
Across single concept and multi-concept erasure, our method achieves the highest $\mathrm{LPIPS}_\mathrm{da}$ score, surpassing the second-best method by approximately 0.1.
This demonstrates that our method achieves a much better trade-off between erasure and preservation capabilities.

% For single concept removal, ESD yields significant impairment to the generative ability of other cartoon concepts, suggesting that ESD isn't capable of decoupling these cartoon concepts. 
% Although the preservation effect of UCE has improved compared to ESD, it remains unsatisfactory. 
% An examination of the $\mathrm{LPIPS}_\mathrm{u}$ reveals that our method and SPM demonstrate incredible retention of other concepts in comparison to MACE. 
% Furthermore, our method exhibits a more pronounced ability to erase these concepts than SPM while maintaining good generative power. Remarkably, for multi-concept erasure, erasure capability of our model is continuously improved with relatively little impairment to model's generation capability.

\subsection{Artistic Style Erasure}

\subsubsection{Experiment Setup}

ESD has provided 20 artist-specific prompts for each of five renowned artists, Van Gogh, Picasso, Rembrandt, Andy Warhol, and Caravaggio. The images generated with these prompts accurately capture each artist's style. %Experiment is conducted with "Van Gogh", "Rembrandt", "Andy Warhol" styles. 
We remove specific artist styles from the model, qualitatively and quantitatively evaluating the erasure effect of the target style and the preservation effects of other artist styles. 
We also utilize $\mathrm{LPIPS}_\mathrm{e}$, $\mathrm{LPIPS}_\mathrm{u}$, and $\mathrm{LPIPS}_\mathrm{da}$ as the quantitative metrics.
% The hyperparameters of our model are consistent with the Cartoon Concept Removal~\cite{}.

\begin{table}[h]
\centering
\begin{adjustbox}{max width=0.47\textwidth}
\begin{tabular}{lccccccccccc}
\toprule
\multirow{2}{*}{Method} & \multicolumn{3}{c}{Erasing \textit{``Van Gogh"}} & \multicolumn{3}{c}{Erasing \textit{``Rembrandt"}} &  \\ \cmidrule(l){2-4} \cmidrule(l){5-7} & $\mathrm{LPIPS}_\mathrm{e}\uparrow$  & $\mathrm{LPIPS}_\mathrm{u}\downarrow$ & $\mathrm{LPIPS}_\mathrm{da}\uparrow$  & $\mathrm{LPIPS}_\mathrm{e}\uparrow$ & $\mathrm{LPIPS}_\mathrm{u}\downarrow$   & $\mathrm{LPIPS}_\mathrm{da}\uparrow$       \\ \midrule
ESD-X & \textbf{0.4} & 0.26 & 0.14 & \textbf{0.473} & 0.277 & 0.196 \\
UCE & 0.25 & 0.05 & 0.2 & 0.2 & 0.056 & 0.144 \\
MACE & 0.309 & 0.095 & 0.214 & 0.316 & 0.1 & 0.216 \\
SPM & 0.255 & \textbf{0.01} & \underline{0.245} & 0.363 & \textbf{0.011} & \underline{0.352} \\
Ours & \underline{0.383} & \underline{0.025} & \textbf{0.358} & \underline{0.458} & \underline{0.043} & \textbf{0.415} \\
\bottomrule
\end{tabular}

\end{adjustbox}
\caption{LPIPS scores for Artistic Style Erasure. \textbf{Bold}: best. \underline{Underline}: second-best.}
\label{tab:metrics of artist erasure}
\vspace{-0.5cm}
\end{table}

\subsubsection{Results of Artistic Style Erasure}
Fig. \textcolor{red}{\ref{fig:artistcomparison}} visualizes the crafted images after erasure.
We can observe that our method performs well in erasing the global artist style of images and preserving non-target concepts. 
However, the images generated by ESD in Van Gogh style have structures that differ significantly from the original image. It suggests that ESD exhibits excessive erasure for the concept ``Van Gogh".
Besides, MACE has a adverse impact on the generative capability. For example, the image generated by MACE in Rembrandt style (the second row) looks significantly different from the original images.
When erasing the concept ``Rembrandt", SPM fails to remove it completely (the fourth row).
% depicts that our method performs well on erasing the global artist style of images and preserving non-target concepts. ESD exhibits excessive erasure for concept ``Van Gogh" and MACE has a pronounced impact on the generative capability. For the concept ``Rembrandt", SPM cannot erase completely.

The quantitative result is presented in Tab. \textcolor{red}{\ref{tab:metrics of artist erasure}}. Similar to the results of Tab. \textcolor{red}{\ref{tab:metrics of Cartoon Concept Removal}}, our method also achieves the highest $\mathrm{LPIPS}_\mathrm{da}$. This demonstrates the scalability of our method across erasing different concepts.

% our method achieves a good trade-off between preservation and erasure capabilities.

\begin{table}[h]
\centering
\begin{adjustbox}{max width=0.47\textwidth}
\begin{tabular}{cccc}
\toprule
& $\mathrm{LPIPS}_\mathrm{e}\uparrow$  & $\mathrm{LPIPS}_\mathrm{u}\downarrow$ & $\mathrm{LPIPS}_\mathrm{da}\uparrow$  \\ \midrule 
Timestep & \underline{0.338} & \underline{0.020} & \underline{0.318} \\
Layer & 0.194 & \textbf{0.01} & 0.184\\
Timestep+Layer & \textbf{0.383} & 0.025 & \textbf{0.358} \\
\bottomrule
\end{tabular}

\end{adjustbox}
\caption{LPIPS scores for different modulation settings. \textbf{Bold}: best. \underline{Underline}: second-best.}
\label{tab:ablation1}
\vspace{-0.5cm}
\end{table}

\subsection{Ablation Study}
%We perform two ablation experiments. The first one confirms that both temporal information and layer information in the modulation process are useful in obtaining a trade-off between retention and erasure effects. The second one displays the effect of EPR and the Modulation Process.

\subsubsection{Effect of timestep and layer modulation factors for TLMO.}

To evaluate the effect of timestep modulation factors and layer modulation factors, we conducted three separate experiments on the concept 'Van Gogh', utilizing only timestep factors, only layer factors, both timestep and layer factors. Quantitative results on $\mathrm{LPIPS}_\mathrm{e}$, $\mathrm{LPIPS}_\mathrm{u}$ and $\mathrm{LPIPS}_\mathrm{da}$ scores are shown in Tab. \textcolor{red}{\ref{tab:ablation1}}. From $\mathrm{LPIPS}_\mathrm{da}$, we can conclude that two types of modulation factors are all helpful in striking a balance between erasing ability and generative ability. Meanwhile, the optimal erasure results are attained when both factors are employed concurrently.

\begin{figure}[t]
    \centering
    \begin{adjustbox}{max width=0.4\textwidth}
    \begin{tabular}{c: c@{\hskip 0.06in} c@{\hskip 0.06in} c@{\hskip 0.06in}}
        \textbf{Original} & \textbf{Finetune} & \textbf{EPR} & \textbf{EPR+TLMO} \\

        \begin{subfigure}{.14\textwidth}
        \includegraphics[width=\linewidth]{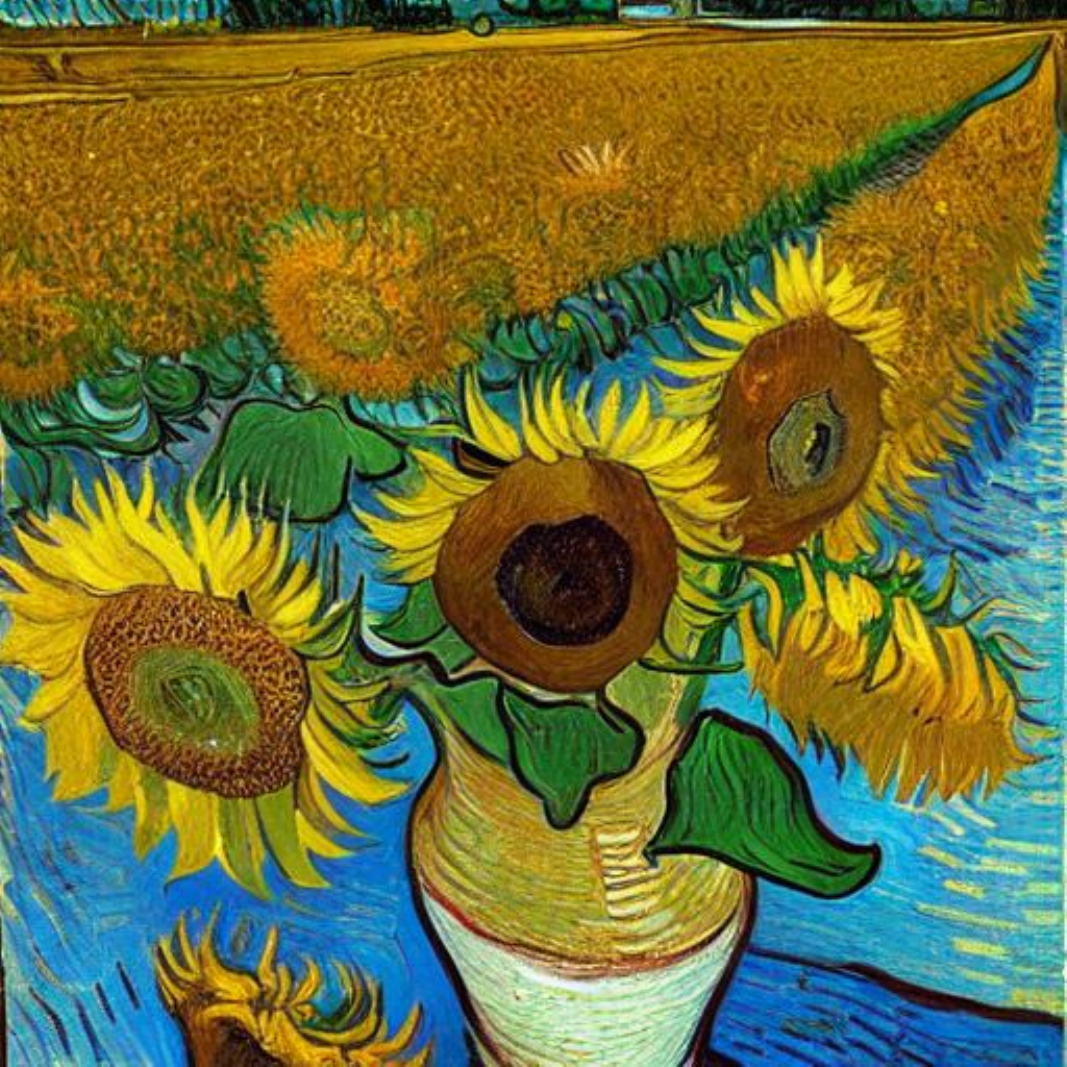}
        \end{subfigure} &
        \begin{subfigure}{.14\textwidth}
        \includegraphics[width=\linewidth]{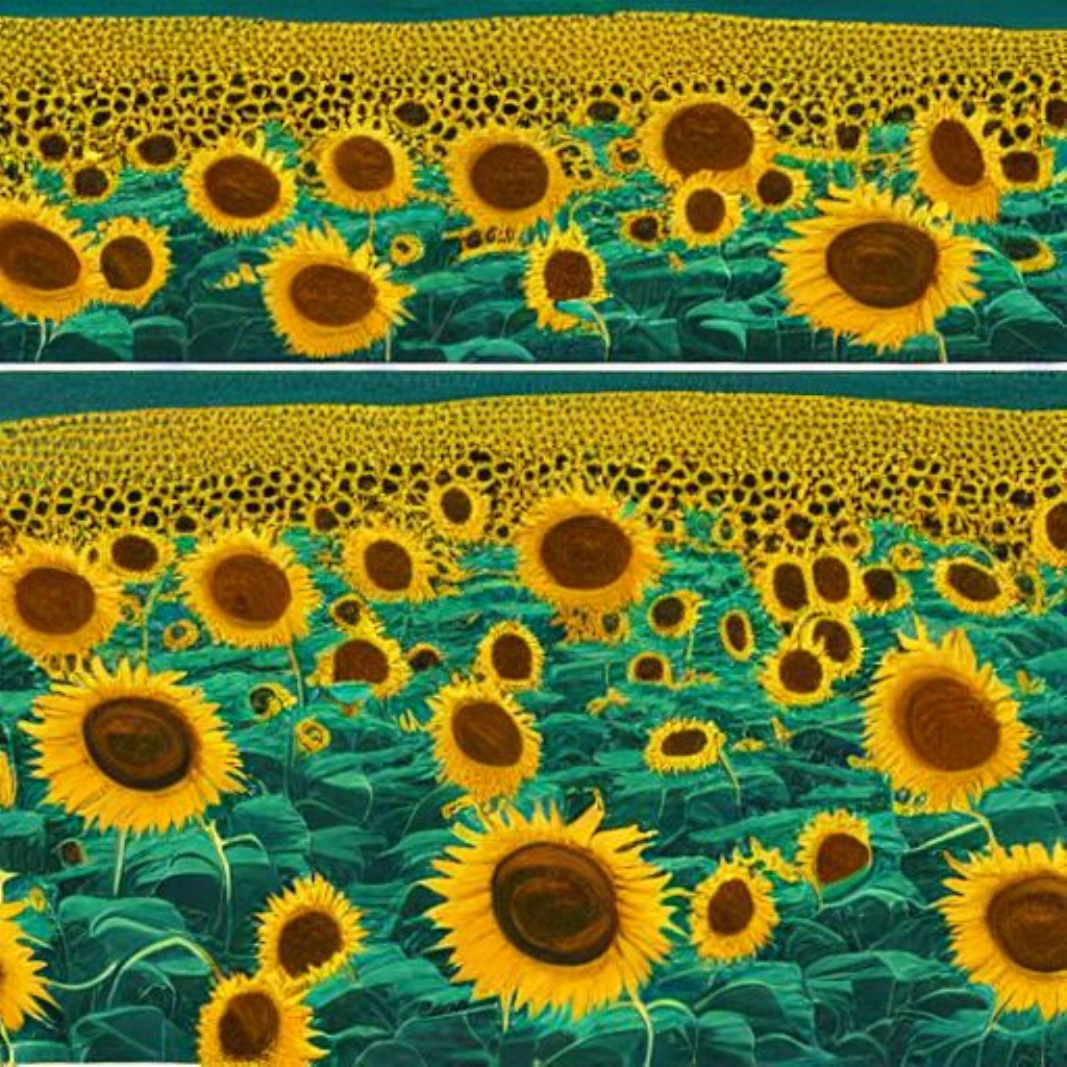}
        \end{subfigure} &
        \begin{subfigure}{.14\textwidth}
        \includegraphics[width=\linewidth]{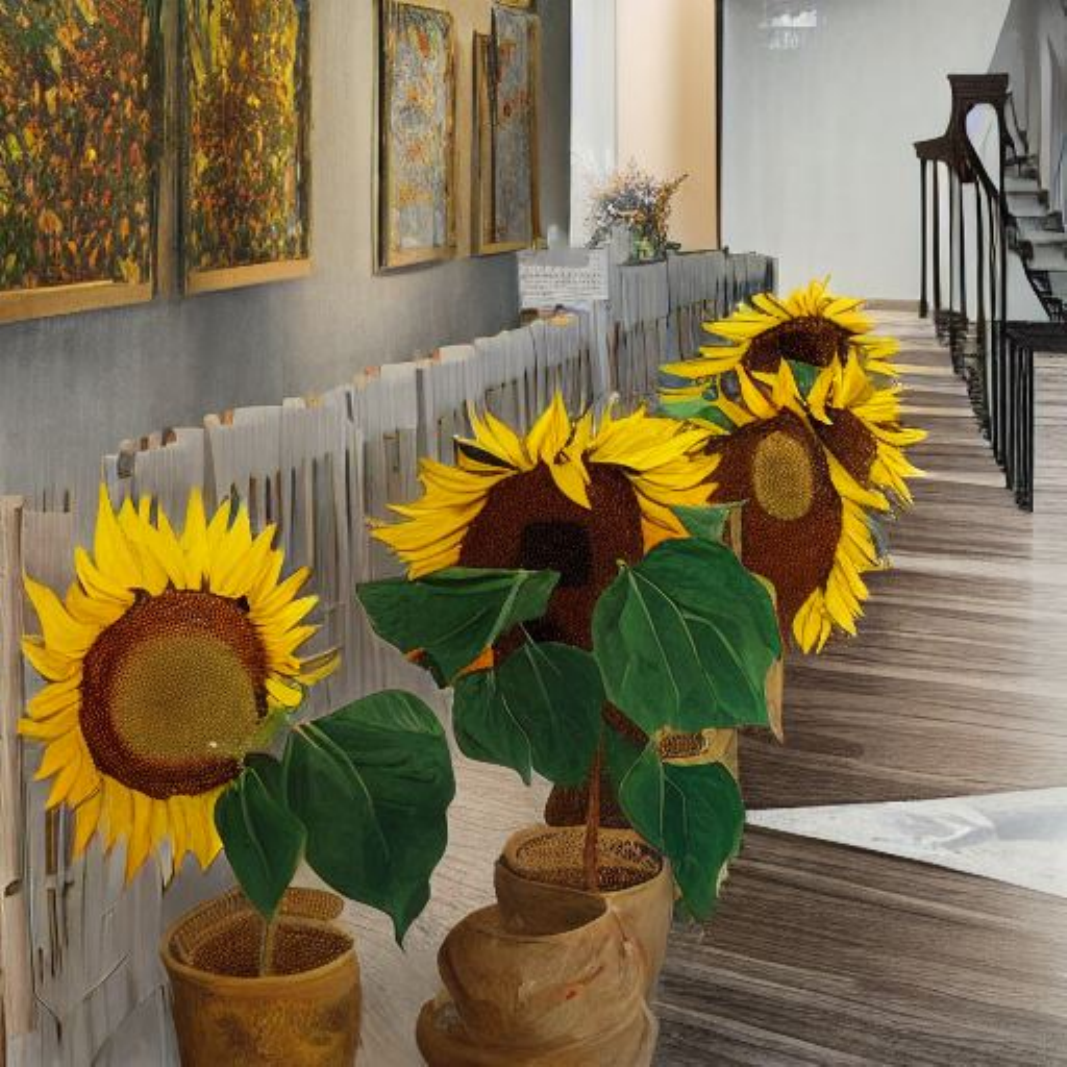}
        \end{subfigure} &
        \begin{subfigure}{.14\textwidth}
        \includegraphics[width=\linewidth]{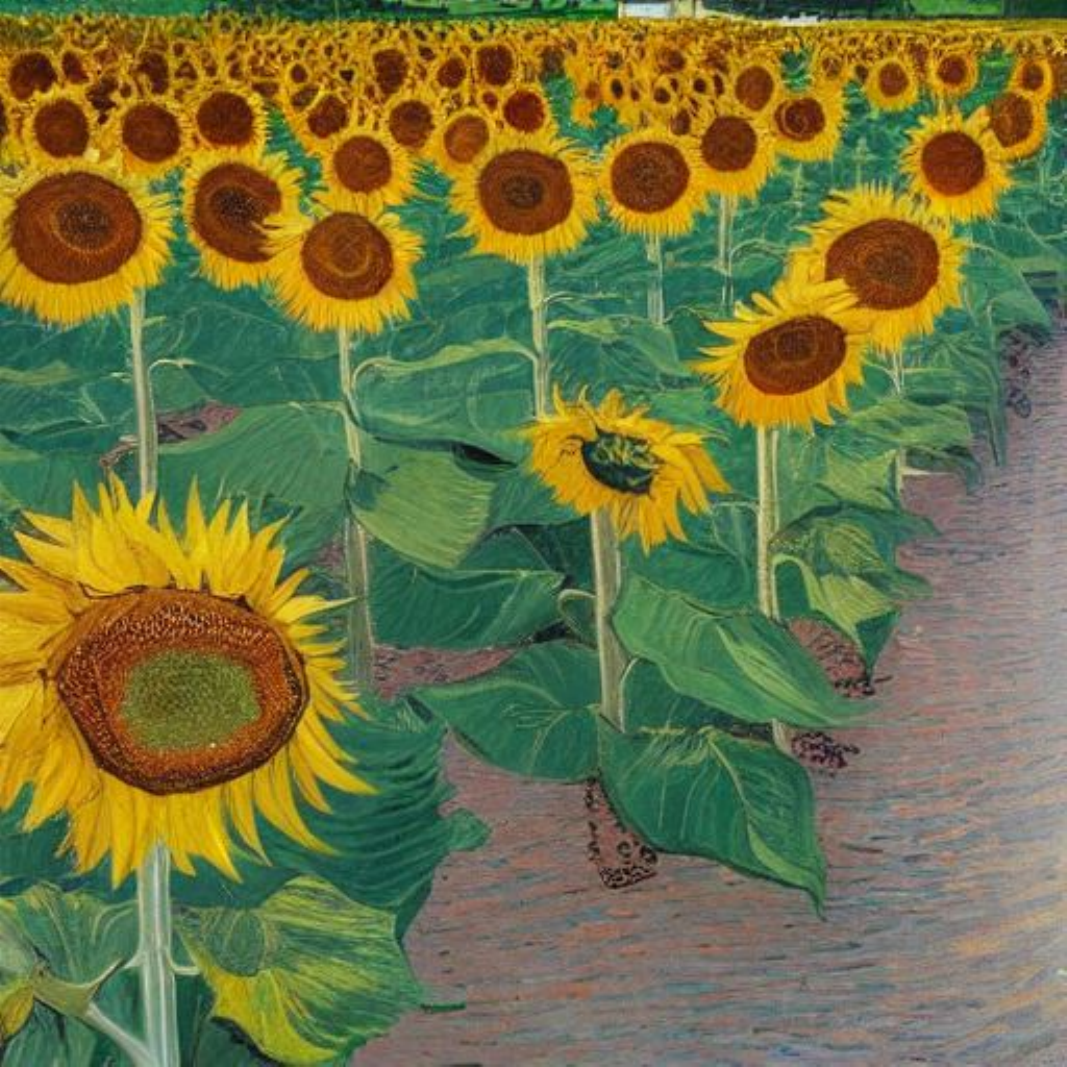}
        \end{subfigure} \\

        \begin{subfigure}{.14\textwidth}
        \includegraphics[width=\linewidth]{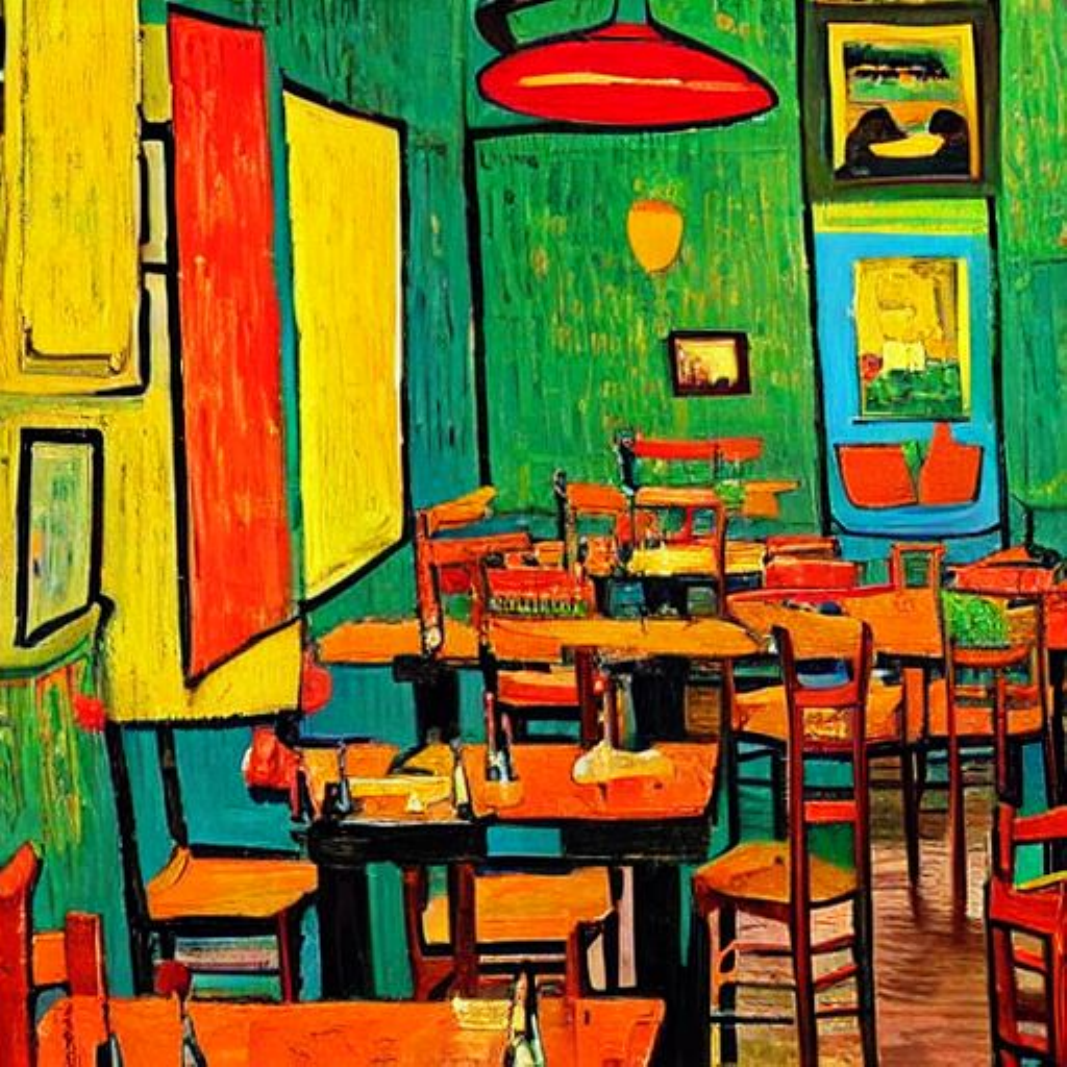}
        \end{subfigure} &
        \begin{subfigure}{.14\textwidth}
        \includegraphics[width=\linewidth]{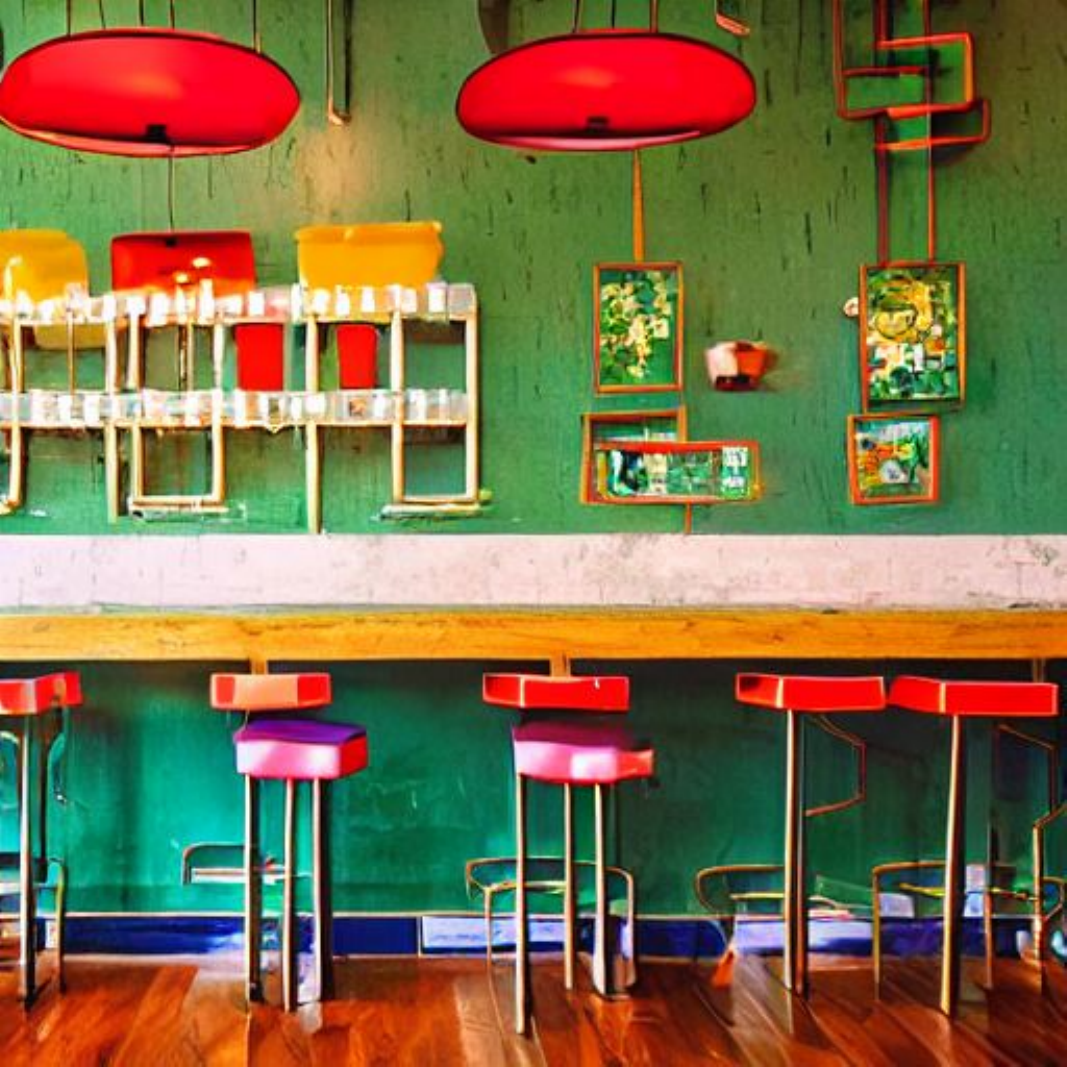}
        \end{subfigure} &
        \begin{subfigure}{.14\textwidth}
        \includegraphics[width=\linewidth]{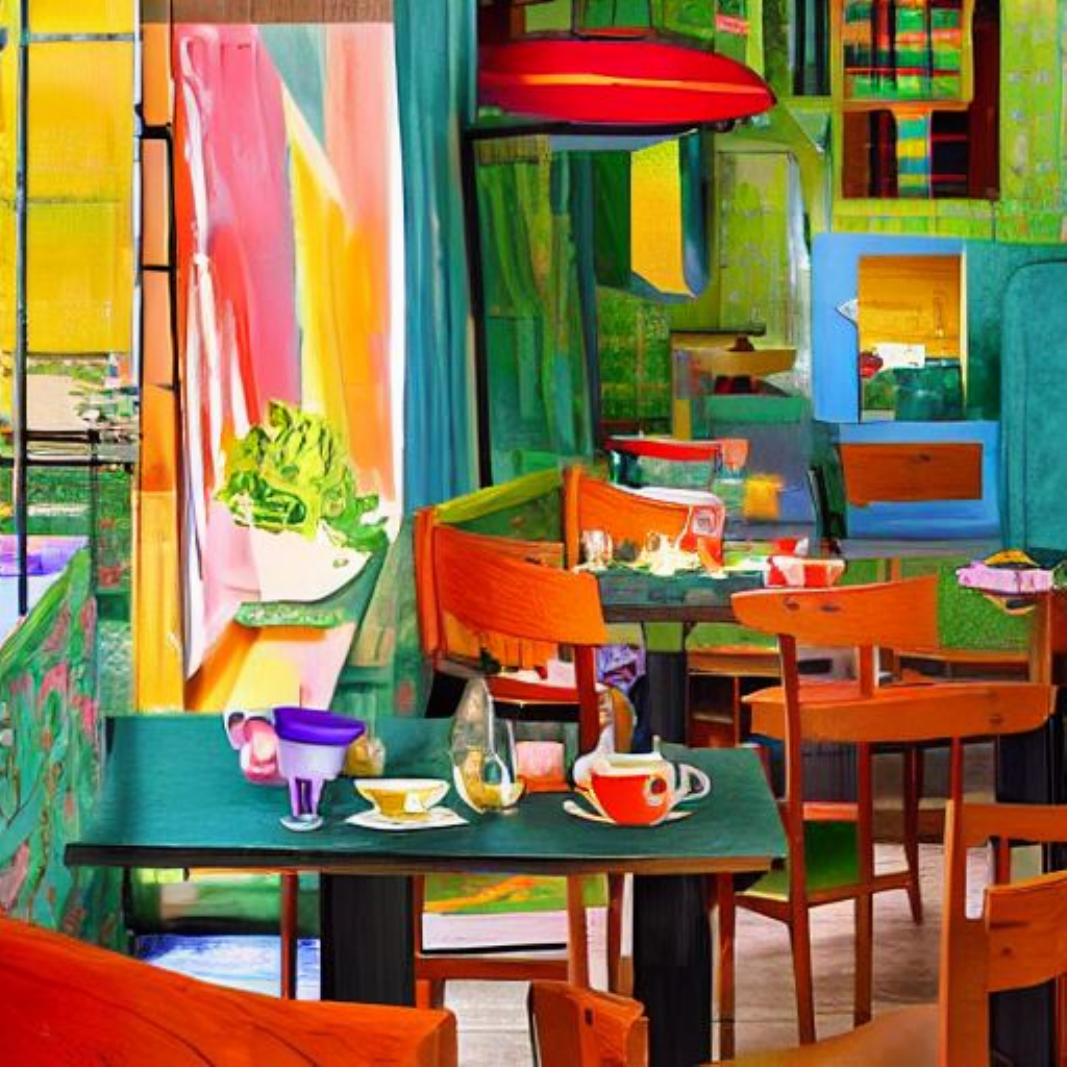}
        \end{subfigure} &
        \begin{subfigure}{.14\textwidth}
        \includegraphics[width=\linewidth]{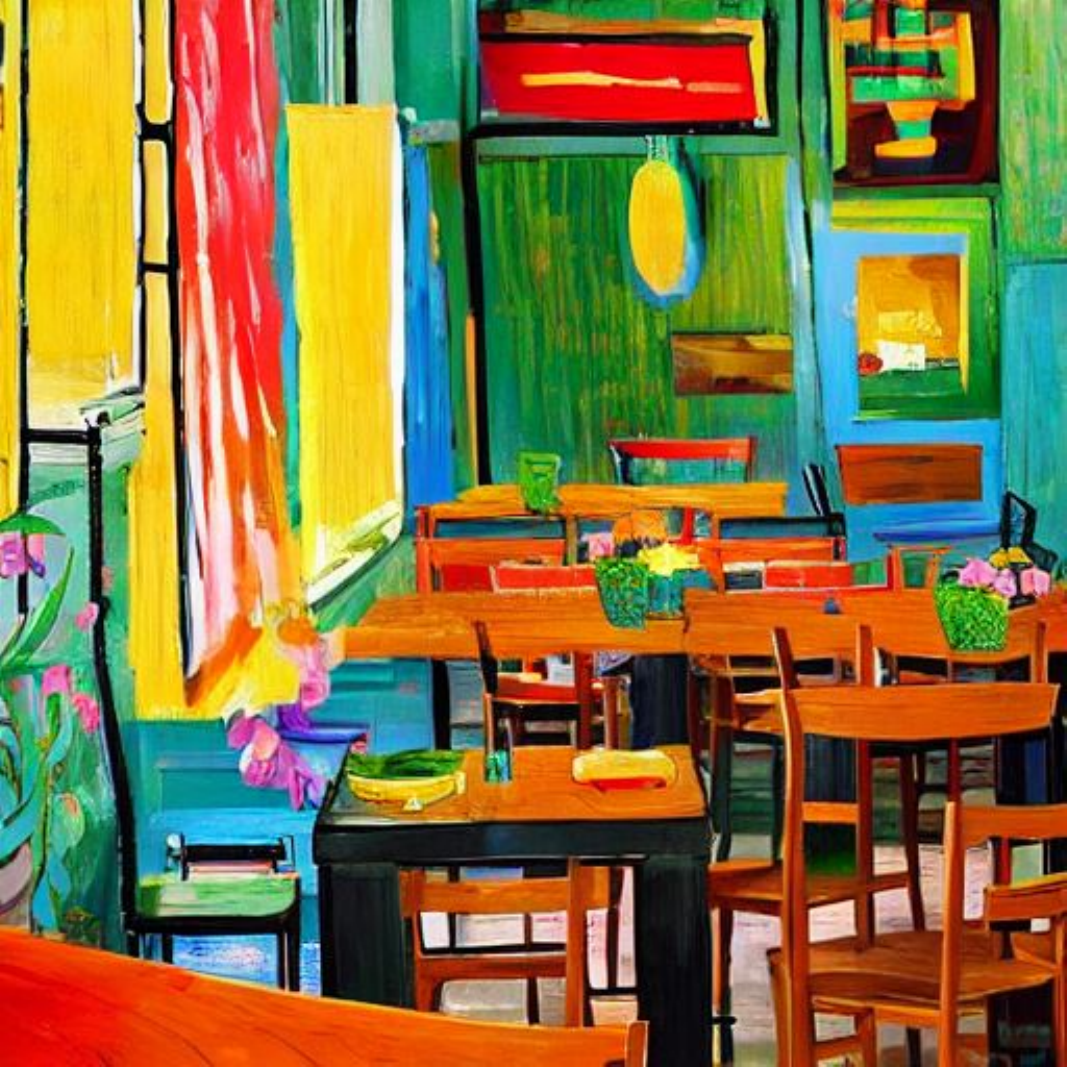}
        \end{subfigure} \\

    \end{tabular}
    \end{adjustbox}
    \caption{Image visualization of our Component verification experiment.}
    \label{fig:ablation2comparison}
    \vspace{-0.5cm}
\end{figure}

\begin{table}[h]
\centering
\begin{adjustbox}{max width=0.47\textwidth}
\begin{tabular}{cccc}
\toprule
 & $\mathrm{LPIPS}_\mathrm{e}\uparrow$  & $\mathrm{LPIPS}_\mathrm{u}\downarrow$ & $\mathrm{LPIPS}_\mathrm{da}\uparrow$  \\ \midrule
Finetune & 0.37 & 0.24 & 0.13 \\
EPR & \textbf{0.459} & \underline{0.033} & \textbf{0.426} \\
EPR+TLMO & \underline{0.383} & \textbf{0.025} & \underline{0.358} \\
\bottomrule
\end{tabular}
\end{adjustbox}
\caption{LPIPS scores for Component Verification. \textbf{Bold}: best. \underline{Underline}: second-best.}
\label{tab:ablation2}
\vspace{-0.5cm}
\end{table}

\subsubsection{Component Verification of DuMo}

We then verify the impact of the EPR and TLMO in DuMo on the concept "Van Gogh". In comparison to directly fine-tuning the cross-attention layers of the encoder, applying the EPR module enables a substantial enhancement in preserving non-target concepts, with an improvement in $\mathrm{LPIPS}_\mathrm{u}$ from 0.24 to 0.033, a substantial increase of 0.207 (Tab. \textcolor{red}{\ref{tab:ablation2}}). Besides, Compared to using EPR alone, result in Fig. \textcolor{red}{\ref{fig:ablation2comparison}} shows that the combination of EPR and TLMO better preserves the structural integrity of the image.

% Direct fine tuning to the encoder substantially alters the backbone feature, heavily damaging the structure of images (Fig. \textcolor{red}{\ref{fig:ablation2comparison}}). While the EPR module relieves this problem to a certain degree and achieves better $\mathrm{LPIPS}_\mathrm{e}$, it is not a comprehensive solution. Notably, TLMO effectively preserves the structure of images, facilitating precise erasure of the target concept and fabulous generative ability of non-target concepts.

\section{Conclusion}

In this paper, we propose a novel framework DuMo which conducts precise erasure for multiple concepts and ensure minimum impairment to non-target concepts generation. While previous method alters both the backbone feature and the skip connection feature, they destruct the generative ability of DM. Our EPR module only modifies the skip connection feature and utilizes the prior knowledge of original skip features to mitigate the impact on non-target concepts. Furthermore, to strengthen the generative ability and realise precise erasure of target concepts, a novel Timestep-Layer Modulation process is introduced to calibrate each output of the EPR module during the denoising process. We believe that DuMo would be illuminating to T2I providers in mitigating the generation of insecure content, thereby contributing to the advancement of a safer AI community.

\section{Acknowledgements}

This project was supported by National Key R\&D Program of China (No.
2021ZD0112804).

\bibliography{aaai25}

% \onecolumn
% \input{checklist}

\end{document}